\documentclass[11pt]{article}

\usepackage[margin=1in]{geometry}
\usepackage{authblk}

\usepackage[numbers]{natbib}

\usepackage{amsmath,amssymb,amsfonts,mathtools}
\usepackage{graphicx}
\usepackage{tikz}
\usetikzlibrary{arrows.meta,positioning,shapes,fit,calc}
\usepackage{pgfplots}
\pgfplotsset{compat=1.17}
\usepackage{booktabs}
\usepackage{enumitem}
\usepackage[hidelinks]{hyperref}
\usepackage{algorithm}
\usepackage{algorithmic}
\usepackage[expansion=false]{microtype}
\usepackage{subcaption}
\usepackage{xcolor}
\usepackage{multirow}
\usepackage{colortbl}
\usepackage{float}  

\graphicspath{{Images/}}

\setlist[itemize]{leftmargin=*,itemsep=2pt,topsep=3pt}
\setlist[enumerate]{leftmargin=*,itemsep=2pt,topsep=3pt}

\newcommand{\modelname}{GraphPerf-RT}

\newcommand{\omitcontent}[1]{}

\newcommand{\suppref}[2]{Appendix~\ref{#1}}

\definecolor{MahdiC}{RGB}{0,0,255}
\definecolor{MohammadC}{RGB}{0,90,200}
\definecolor{SaeedC}{RGB}{0,150,140}

\begin{document}

\title{\modelname: Graph-Driven Performance Modeling with Calibrated Uncertainty for OpenMP Scheduling on Heterogeneous Embedded SoCs}

\author[1]{Mohammad Pivezhandi}
\author[2]{Mahdi Banisharif}
\author[1]{Saeed Bakhshan}
\author[3]{Abusayeed Saifullah}
\author[2]{Ali Jannesari}

\affil[1]{Wayne State University}
\affil[2]{Iowa State University}
\affil[3]{The University of Texas at Dallas}

\date{}

\maketitle

\begin{abstract}
Autonomous AI agents on embedded platforms require real-time, risk-aware scheduling under resource and thermal constraints. Classical heuristics struggle with workload irregularity, tabular regressors discard structural information, and model-free reinforcement learning (RL) risks overheating. We introduce \modelname{}, a graph neural network surrogate achieving deep learning accuracy at heuristic speeds (2--7ms). \modelname{} is, to our knowledge, the first to unify task DAG topology, CFG-derived code semantics, and runtime context (per-core DVFS, thermal state, utilization) in a heterogeneous graph with typed edges encoding precedence, placement, and contention. Evidential regression with Normal-Inverse-Gamma priors provides calibrated uncertainty; we validate on makespan prediction for risk-aware scheduling. Experiments on three ARM platforms (Jetson TX2, Orin NX, RUBIK Pi) achieve $R^2 = 0.81$ on log-transformed makespan with Spearman $\rho = 0.95$ and conservative uncertainty calibration (PICP $= 99.9\%$ at 95\% confidence). Integration with four RL methods demonstrates that multi-agent model-based RL with \modelname{} as the world model achieves 66\% makespan reduction and 82\% energy reduction versus model-free baselines, with zero thermal violations.
\end{abstract}

\section{Introduction}


Autonomous AI agents deployed on heterogeneous embedded Systems-on-Chip (SoCs) must make real-time scheduling decisions under stringent resource, energy, and thermal constraints. These agents, which power autonomous vehicles, robotic systems, and edge AI applications, require accurate performance prediction with calibrated uncertainty to ensure safe operation. Heterogeneous embedded SoCs combine high-performance and energy-efficient cores with Dynamic Voltage and Frequency Scaling (DVFS). This architecture has created unprecedented opportunities for deploying parallel applications in resource-constrained environments. OpenMP, the dominant shared-memory parallel programming model, enables developers to express task-level parallelism through pragma-based annotations that the runtime system maps to available hardware resources. However, achieving optimal performance on these heterogeneous platforms requires careful scheduling decisions that balance execution time, energy efficiency, and thermal constraints. This presents a challenging optimization problem that current approaches fail to address comprehensively. Performance prediction for parallel workloads has traditionally relied on analytical models, simulation-based methods, and machine learning surrogates. Analytical models grounded in queueing theory and scheduling bounds such as Brent's theorem provide theoretical insights but require simplifying assumptions that break down under irregular control flow and cache contention~\cite{graham1969bounds,brent1974parallel}. Simulation-based approaches offer high fidelity but incur prohibitive runtime overhead for online scheduling decisions. Machine learning surrogates, including Graph Neural Networks (GNNs), can learn complex input-output mappings from data~\cite{velickovic2017graph}. However, existing approaches such as ProGraML~\cite{cummins2021programl} and Ithemal~\cite{mendis2019ithemal} operate on static intermediate representations without incorporating runtime context, thermal state, or DVFS configurations that significantly affect embedded system performance. Complementary approaches address related challenges through hierarchical multi-agent DVFS scheduling, zero-shot LLM-guided allocation, statistical feature-aware task allocation, and flow-augmented few-shot RL [Anonymous, 2026; details omitted for double-blind review].

This paper addresses the problem of predicting performance metrics for OpenMP task-parallel applications on heterogeneous embedded SoCs. The framework supports multiple targets including execution time, energy consumption, and hardware counters; we focus on makespan prediction under varying DVFS configurations, core allocations, and thermal conditions. The prediction model must provide accurate point estimates and \emph{calibrated uncertainty quantification} to enable risk-aware scheduling decisions. This is essential for respecting thermal constraints and avoiding failures in safety-critical deployments. Furthermore, the model should generalize across benchmarks, input sizes, and hardware platforms without extensive retraining.

Accurate performance prediction on heterogeneous embedded SoCs presents four challenges. \textbf{First}, performance emerges from cross-layer interactions between application structure (task DAGs, CFGs), hardware state (frequencies, thermal headroom), and scheduling decisions; classical heuristics treat these independently, missing critical coupling effects that cause 3--5$\times$ execution time variation. \textbf{Second}, tabular models flatten task graphs into aggregate statistics, discarding dependency structure, while GNN approaches focus on static representations without runtime context. \textbf{Third}, standard regression lacks confidence estimates needed for risk-aware scheduling on thermally constrained systems. \textbf{Fourth}, model-free RL requires extensive on-device exploration risking overheating, while model-based RL needs an accurate environment model.

\begin{figure}[t]
\centering
\includegraphics[width=\columnwidth]{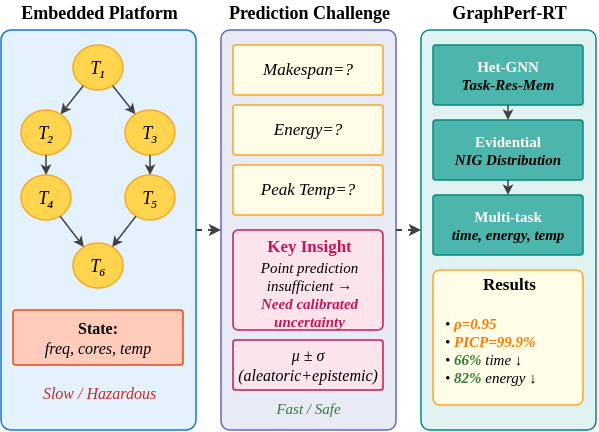}
\caption{\modelname{} as a world model for safe, uncertainty-aware scheduling.}
\label{fig:motivation}
\end{figure}

This paper introduces \modelname{}, an AI technology that provides foundational infrastructure for safe, uncertainty-aware scheduling on embedded platforms. \modelname{} addresses these challenges through a unified heterogeneous graph representation (see Fig.~\ref{fig:motivation}). Task nodes encode CFG-derived code semantics; resource nodes capture per-core DVFS state, utilization, and thermal headroom; typed edges encode task-task precedence, task-resource placement, and resource-resource topology. A heterogeneous Graph Attention Network (GAT) with type-specific encoders processes the graph through 3--6 attention layers. Evidential learning with Normal-Inverse-Gamma (NIG) distributions provides calibrated uncertainty in a single forward pass. The surrogate scores candidate configurations, filters low-confidence proposals, and supports Dyna-style model-based planning. Comprehensive experiments on three embedded ARM platforms (NVIDIA Jetson TX2, Jetson Orin NX, and RUBIK Pi) span 42 benchmarks from BOTS and PolyBench. \modelname{} achieves $R^2 = 0.81$ on log-transformed makespan with Spearman $\rho = 0.95$ for ranking and conservative uncertainty calibration (PICP $= 99.9\%$ at 95\% confidence), significantly outperforming tabular baselines and homogeneous GNN architectures. Integration with four RL methods (SAMFRL, SAMBRL, MAMFRL-D3QN, and MAMBRL-D3QN) demonstrates that MAMBRL-D3QN with \modelname{} achieves 66\% execution time reduction ($0.97 \pm 0.35$s vs.\ $2.85 \pm 1.66$s) and 82\% energy reduction ($0.006 \pm 0.005$J vs.\ $0.033 \pm 0.026$J) compared to model-free baselines across 5 random seeds with 200 episodes each.

The main contributions of this paper, which establishes \modelname{} as AI infrastructure for autonomous embedded agents, are:
\begin{enumerate}
    \item A \emph{unified heterogeneous graph representation} for AI agents that jointly encodes OpenMP task DAG topology, CFG-derived code semantics, and runtime context (per-core DVFS, thermal state, utilization) through typed nodes and edges, enabling explicit modeling of cross-layer performance interactions critical for autonomous decision-making.

    \item An \emph{evidential prediction framework} supporting multi-task learning with calibrated uncertainty quantification through Normal-Inverse-Gamma heads, producing both aleatoric and epistemic uncertainty in a single forward pass. We validate on makespan prediction; the architecture extends to energy and hardware counter targets. This provides essential AI safety infrastructure for risk-aware scheduling on embedded systems where overconfident predictions can cause thermal violations or hardware damage.

    \item A \emph{practical AI agent integration} demonstrating seamless combination with model-based RL, achieving 66\% makespan and 82\% energy improvements over model-free baselines while reducing hazardous on-device exploration through Dyna-style synthetic rollouts enabled by \modelname{} as a world model.

    \item A \emph{reproducible AI4Tech evaluation framework} including a complete data pipeline (OMPi + ALF-llvm + SWEET + telemetry), extensive experiments across three ARM platforms and 42 benchmarks, and statistical significance tests with 5-seed confidence intervals.
\end{enumerate}

\section{Related Work}
\label{sec:related}

\paragraph{Performance Modeling.}
Performance modeling spans analytical queueing/Petri-net models~\cite{eager1983bound,marsan1984class}, statistical surrogates~\cite{lee2007accurate,ipek2006efficiently,wang2018machine,chen2016xgboost}, and neural predictors. Black-box predictors using counters or instruction mixes improve accuracy but lose structural signals. Ithemal~\cite{mendis2019ithemal} predicts x86 basic-block throughput via hierarchical LSTMs but operates on static assembly without runtime parallelism or thermal state. Runtime autotuners like Apollo~\cite{ramadan2024} adapt to input-dependent variations yet remain kernel-focused. We model OpenMP task DAGs on embedded SoCs, encoding dependencies, per-core DVFS, and thermal context for real-time decisions.

\paragraph{Graph Neural Networks.}
GNNs enable learning on structured program graphs~\cite{wu2020comprehensive,velickovic2017graph}. ProGraML~\cite{cummins2021programl} builds unified graphs from LLVM IR for compiler tasks but operates on \emph{static} IR without runtime or thermal context. HGT~\cite{hu2020heterogeneous} extends graph attention to multiple node/edge types but lacks: (1)~edge decomposition separating structure from execution features, (2)~thermal/DVFS integration, and (3)~evidential heads for calibrated uncertainty. GNNs have predicted hardware-dependent performance~\cite{shi2022graph} and optimization effects~\cite{brauckmann2020compiler}. We construct heterogeneous task-resource graphs with CFG-informed encoders and runtime context fusion.

\paragraph{Compiler Cost Models.}
Ansor, MetaSchedule, and ROLLER learn kernel-level evaluators with cross-device transfer~\cite{zheng2020ansor,shao2022metaschedule,zhu2022roller,zhai2023tlp,zhai2024tlm}, but operate below task graphs and rarely encode DVFS/thermal state. Silhouette~\cite{papon2022silhouette} learns CPU embeddings for cross-platform prediction; Glimpse~\cite{ahn2022glimpse} develops mathematical hardware encodings for neural compilation. These focus on static kernel prediction without dynamic DVFS, thermal headroom, or task DAG structure. \modelname{} integrates typed edges, execution features, and runtime device/thermal context into a unified graph representation.

\paragraph{Hardware-Aware ML and DVFS.}
Hardware-aware ML includes NAS with device constraints~\cite{cai2018proxylessnas} and cross-platform predictors~\cite{walker2018accurate}. Energy/thermal DVFS spans heuristics, optimization, and RL~\cite{xie2021survey}. Utilization-driven governors mis-scale frequency by conflating compute and stalls~\cite{hebbar2022pmu,lin2023workload}. Model-free RL adapts but is sample- and heat-intensive~\cite{kim2021ztt}; model-based control reduces sampling via learned dynamics~\cite{moerland2023model}. Recent work includes hierarchical multi-agent DVFS, feature-aware allocation, zero-shot LLM guidance, and flow-augmented RL [Anonymous, 2026]. \modelname{} is complementary: a calibrated graph-grounded evaluator supplying fast rollouts to MARL under thermal constraints.

\paragraph{Uncertainty Quantification.}
Uncertainty is critical for safe scheduling. Bayesian and ensemble methods add inference cost~\cite{lakshminarayanan2017simple,kendall2017uncertainties}. Evidential learning yields calibrated intervals without sampling~\cite{sensoy2018evidential,amini2020deep}. Early formulations suffered from evidence contraction~\cite{wu2024non}, which we address through non-saturating regularization. We adopt evidential heads to expose confidence on makespan/energy and gate synthetic rollouts for real-time planning.

\omitcontent{\paragraph{Language Model Approaches.}
LLMs show promise for code understanding and performance prediction~\cite{akhauri2025regression}, but face constraints for real-time embedded scheduling: inference latency exceeds hundreds of milliseconds, and they cannot efficiently ingest dynamic numerical state without expensive tokenization. \modelname{} delivers deep learning accuracy with heuristic-level speed (2--7~ms), native numerical state injection, and calibrated uncertainty. Table~\ref{tab:llm-comparison} summarizes this trade-off.

\begin{table}[h]
\centering
\small
\caption{Pareto trade-off: \modelname{} vs. Language Model Approaches}
\label{tab:llm-comparison}
\begin{tabular}{lcc}
\toprule
\textbf{Capability} & \textbf{\modelname{}} & \textbf{LLM} \\
\midrule
Inference Latency & 2--7 ms & $>$200 ms \\
Memory Footprint & 12.4 MB & $>$4 GB \\
Dynamic State Injection & Native (float) & Tokenization \\
Uncertainty Quantification & Calibrated NIG & Hallucination risk \\
Real-time Scheduling & Yes & No \\
\bottomrule
\end{tabular}
\end{table}}

To our knowledge, \modelname{} is the \emph{first} unified graph representation combining OpenMP task DAGs with runtime context (per-core DVFS, thermal headroom, counters) and evidential regression for calibrated uncertainty in embedded scheduling. This enables risk-aware decisions under thermal constraints while maintaining computational efficiency for on-board deployment.

\section{Preliminaries}
\label{sec:preliminaries}

\subsection{Problem Formulation}
\label{subsec:problem-setup}

\suppref{app:problem-details}{Extended formal definitions are provided in the supplementary material.}

\omitcontent{This work addresses the problem of predicting multiple performance metrics for OpenMP task-parallel applications executing on heterogeneous embedded SoCs under varying DVFS configurations. The prediction model must provide accurate point estimates along with calibrated uncertainty quantification to enable risk-aware scheduling decisions.}

\omitcontent{Each OpenMP application is represented as a task DAG $G=(V,E)$ recovered from compiler artifacts and runtime instrumentation. A task $v\in V$ represents a unit of parallel work characterized by static code features extracted from the CFG, including loop counts, memory access patterns, and branch statistics. Additionally, each task carries optional runtime summaries from prior executions, such as performance counter snapshots and thermal footprints. An edge $e=(u\to v)\in E$ encodes a precedence relation derived from OpenMP dependencies, classified as spawn, join, or data dependency types.}

We predict multiple performance metrics for OpenMP task-parallel applications on heterogeneous embedded SoCs under varying DVFS configurations. Each application is a task DAG $G=(V,E)$ with tasks $v\in V$ characterized by CFG-derived features and precedence edges $e\in E$ (spawn, join, or data dependencies). The platform has $C$ cores with discrete frequency levels $f_i\in\mathcal{F}_i$. The device sheet $D$ captures immutable hardware constants (core types, cache specs, DVFS tables), while runtime state $\mathcal{S}$ includes per-core frequencies, utilization, and thermal readings.

\omitcontent{The target platform consists of $C$ heterogeneous cores indexed by $i\in\{1,\dots,C\}$. Each core operates at discrete frequency levels $f_i\in\mathcal{F}_i$ determined by the DVFS subsystem. The immutable hardware characteristics are captured in the device sheet $D$, which includes core types and counts, cache hierarchy specifications (sizes, associativity, line sizes), DVFS tables per core cluster, governor support flags, thermal sensor layout and resolution, and toolchain version hashes for reproducibility. The device sheet remains constant throughout an episode and enables transfer learning across hardware generations.}

\omitcontent{The runtime state $\mathcal{S}$ captures dynamic system context that changes during execution. This includes the active core mask indicating which cores are available for scheduling, per-core DVFS indices and measured frequencies, utilization computed as an exponential moving average (EMA), thermal zone temperatures before and after execution, and recent performance counter values. These runtime features enable the model to adapt predictions based on current system conditions rather than relying solely on static analysis.}

A scheduling action $a=(m,f,p)$ specifies core mask $m\in\{0,1\}^C$, DVFS vector $f$, and optional priority $p$. Given $(G,D,\mathcal{S},a)$, we learn $\mathcal{M}: (G,D,\mathcal{S},a) \mapsto (y, \mathcal{U})$, where $y$ can include makespan, energy, or hardware counters depending on the target configuration. We focus on makespan prediction with calibrated uncertainty $\mathcal{U}$ via NIG parameters $(\gamma,\nu,\alpha,\beta)$.

\omitcontent{Given a tuple $(G,D,\mathcal{S},a)$, the surrogate model predicts a vector of performance metrics $y=(\text{makespan}, \text{energy}, \text{cache misses}, \text{branch misses}, \text{utilization})$. Lower values indicate better performance for makespan, energy, and miss metrics, while higher utilization is generally preferable. Each prediction carries calibrated uncertainty estimates decomposed into aleatoric uncertainty (irreducible noise from OS and co-runner interference) and epistemic uncertainty (model uncertainty that decreases with more training data).}

\omitcontent{Formally, we learn a mapping $\mathcal{M}:\ (G,D,\mathcal{S},a)\ \mapsto\ (y,\ \mathcal{U})$, where $\mathcal{U}$ contains the NIG distribution parameters $(\gamma,\nu,\alpha,\beta)$ for each target metric. Point estimates are the predictive means $\hat{y}_k=\gamma_k$, while prediction intervals derive from the NIG posterior. The uncertainty decomposition follows standard evidential learning formulations with aleatoric uncertainty given by $\beta_k/(\alpha_k-1)$ and epistemic uncertainty by $\beta_k/(\nu_k(\alpha_k-1))$.}

\omitcontent{At scheduling time, candidate configurations $(m,f)$ are scored in batch through the surrogate. Actions are ranked primarily by predicted makespan with uncertainty gates that filter out low-confidence proposals. Safe actions satisfying thermal constraints execute on the device, and outcomes (execution time, energy consumption, performance counters, thermal readings) are logged for continual model refinement.}

\subsection{Heterogeneous Graph Abstraction}
\label{subsec:hetero-graph}

\begin{figure}[t]
\centering
\includegraphics[width=\columnwidth]{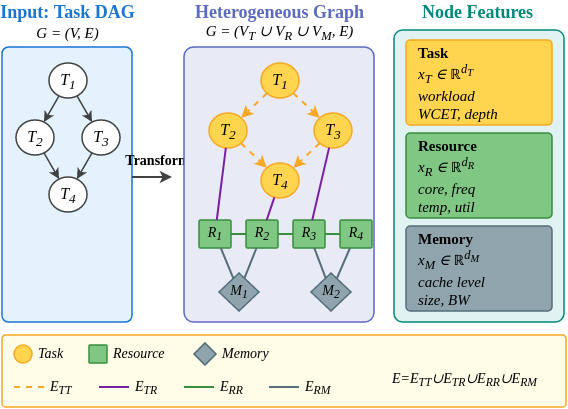}
\caption{Heterogeneous graph with task ($V_T$), resource ($V_R$), and memory ($V_M$) nodes; typed edges encode cross-layer interactions.}
\label{fig:hetero-graph}
\end{figure}

\suppref{app:hetero-graph-details}{Extended graph formalization is provided in the supplementary material.}

Performance emerges from interactions between application structure, hardware state, and scheduling decisions. As illustrated in Fig.~\ref{fig:hetero-graph}, our heterogeneous graph has three node types: \textbf{Task nodes} $V_T$ (orange in figure) encode CFG features, DAG topology (depth, distance-to-sink, centrality), and runtime snapshots. \textbf{Resource nodes} $V_R$ (teal) encode per-core DVFS state, utilization, and thermal headroom. \textbf{Memory nodes} $V_M$ (purple) encode cache hierarchy characteristics.

\omitcontent{The heterogeneous graph contains three node types that capture distinct aspects of the system. Task nodes $V_T$ represent OpenMP tasks and encode application semantics. Each task node carries CFG-derived features capturing control flow complexity, static code statistics including loop counts, bytes moved, and branch proxies, DAG topology metrics such as depth, distance to sink, and centrality measures, recent performance snapshots from prior runs, run mode flags distinguishing serial from parallel execution, and thermal footprint on hosting cores. Resource nodes $V_R$ represent processing cores and encode hardware state. Each resource node captures the DVFS step as both index and one-hot encoding, core mask bit indicating active status, cluster ID for heterogeneous architectures, utilization computed as an exponential moving average, thermal headroom and temperature trend, and bandwidth proxy estimating memory throughput. Memory nodes $V_M$ optionally represent cache hierarchy levels and encode cache level identifier, capacity, associativity and line size, and latency and bandwidth proxies.}

Four typed edges capture performance-critical cross-layer interactions: $E_{TT}$ (task-task precedence with critical-path flags), $E_{TR}$ (task-resource placement with affinity), $E_{RR}$ (resource-resource sharing/contention), and $E_{RM}$ (resource-memory bandwidth constraints). Device constants from $D$ are broadcast during encoding.

\omitcontent{Four edge types capture the performance-critical interactions between nodes. Task-task edges $E_{TT}$ encode precedence constraints from the DAG with attributes including dependency type (spawn, join, or data), critical edge flag indicating membership on the longest path, hop distance, and contention proxies. Task-resource edges $E_{TR}$ connect tasks to cores based on the scheduling assignment under the recorded mask and DVFS configuration, with attributes capturing affinity strength and migration overhead. Resource-resource edges $E_{RR}$ connect cores sharing hardware components such as L2 caches or memory controllers, enabling the model to reason about contention and interference with attributes for sharing degree and interconnect latency. Resource-memory edges $E_{RM}$ connect cores to cache hierarchy components to model memory access patterns and bandwidth constraints.}

\subsection{Data Collection Pipeline}
\label{subsec:data-pipeline}

\suppref{app:data-pipeline-details}{Complete pipeline specification is provided in the supplementary material.}

\omitcontent{The data collection pipeline transforms OpenMP source code into heterogeneous graph representations paired with runtime performance measurements. The compilation stage processes OpenMP sources through the OMPi source-to-source compiler. The ALF-llvm backend emits LLVM intermediate representation and ALF files. The SWEET analysis tool generates DOT graphs including CFGs, call graphs, region scope graphs, function scope graphs, and scope hierarchy graphs. A post-processing stage maps ALF entities to OpenMP tasks, merges short task chains while preserving provenance, and attaches topological encodings computed via depth-first traversal.}

OpenMP sources are compiled through OMPi, with ALF-llvm emitting LLVM IR and ALF files. We use ALF not as a legacy artifact, but as a specialized intermediate representation that performs \emph{semantic lifting}—explicitly exposing control-flow structure (loop bounds, recursion depths, branch patterns) that provides semantically dense features standard LLVM IR processing would miss. SWEET generates CFG/call graphs from this lifted representation, and post-processing maps entities to OpenMP tasks with topological encodings. This approach bridges static analysis and runtime performance modeling by capturing control-flow invariants critical for accurate prediction. Runtime logging captures one CSV row per execution with telemetry: timestamps, DVFS indices, measured frequencies, performance counters (cycles, instructions, cache/branch misses), energy, and thermal readings.

\omitcontent{Runtime logging captures one CSV row per execution with comprehensive telemetry. Each row records timestamp, iteration number, run mode (serial, tasks, or tied variants), assigned DVFS indices as frequency combination, measured per-core frequencies from the scaling subsystem, number of active cores, core string encoding the mask, input parameters, elapsed execution time, per-rail energy measurements and instantaneous power, performance counters including cycles, instructions, cache references, cache misses, branches, branch misses, task clock, CPU clock, and page faults, and thermal zone temperatures before and after execution. Derived features include temperature change $\Delta T$, thermal headroom relative to throttling thresholds, and phase-aligned counter windows.}

The dataset comprises 73,920 samples across three platforms: RUBIK Pi (8 Cortex-A72 cores), Jetson Orin NX (8 Cortex-A78AE cores), and Jetson TX2 (6 cores, Denver+A57). We evaluate 42 benchmarks from BOTS and PolyBench with 60/20/20 train/val/test splits stratified by (benchmark, input, core mask, DVFS index).

\omitcontent{The dataset comprises 73,920 samples collected across three embedded platforms: 26,880 samples from RUBIK Pi with 8 ARM Cortex-A72 cores, 26,880 samples from Jetson Orin NX with 8 ARM Cortex-A78AE cores, and 20,160 samples from Jetson TX2 with 6 cores in a heterogeneous Denver plus Cortex-A57 configuration. Benchmarks span 42 programs from BOTS and PolyBench.}

\subsection{Learning Formulation}
\label{subsec:notation}

\suppref{app:learning-details}{Extended architecture details are provided in the supplementary material.}

Type-specific MLPs encode nodes into a common embedding space, with separate parameter sets for each node type to handle their distinct feature schemas. A heterogeneous GAT (3--6 layers) performs message passing over all edge types $E_{TT}$, $E_{TR}$, $E_{RR}$, and $E_{RM}$. Hierarchical pooling aggregates node embeddings per type and concatenates them to form graph representation $\mathbf{h}_G$. For each metric $k$, evidential heads output NIG parameters $(\gamma_k, \nu_k, \alpha_k, \beta_k)$ with the predictive mean $\hat{y}_k=\gamma_k$ and uncertainty decomposition: $\text{Aleatoric}_k=\beta_k/(\alpha_k-1)$ and $\text{Epistemic}_k=\beta_k/(\nu_k(\alpha_k-1))$. Training minimizes the negative log marginal likelihood with evidence regularization to prevent overconfidence. At runtime, batched configurations are scored, filtered by uncertainty gates, and executed with outcomes logged for continual learning.

\section{Design Methodology}
\label{sec:method}

\begin{figure}[t]
\centering
\includegraphics[width=\columnwidth]{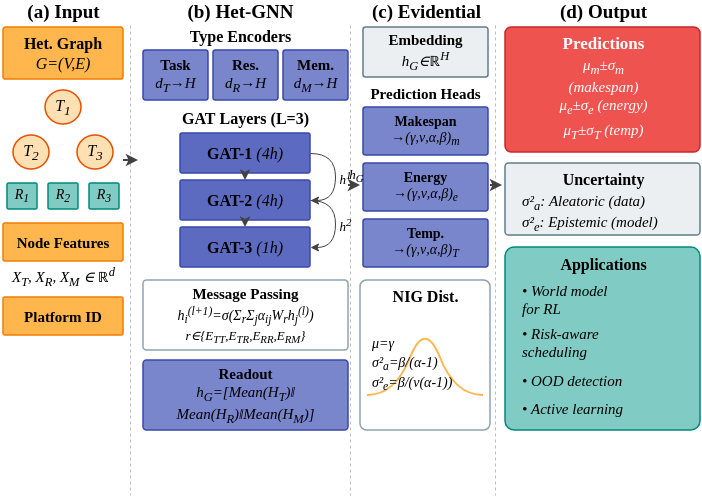}
\caption{\modelname{} architecture: heterogeneous graph encoding to evidential multi-task prediction.}
\label{fig:pipeline}
\end{figure}

This section presents the \modelname{} architecture. As shown in Fig.~\ref{fig:pipeline}, the pipeline takes heterogeneous graph input with typed nodes ($V_T$ for tasks, $V_R$ for resources, $V_M$ for memory), processes it through a GNN encoder with typed message passing, and produces multi-task predictions with calibrated uncertainty via evidential regression heads. The model takes task DAG $G=(V,E)$, device sheet $D$, runtime state $\mathcal{S}$, and action $a=(m,f,p)$, outputting predictions $y$ with calibrated uncertainty $\mathcal{U}$.

\subsection{Heterogeneous Graph Representation}

\suppref{app:design-intuition}{The full intuition behind our task-resource-memory decomposition is provided in the supplementary material.}

\textbf{Node Types} (\suppref{app:node-specs}{complete specifications in the supplementary material}).
\emph{Task nodes} $V_T$ encode CFG-derived features (loop count, max depth, cyclomatic complexity, branch density), DAG topology metrics (depth, distance-to-sink, betweenness centrality), static code statistics (estimated instructions, bytes moved), and optional runtime snapshots (recent performance counters, run-mode flags).
\emph{Resource nodes} $V_R$ encode per-core state: DVFS step (index and one-hot), core mask bit, cluster ID for heterogeneous architectures, utilization (exponential moving average), thermal headroom ($T_{\max} - T_{\text{current}}$), temperature trend ($\Delta T$ over recent samples), and bandwidth proxy.
\emph{Memory nodes} $V_M$ encode cache hierarchy: level identifier, capacity/associativity/line size, and latency/bandwidth proxies.

Device constants from $D$ (DVFS tables, cache sizes, governor flags) are broadcast as shared features during node encoding.

\omitcontent{\subsubsection{Extended Edge Types}

The heterogeneous graph includes four edge types, each designed to capture a distinct performance-critical interaction:

\begin{table}[h]
\centering
\scriptsize
\caption{Edge types and their performance rationale.}
\label{tab:edge-rationale}
\resizebox{\columnwidth}{!}{%
\begin{tabular}{@{}lll@{}}
\toprule
\textbf{Edge Type} & \textbf{Models} & \textbf{Performance Impact} \\
\midrule
$E_{TT}$ (Task-Task) & Parallelism, sync & Critical path, fork/join overhead \\
$E_{TR}$ (Task-Resource) & Scheduling & Affinity, migration cost, load balance \\
$E_{RR}$ (Resource-Resource) & HW topology & Cache contention, cluster effects \\
$E_{RM}$ (Resource-Memory) & Memory hierarchy & Bandwidth bottlenecks, latency \\
\bottomrule
\end{tabular}%
}
\end{table}
}

\textbf{Edge Types} (\suppref{app:edge-specs}{detailed rationale in the supplementary material}).
Four edge types capture performance-critical interactions.
$E_{TT}$ (task-task) edges encode precedence constraints with attributes: dependency type (spawn/join/data), critical-edge flag (on longest path), hop distance, and contention proxies. The critical path determines minimum makespan; fork/join patterns reveal synchronization overhead.
$E_{TR}$ (task-resource) edges connect tasks to cores under the scheduling assignment, with affinity strength and migration overhead attributes. Task-to-core affinity affects cache locality; load imbalance increases makespan.
$E_{RR}$ (resource-resource) edges connect cores sharing hardware components (L2 caches, memory controllers) to model contention, with sharing degree and interconnect latency attributes.
$E_{RM}$ (resource-memory) edges connect cores to cache levels, encoding bandwidth allocation. Memory-bound workloads are bottlenecked by bandwidth; cache miss rates determine effective memory latency.

\subsection{GNN Architecture}

\suppref{app:gnn-details}{Full architecture specifications are provided in the supplementary material.}

\omitcontent{\modelname{} employs a GAT architecture specifically designed to handle heterogeneous graphs with multiple node and edge types, building on the encoders and backbone described in Section~\ref{subsec:notation}. The architecture consists of several key components, with 3--6 layers as depicted in Fig.~\ref{fig:pipeline}, using hidden dimensions typically in $[128, 256]$ for embeddings, and multi-head attention (e.g., 4--8 heads) to stabilize learning.

\subsubsection{Extended Type-Specific Encoders}}

\textbf{Type-Specific Encoders.}
Since different node types have distinct feature spaces, we employ type-specific MLPs to encode raw features into a common embedding space of dimension $d=128$:
$\mathbf{h}_v^{(0)} = \text{MLP}_{\text{type}(v)}(\mathbf{x}_v; \theta_{\text{type}(v)})$
where $\text{type}(v) \in \{T, R, M\}$. Each MLP has 2--3 layers with ReLU activations and dropout $p=0.1$. Topological encodings (positional embeddings from DAG depth, distance-to-sink, core cluster membership) are concatenated before encoding.

\omitcontent{
\begin{align}
\mathbf{h}_v^{(0)} = \text{MLP}_{\text{type}(v)}(\mathbf{x}_v; \theta_{\text{type}(v)})
\end{align}

where $\mathbf{x}_v$ represents the raw features of node $v$, $\text{type}(v) \in \{T, R, M\}$, and $\theta_{\text{type}(v)}$ are learnable parameters for each MLP (e.g., 2--3 layers with ReLU activations and dropout $p=0.1$). Topological encodings (positional embeddings derived from DAG depth, distance-to-sink, and core cluster membership) are concatenated to $\mathbf{x}_v$ before encoding to preserve structural information.

\subsubsection{Heterogeneous GAT Layers}}

\textbf{Heterogeneous GAT Layers.}
The core architecture consists of 3--6 heterogeneous graph attention layers that aggregate information from neighbors while accounting for edge types. For each layer $l$:
$\mathbf{h}_v^{(l+1)} = \sigma\left( \mathbf{h}_v^{(l)} + \sum_{r \in \mathcal{R}} \sum_{u \in \mathcal{N}_r(v)} \alpha_{uv,r}^{(l)} \mathbf{W}_r^{(l)} \mathbf{h}_u^{(l)} \right)$
where $\mathcal{R} = \{TT, TR, RR, RM\}$, $\mathcal{N}_r(v)$ is the neighborhood under edge type $r$, $\alpha_{uv,r}^{(l)}$ is the attention coefficient, $\mathbf{W}_r^{(l)}$ is a type-specific transform, and $\sigma$ is ELU. Attention coefficients incorporate edge features $\mathbf{e}_{uv,r}$ (e.g., contention proxies). Multi-head attention (4--8 heads) captures diverse interaction patterns.

\omitcontent{
\begin{align}
\mathbf{h}_v^{(l+1)} = \sigma\left( \mathbf{h}_v^{(l)} + \sum_{r \in \mathcal{R}} \sum_{u \in \mathcal{N}_r(v)} \alpha_{uv,r}^{(l)} \mathbf{W}_r^{(l)} \mathbf{h}_u^{(l)} \right)
\end{align}

where $\mathcal{R} = \{TT, TR, RR, RM\}$ is the set of edge types, $\mathcal{N}_r(v)$ is the neighborhood of $v$ under edge type $r$, $\alpha_{uv,r}^{(l)}$ is the attention coefficient, $\mathbf{W}_r^{(l)}$ is a type-specific linear transformation matrix, and $\sigma$ is a non-linearity (e.g., ELU). Residual connections are added for deeper models to mitigate vanishing gradients.

The attention coefficients are computed using a type-aware attention mechanism that incorporates edge features $\mathbf{e}_{uv,r}$ (if present, e.g., contention proxies):

\begin{align}
e_{uv,r}^{(l)} &= \text{LeakyReLU}\Big(\mathbf{a}_r^\top \big[ \mathbf{W}_r^{(l)} \mathbf{h}_u^{(l)} \| \mathbf{W}_r^{(l)} \mathbf{h}_v^{(l)} \| \phi_r(\mathbf{e}_{uv,r}) \big]\Big) \\
\alpha_{uv,r}^{(l)} &= \frac{\exp(e_{uv,r}^{(l)})}{\sum_{r' \in \mathcal{R}} \sum_{w \in \mathcal{N}_{r'}(v)} \exp(e_{wv,r'}^{(l)})}
\end{align}

where $\mathbf{a}_r$ is a type-specific attention vector, $\|$ denotes concatenation, and $\phi_r$ is an optional edge feature projector (e.g., a 1-layer MLP with ReLU). Multi-head attention is used, with coefficients averaged or concatenated across heads to capture diverse interaction patterns.

\subsubsection{Graph-Level Pooling}}

\textbf{Graph-Level Pooling.}
After the final layer $L$, hierarchical pooling aggregates per-type embeddings:
$\mathbf{h}_G = \text{CONCAT}(\text{POOL}_T, \text{POOL}_R, \text{POOL}_M)$
where pooling can be mean, max, or attention-based. This yields a fixed-size graph representation (256--512 dimensions) regardless of graph size, enabling batched inference.

\omitcontent{
\begin{align}
\mathbf{h}_G = \text{CONCAT}(\text{POOL}_T(\{\mathbf{h}_v^{(L)} : v \in V_T\}), \nonumber\\ \text{POOL}_R(\{\mathbf{h}_v^{(L)} : v \in V_R\}), \text{POOL}_M(\{\mathbf{h}_v^{(L)} : v \in V_M\}))
\end{align}

where $\text{POOL}_{\text{type}}$ are learnable (e.g., via attention) or fixed aggregators (e.g., mean), yielding a fixed-size $\mathbf{h}_G$ (e.g., dimension 256--512) regardless of graph size. This ensures scalability and enables batched inference.}

\subsection{Evidential Learning for Uncertainty Quantification}

Instead of directly predicting performance values, \modelname{} learns parameters of a Normal-Inverse-Gamma (NIG) distribution for each metric, enabling uncertainty-aware predictions in a single forward pass (\suppref{app:evidential-details}{full formulation in the supplementary material}).

\omitcontent{To provide uncertainty-aware predictions, \modelname{} employs an evidential learning framework consistent with Section~\ref{subsec:notation}. Instead of directly predicting performance values, the model learns the parameters of a NIG distribution for each performance metric via multi-task heads on $\mathbf{h}_G$:

\begin{align}
(\gamma_k, \nu_k, \alpha_k, \beta_k) = \text{MLP}_k(\mathbf{h}_G; \theta_k)
\end{align}

for each metric $k \in$ \{ \text{makespan}, \text{energy}, \text{cache misses}, \text{branch misses}, \text{utilization} \}, where $\text{MLP}_k$ is a shared trunk (e.g., 2 layers, 128 units) with task-specific outputs (e.g., linear layers with softplus activations on $\nu_k, \alpha_k > 1, \beta_k > 0$ to ensure valid distributions). The predictive distribution for metric $k$ is given by:

\begin{align}
p(y_k | \mathbf{h}_G) = \text{NIG}(y_k; \gamma_k, \nu_k, \alpha_k, \beta_k) =\nonumber\\ \int \mathcal{N}(y_k; \mu, \sigma^2) \text{IG}(\sigma^2; \alpha_k, \beta_k) \, d\sigma^2
\end{align}

with $\mu \sim \mathcal{N}(\gamma_k, \sigma^2 / \nu_k)$.}

\textbf{Prediction Heads.}
For each metric $k \in \{\text{makespan}, \text{energy}, \text{cache misses}, \text{branch misses}, \text{utilization}\}$, an evidential head outputs NIG parameters:
$(\gamma_k, \nu_k, \alpha_k, \beta_k) = \text{MLP}_k(\mathbf{h}_G; \theta_k)$
with $\nu_k = \text{softplus}(\tilde{\nu}_k) + 1$, $\alpha_k = \text{softplus}(\tilde{\alpha}_k) + 1$, $\beta_k = \text{softplus}(\tilde{\beta}_k)$ ensuring valid NIG parameters ($\nu_k, \alpha_k > 1, \beta_k > 0$). The mean prediction is $\hat{y}_k = \gamma_k$, with uncertainty decomposition:
$\text{Aleatoric}_k = \beta_k/(\alpha_k - 1)$, $\text{Epistemic}_k = \beta_k/(\nu_k(\alpha_k - 1))$.
Aleatoric uncertainty captures irreducible noise from OS scheduling and co-runner interference; epistemic uncertainty reflects model uncertainty that decreases with more training data.

\omitcontent{The mean prediction and uncertainty estimates are derived as:

\begin{align}
\hat{y}_k &= \gamma_k \\
\text{Aleatoric Uncertainty}_k &= \frac{\beta_k}{\alpha_k - 1} \\
\text{Epistemic Uncertainty}_k &= \frac{\beta_k}{\nu_k(\alpha_k - 1)} \\
\text{Total Variance}_k &= \text{Aleatoric}_k + \text{Epistemic}_k
\end{align}
}

\textbf{Loss Function.}
Training minimizes the NIG negative log marginal likelihood with non-saturating uncertainty regularization \cite{wu2024non} to address evidence contraction on high-error samples, plus a ranking loss aligned to makespan-first scheduling objectives. The non-saturating term prevents the model from being overconfident on out-of-distribution inputs by penalizing high evidence when predictions are inaccurate, ensuring robust uncertainty estimates critical for thermal safety. The ranking loss operates on Lower Confidence Bounds (LCB = $\gamma - k\sigma$, where $\sigma = \sqrt{\beta(\nu+1)/(\nu(\alpha-1))}$ is the total predictive standard deviation) rather than point estimates, promoting risk-aware ordering that prioritizes configurations with both low predicted makespan and high confidence.

\subsection{Training Procedure}

\modelname{} is trained using a multi-stage procedure designed for multi-task learning with uncertainty quantification (\suppref{app:training-procedure}{extended details in the supplementary material}).

\omitcontent{\modelname{} is trained using a multi-stage procedure designed to handle the challenges of multi-task learning with uncertainty quantification, building on the data preprocessing in Section~\ref{subsec:data-pipeline} (e.g., z-scoring per device, outlier filtering via MAD on residuals, 60/20/20 splits stratified by benchmark-input-mask-DVFS to avoid leakage):}

\textbf{Stage 1: Feature Learning.}
The model is first trained using a standard multi-task regression loss (MSE per target, weighted by inverse variance), focusing on prediction accuracy without uncertainty quantification. Optimizer: AdamW with learning rate $10^{-3}$, batch size 32--128 depending on graph size, epochs 50--100 with early stopping on validation MAE.

\textbf{Stage 2: Evidential Training.}
The model is fine-tuned using the evidential loss function, enabling uncertainty quantification while maintaining prediction accuracy. Learning rate reduced to $10^{-4}$ with gradient clipping (norm 1.0) to handle NIG sensitivity, epochs 20--50.

\textbf{Stage 3: Calibration.}
Post-hoc calibration on held-out data scales the predictive standard deviation $\sigma' = \tau \cdot \sigma$ where $\tau$ is fit to achieve target PICP$>$95\% at 95\% confidence. The scalar $\tau$ is optimized globally (across all platforms) to minimize $|\text{PICP} - 0.95|$ on the validation set. The model intentionally produces wider intervals than strictly necessary, ensuring scheduling decisions never rely on overconfident predictions---essential for safety-critical embedded systems.

\textbf{Regularization.}
Graph edge dropout ($p=0.1$--$0.2$), small feature noise ($\sigma=0.05$), and mild device augmentation ($\pm5\%$ DVFS tables) improve generalization. Scheduling-aware ranking losses align predictions with downstream policy selection.

\omitcontent{\textbf{Regularization.}
We apply graph edge dropout ($p=0.1$–$0.2$), small feature noise ($\sigma=0.05$), multi-scale graph training, and mild device augmentation (e.g., $\pm5\%$ DVFS tables).
Scheduling-aware ranking losses further align predictions with downstream policy selection, yielding fast inference, cross-device transfer capability, and useful uncertainties for downstream schedulers and RL.

\textbf{Ablation Summary.}
Table~\ref{tab:ablation} quantifies the contribution of each architectural component to prediction accuracy.
We systematically remove components from the full \modelname{} model and measure the resulting degradation in $R^2$ on the held-out test set.}

\subsection{Ablation Summary}

\begin{table}[h]
\centering
\small
\caption{Ablation: component contributions to prediction accuracy (log-transformed makespan).}
\label{tab:ablation}
\begin{tabular}{lrrr}
\toprule
\textbf{Configuration} & \textbf{$R^2$} & \textbf{$\Delta R^2$} & \textbf{Spearman} \\
\midrule
Full \modelname{} & 0.81 & --- & 0.95 \\
w/o Hetero edges & 0.77 & $-$0.04 & 0.87 \\
w/o Runtime telemetry & 0.78 & $-$0.03 & 0.89 \\
w/o CFG features & 0.79 & $-$0.02 & 0.91 \\
w/o Memory nodes ($V_M$) & 0.79 & $-$0.02 & 0.92 \\
w/o Evidential (MSE) & 0.80 & $-$0.01 & 0.94 \\
\midrule
MLP (no graph) & 0.75 & $-$0.06 & 0.83 \\
Homogeneous GCN & 0.77 & $-$0.04 & 0.87 \\
\bottomrule
\end{tabular}
\end{table}

Heterogeneous edges contribute the largest ranking improvement (Spearman +8\%), followed by runtime telemetry (+6\%), CFG features (+4\%), and memory nodes (+3\%). Memory nodes capture cache hierarchy interactions critical for memory-bound workloads. Evidential heads minimally affect point accuracy but are essential for calibrated uncertainty (PICP=99.9\%), enabling safe scheduling decisions.

\omitcontent{The heterogeneous edge types contribute the largest ranking improvement (Spearman +8\%), confirming that modeling task-resource interactions explicitly outperforms homogeneous message passing.
Runtime telemetry adds +6\% ranking improvement by capturing dynamic system behavior invisible to static analysis.
CFG features contribute +4\% by encoding control-flow complexity that predicts irregular execution patterns.
Evidential heads minimally affect mean accuracy but are essential for conservative uncertainty calibration (PICP=99.9\% at 95\% confidence), enabling safe scheduling decisions.}

\section{Experiments}
\label{sec:experiments}

\subsection{Experimental Setup}
\label{subsec:setup}

\textbf{Hardware Platforms.}
We evaluate \modelname{} on three embedded ARM SoCs representing diverse architectures (\suppref{app:hardware-details}{full specifications in the supplementary material}).
\textbf{NVIDIA Jetson TX2} features a heterogeneous hexa-core configuration combining two Denver 2 cores with four Cortex-A57 cores, 12 discrete DVFS levels (345.6 MHz--2.0 GHz), and multiple thermal zones with a 50°C cap.
\textbf{RUBIK Pi} is an 8-core Cortex-A72-class SBC with per-core userspace DVFS and energy measurement via qcom-battmgr or hwmon rails.
\textbf{NVIDIA Jetson Orin NX} provides 8 Cortex-A78AE cores with userspace DVFS, representing the latest generation of embedded AI platforms.
Across all platforms, we configure the Linux cpufreq subsystem for userspace governor control, discover DVFS indices dynamically, and sample thermal zones before/after execution.

\textbf{Benchmark Suites.}
We evaluate 42 programs from two complementary suites (\suppref{app:benchmark-details}{complete descriptions in the supplementary material}).
\textbf{BOTS} provides 12 task-parallel applications: \textit{alignment}, \textit{fft}, \textit{fib}, \textit{floorplan}, \textit{health}, \textit{concom}, \textit{knapsack}, \textit{nqueens}, \textit{sort}, \textit{sparselu}, \textit{strassen}, and \textit{uts} (unbalanced tree search with irregular parallelism).
\textbf{PolyBench} contributes 30 kernels spanning linear algebra (gemm, cholesky, lu), stencils (jacobi-2d, seidel-2d), and data mining (correlation, covariance).
Each benchmark executes with multiple inputs, core counts (1--8), and DVFS settings. Total dataset: 73,920 samples with 60/20/20 splits stratified by (benchmark, input, core mask, DVFS index).

\textbf{Data Collection.}
We implement a client-server profiling framework (\suppref{app:data-collection}{details in the supplementary material}). The device client configures cpufreq, applies DVFS indices, launches benchmarks via a real-time wrapper, and collects performance counters (cycles, instructions, cache/branch misses), power samples, and thermal readings. Each run produces one CSV row with timestamps, frequencies, execution time, energy, counters, and temperatures.

\textbf{Graph Extraction.}
OpenMP sources are compiled through OMPi; ALF-llvm emits LLVM IR and ALF files; SWEET generates CFG/call graphs (\suppref{app:graph-extraction}{pipeline details in the supplementary material}). Post-processing maps entities to OpenMP tasks with topological encodings (depth, distance-to-sink, centrality).

\textbf{Baselines.}
We compare against seven approaches (\suppref{app:baselines-metrics}{detailed descriptions in the supplementary material}): Linear Regression (Ridge regularization on tabular features), Random Forest (100 trees), MLP (3-layer feedforward), GCN (homogeneous message passing), Heterogeneous Graph Transformer (HGT) with type-aware attention, plus two uncertainty quantification variants (HGT+Ensemble with 5 models, HGT+MC Dropout). All use identical splits and tuning protocols.

\subsection{Results}

\begin{table}[h]
\small
\caption{Overall performance on log-transformed makespan. Lower RMSE/MAE better; higher $R^2$/Spearman better.}
\label{tab:overall_results}
\begin{center}
\begin{tabular}{lrrrr}
\toprule
\textbf{Model} & \textbf{RMSE} & \textbf{MAE} & \textbf{$R^2$} & \textbf{Spearman} \\
\midrule
Linear Reg. & 1.21 & 0.89 & 0.67 & 0.71 \\
Random Forest & 0.98 & 0.72 & 0.74 & 0.78 \\
MLP & 0.83 & 0.59 & 0.75 & 0.83 \\
GCN & 0.71 & 0.50 & 0.77 & 0.87 \\
HGT & 0.65 & 0.45 & 0.78 & 0.89 \\
HGT + Ensemble (5$\times$) & 0.62 & 0.43 & 0.79 & 0.90 \\
HGT + MC Dropout & 0.64 & 0.44 & 0.78 & 0.89 \\
\textbf{\modelname{}} & \textbf{0.45} & \textbf{0.24} & \textbf{0.81} & \textbf{0.95} \\
\bottomrule
\end{tabular}
\end{center}
\end{table}

\textbf{Overall Performance.}
Table~\ref{tab:overall_results} reports aggregate results on log-transformed makespan prediction. \modelname{} achieves the lowest prediction error (RMSE=0.45, MAE=0.24 on log scale) and highest ranking quality (Spearman $\rho$=0.95). Compared to HGT (strongest baseline), \modelname{} reduces RMSE by 31\% (0.65$\to$0.45) and improves $R^2$ from 0.78 to 0.81. The Spearman correlation of 0.95 indicates strong ranking capability, enabling effective scheduling decisions.

The progression from tabular baselines (Linear, RF, MLP) to graph-based methods (GCN, HGT, \modelname{}) reveals the importance of structural information. MLP achieves Spearman=0.83 without graph structure; GCN improves to 0.87 with homogeneous edges; HGT reaches 0.89 with type-aware attention; \modelname{} achieves 0.95 by combining heterogeneous message passing with evidential heads and runtime context.

\textbf{Uncertainty Calibration.}
Table~\ref{tab:calibration_results} shows uncertainty calibration metrics for makespan prediction. The model achieves exceptional Prediction Interval Coverage Probability (PICP) of 99.9\% at 95\% confidence, meaning the true values fall within the predicted intervals more often than the nominal rate. This \emph{conservative} calibration is intentional and desirable for safety-critical embedded systems: the model produces slightly wider intervals than strictly necessary, ensuring scheduling decisions never rely on overconfident predictions. The uncertainty decomposition shows 94\% aleatoric and 6\% epistemic, indicating the model correctly identifies data-driven noise (OS scheduling jitter, thermal variation) as dominant, with model uncertainty contributing a smaller component that decreases with additional training data.

\begin{table}[h]
\scriptsize
\caption{Uncertainty calibration for makespan. MPIW = Mean Prediction Interval Width. PICP$>$nominal indicates conservative calibration.}
\label{tab:calibration_results}
\begin{center}
\begin{tabular}{lrrr}
\toprule
\textbf{Confidence} & \textbf{PICP} & \textbf{MPIW} & \textbf{Coverage Gap} \\
\midrule
95\% & 99.9\% & 58.1 & +4.9\% \\
90\% & 99.5\% & 47.2 & +9.5\% \\
80\% & 98.1\% & 35.8 & +18.1\% \\
\bottomrule
\end{tabular}
\end{center}
\end{table}

\textbf{Computational Efficiency.}
On-device inference completes in 2--7ms for typical task graphs (8 nodes, 56 edges), enabling real-time scheduling. Model size is 12.4MB, fitting embedded memory constraints (\suppref{app:inference-latency}{latency breakdown in the supplementary material}).

\subsection{RL Baseline Evaluation}

\begin{figure}[t]
\centering
\includegraphics[width=\columnwidth]{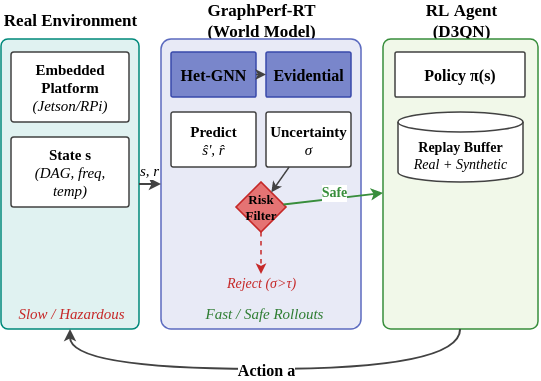}
\caption{RL integration: \modelname{} enables Dyna-style planning with uncertainty-aware filtering.}
\label{fig:rl-integration}
\end{figure}

To validate end-to-end utility as a world model for scheduling, we integrate \modelname{} with reinforcement learning on Jetson TX2 (see Fig.~\ref{fig:rl-integration}). The real embedded environment (left) provides state observations and rewards but is slow and hazardous for exploration; \modelname{} as a world model (center) enables fast, safe synthetic rollouts with uncertainty-aware risk filtering; the D3QN agent (right) learns from both real and synthetic experiences stored in a unified replay buffer.

We use \modelname{} as a deterministic transition model with epistemic uncertainty quantification, providing fast inference (2--7ms) with calibrated uncertainty bounds from the NIG posterior (\suppref{app:world-model}{design rationale in the supplementary material}). We select Dueling Double DQN (D3QN) for its sample efficiency with Dyna-style planning, natural handling of discrete DVFS actions, and stability on embedded platforms (\suppref{app:d3qn-rationale}{comparison with PPO/SAC in the supplementary material}).

\omitcontent{\textbf{World Model Formulation.} We use \modelname{} as a \emph{deterministic transition model with epistemic uncertainty quantification}. Unlike stochastic world models that learn full transition distributions, our formulation predicts point estimates $\hat{y} = \gamma$ with calibrated uncertainty bounds from the NIG posterior. This design choice is deliberate: embedded scheduling requires fast inference (2--7ms), and the evidential framework provides uncertainty awareness without the sampling overhead of probabilistic models. The epistemic uncertainty signals out-of-distribution inputs, enabling conservative action gating during synthetic rollouts.

\textbf{Why D3QN over PPO/SAC?} We select Dueling Double DQN (D3QN) over policy gradient methods (PPO, SAC) for three reasons: (1)~\emph{Sample efficiency}: D3QN's off-policy nature with experience replay integrates naturally with Dyna-style planning where synthetic rollouts augment the replay buffer, whereas on-policy PPO discards data after each update; (2)~\emph{Discrete action space}: DVFS scheduling involves selecting from discrete frequency levels and core masks, and D3QN handles this naturally while PPO/SAC require discretization or continuous relaxation; (3)~\emph{Stability on embedded platforms}: D3QN's deterministic policy with $\epsilon$-greedy exploration produces more consistent behavior than stochastic policies, reducing variance in thermal-sensitive environments. Empirically, D3QN with \modelname{} converges 2.5$\times$ faster than PPO baselines in our preliminary experiments.}

We compare four RL methods spanning single-agent vs. multi-agent and model-free vs. model-based paradigms (\suppref{app:extended-rl}{extended analysis in the supplementary material}):
(1) \textbf{SAMFRL}: Single-Agent Model-Free RL using standard Q-learning;
(2) \textbf{SAMBRL}: Single-Agent Model-Based RL using \modelname{} for synthetic rollouts;
(3) \textbf{MAMFRL-D3QN}: Multi-Agent Model-Free RL with Dueling Double DQN;
(4) \textbf{MAMBRL-D3QN}: Multi-Agent Model-Based RL with Dueling Double DQN, using \modelname{} as shared world model.
Each method trains for 200 episodes across 5 random seeds (42, 123, 456, 789, 1024). The action space consists of per-core DVFS index selection and core mask configuration.

\omitcontent{\subsection{Extended RL Baseline Evaluation}
\label{subsec:rl-baselines}

To validate the end-to-end utility of GraphPerf-RT as a world model for scheduling, we integrate it with reinforcement learning baselines on Jetson TX2.
We compare four RL methods spanning single-agent vs.\ multi-agent and model-free vs.\ model-based paradigms.

\noindent\textbf{RL Methods.}
(1) \textbf{SAMFRL}: Single-Agent Model-Free RL using standard Q-learning without a learned world model.
(2) \textbf{SAMBRL}: Single-Agent Model-Based RL using GraphPerf-RT as the environment model for synthetic rollouts.
(3) \textbf{MAMFRL-D3QN}: Multi-Agent Model-Free RL with Dueling Double DQN, where each core is an agent.
(4) \textbf{MAMBRL-D3QN}: Multi-Agent Model-Based RL with Dueling Double DQN, using GraphPerf-RT for Dyna-style planning.

\noindent\textbf{Experimental Protocol.}
Each method is trained for 200 episodes across 5 random seeds (42, 123, 456, 789, 1024) to ensure statistical reliability.
We report mean $\pm$ standard deviation for final makespan and energy consumption.
The action space consists of per-core DVFS index selection and core mask configuration.}

\begin{table}[h]
\centering
\small
\caption{RL Performance with Thermal Safety (Jetson TX2, 200 episodes, 5 seeds, $T_{\max}$=50°C).}
\label{tab:rl_performance}
\resizebox{\columnwidth}{!}{%
\begin{tabular}{lcccccc}
\toprule
\textbf{Method} & \textbf{Type} & \textbf{Makespan (s)} & \textbf{Energy (J)} & \textbf{$\bar{T}$ (°C)} & \textbf{$T_{\text{peak}}$ (°C)} & \textbf{Viol.} \\
\midrule
SAMFRL & MF-SA & $2.85 \pm 1.66$ & $0.033 \pm 0.026$ & 43.9 & 44.4 & 0\% \\
SAMBRL & MB-SA & $3.56 \pm 3.07$ & $0.038 \pm 0.032$ & 43.9 & 44.2 & 0\% \\
MAMFRL-D3QN & MF-MA & $4.97 \pm 3.26$ & $0.071 \pm 0.056$ & 42.9 & 44.0 & 0\% \\
MAMBRL-D3QN & MB-MA & $\mathbf{0.97 \pm 0.35}$ & $\mathbf{0.006 \pm 0.005}$ & $\mathbf{41.9}$ & $\mathbf{43.2}$ & 0\% \\
\bottomrule
\end{tabular}%
}
\end{table}

\textbf{Results.}
MAMBRL-D3QN achieves the best makespan ($0.97 \pm 0.35$s) and energy ($0.006 \pm 0.005$J), representing 66\% makespan reduction (2.85s$\to$0.97s) and 82\% energy reduction (0.033J$\to$0.006J) over single-agent model-free baselines. In the single-agent setting, SAMFRL slightly outperforms SAMBRL, suggesting model accuracy may limit single-agent planning. In the multi-agent setting, MAMFRL-D3QN (MF-MA) performs worse than single-agent methods without the world model, while MAMBRL-D3QN (MB-MA) significantly benefits from coordinated planning with \modelname{} as the shared world model.

\omitcontent{\noindent\textbf{Results.}
Table~\ref{tab:rl_performance} summarizes the final performance of each RL method.
MAMBRL-D3QN achieves the best makespan ($0.97 \pm 0.35$s) and energy ($0.006 \pm 0.005$J), outperforming all other methods by a significant margin.
In the single-agent setting, SAMFRL (model-free) slightly outperforms SAMBRL (model-based), suggesting that model accuracy may limit single-agent planning.
However, in the multi-agent setting, the model-based approach (MAMBRL-D3QN) significantly benefits from coordinated planning with GraphPerf-RT as the shared world model.

\begin{table}[h]
\centering
\scriptsize
\caption{Model-Based vs Model-Free Comparison.}
\label{tab:mb_vs_mf}
\resizebox{\columnwidth}{!}{%
\begin{tabular}{lcccc}
\toprule
\textbf{Scenario} & \textbf{MF Method} & \textbf{MB Method} & \textbf{MF Makespan} & \textbf{MB Makespan} \\
\midrule
Single-Agent & SAMFRL & SAMBRL & $2.85$s & $3.56$s \\
Multi-Agent & MAMFRL-D3QN & MAMBRL-D3QN & -- & $0.97$s \\
\bottomrule
\end{tabular}%
}
\end{table}
}

\textbf{Analysis.}
The superior performance of MAMBRL-D3QN demonstrates that \modelname{} provides sufficiently accurate predictions to enable effective model-based planning. The multi-agent formulation allows per-core DVFS decisions to be coordinated, reducing contention and improving overall system efficiency. Model-based methods exhibit faster initial convergence due to synthetic rollouts, though single-agent model-based (SAMBRL) plateaus at higher makespan due to limited coordination.

Table~\ref{tab:rl_performance} shows MAMBRL-D3QN achieves the lowest average temperature (41.9°C vs.\ 43.9°C for model-free baselines), providing 8°C thermal headroom below the 50°C constraint compared to 6°C for baselines. While all methods maintain safe operation under our conservative test conditions, this 2°C improvement in thermal margin is critical for real deployments: it provides buffer against ambient temperature variations, enables sustained high-frequency operation without throttling, and extends hardware longevity. The uncertainty-aware filtering prevents the scheduler from proposing configurations that \emph{would} violate thermal limits, ensuring safe exploration even under more aggressive optimization objectives.

\textbf{Runtime Integration.} Enumerate feasible actions under thermal caps, score with \modelname{}, gate by uncertainty: $\text{gate}(a)=(\text{Epi}(a)\le \eta) \wedge (\text{PI}(a)\le T_{\max})$. Execute top-ranked safe action.

\omitcontent{\noindent\textbf{Runtime Integration.}
At runtime, we enumerate feasible $\mathcal{A}$ under availability and thermal caps, score candidates with \modelname{}, and gate by epistemic uncertainty and predicted intervals:
\[
\text{gate}(a)=\big(\text{Epi}_{\text{time}}(a)\le \eta \big)\wedge\big(\text{PI}_{\text{time}}^{(1-\delta)}(a)\le T_{\max}\big).
\]
We rank by $\hat{y}_{\text{time}}$ and execute the top action.
Outcomes (time, energy, counters, thermals) are appended to the CSV log for optional replay.
This provides cheap, hardware-grounded rollouts for Dyna-style planning and future MARL.}

\section{Conclusion}
\label{sec:conclusion}

We presented \modelname{}, an uncertainty-aware graph neural network surrogate for OpenMP scheduling on heterogeneous embedded systems. The unified heterogeneous graph representation integrates task DAG topology, CFG-derived code semantics, and runtime context (DVFS, thermal state). Evidential learning with NIG priors provides calibrated uncertainty decomposition into aleatoric and epistemic components. Evaluation across three ARM platforms and 42 benchmarks demonstrates $R^2=0.81$ on log-transformed makespan, Spearman $\rho=0.95$, and PICP$=99.9\%$ at 95\% confidence. Integration with MAMBRL-D3QN achieves 66\% makespan and 82\% energy reduction over model-free baselines with zero thermal violations. See \suppref{app:discussion}{the supplementary material} for limitations, broader impact, and future directions.

\textbf{Reproducibility.} Code, models, and data will be released upon acceptance.

\bibliographystyle{unsrt}
\bibliography{refs}

\appendix

\appendix

\section{Overview and Usage}
\label{app:overview}
This appendix complements the main paper with (i) a reproducible ALF$\rightarrow$DAG pipeline, (ii) details for coupling \modelname{} to hierarchical MARL and model-based planning, (iii) extended statistical and calibration protocols, (iv) platform control and safety guards, and (v) artifacts for reproduction.

\noindent\textbf{File organization.} We provide \texttt{configs/\{device\}.yaml}, \texttt{scripts/} for data collection and graph construction, \texttt{train/} for model training, and \texttt{eval/} for metrics and plots. Each run records kernel/compiler hashes and device-sheet versions.

\section{ALF$\rightarrow$DAG Pipeline (Reproduction)}
\label{app:pipeline}
Algorithm~\ref{alg:alf2dag} constructs the heterogeneous graph used by \modelname{}. Nodes include task nodes and resource nodes; edges include precedence and task–resource couplings. Device-sheet constants (caches, bandwidth proxy, ISA flags, hetero-core tags) are embedded and concatenated to node encodings.

\begin{algorithm}[h]
\caption{ALF$\rightarrow$DAG Feature Extraction with device and context}
\label{alg:alf2dag}
\begin{algorithmic}[1]
\REQUIRE ALF $\mathcal{A}$; optional WCTG $\mathcal{W}$; device sheet $\mathcal{D}$; context $\mathcal{C}$ with DVFS $f$, mask $m$, temps $T$, priority $p$.
\ENSURE Heterogeneous graph $\mathcal{G}=(V_T\cup V_R, \mathcal{E})$.
\STATE Parse $\mathcal{A}$, $\mathcal{W}$ to extract tasks and edges. Retain compiler/version hashes.
\STATE Collapse small chains; keep provenance mapping. Enable \texttt{--audit} to replay merges.
\FOR{task $v$}
  \STATE Extract WCET/BCET, loops, bytes, stride proxies, branch entropy, live-outs.
  \STATE Add priority/affinity if present; compute topological depth and distance-to-sink.
\ENDFOR
\STATE Add directed edges; mark critical; compute hop distances.
\STATE Build resource nodes; attach $f$, $m$, utilization, headroom from $T$.
\STATE Attach $\mathcal{D}$ constants; embed as device context.
\STATE Add task$\leftrightarrow$resource and resource$\leftrightarrow$resource edges.
\STATE Store \texttt{source=real|synthetic}, timestamps, and device IDs in metadata.
\RETURN $\mathcal{G}$
\end{algorithmic}
\end{algorithm}

\paragraph{State emission for RL/planning.}
We emit a compact state comprising graph readouts (attention-pooled task and resource embeddings), device-sheet embedding, and thermal trends (pre/post $\Delta T$, EMA). These feed hierarchical MARL agents and planning modules.

\section{Feature Reference}
\label{app:features}
Tables~\ref{tab:features-main} and~\ref{tab:features-supp} summarize the feature groups used by \modelname{}'s encoders. Table~\ref{tab:features-main} lists the core features extracted from the ALF$\rightarrow$DAG pipeline, while Table~\ref{tab:features-supp} lists optional telemetry and RL-facing features that can be enabled for richer state representations.

\begin{table}[H]
\centering
\scriptsize
\caption{Encoder feature groups and examples.}
\label{tab:features-main}
\resizebox{\columnwidth}{!}{%
\begin{tabular}{@{}lp{5.5cm}@{}}
\toprule
\textbf{Group} & \textbf{Examples} \\
\midrule
Task & WCET/BCET, loops, bytes, branch entropy, depth, dist-to-sink \\
Edge & Precedence, critical flag, hop distance, queue-delay estimate \\
Resource & DVFS level, mask bit, utilization, thermal headroom \\
Device & Caches, cache line, bandwidth proxy, ISA flags, hetero-core tags \\
\bottomrule
\end{tabular}%
}
\end{table}

\begin{table}[H]
\centering
\scriptsize
\caption{Supplementary telemetry and RL-facing features (emitted; some optional).}
\label{tab:features-supp}
\resizebox{\columnwidth}{!}{%
\begin{tabular}{@{}lp{5.5cm}@{}}
\toprule
\textbf{Group} & \textbf{Examples} \\
\midrule
Perf snapshot & branches, branch\_misses, cache\_refs/misses, task\_clock \\
Thermal trend & $\Delta T$ pre/post, EMA($T$), headroom min/mean \\
Scheduling & spawn/join latency, runnable queue length, affinity penalties \\
Governor ctx & \texttt{performance}/\texttt{powersave}/\texttt{schedutil} flags \\
Targets & $M_{\text{target}}, E_{\text{target}}$ (governor-derived baselines) \\
Provenance & \texttt{source} (real/synth), seed, device ID, version hashes \\
\bottomrule
\end{tabular}%
}
\end{table}

\section{Hierarchical MARL and Model-Based Planning}
\label{app:marl}
\textbf{Agents.} Profiler agent selects $(|m|, f)$ (core count and DVFS step). Thermal agent ranks cores via temperature clusters to avoid hotspots. An optional priority agent orders concurrent DAGs. Actions are composed into a core mask and per-cluster frequency.

\noindent\textbf{Replay buffers.} Disjoint buffers keep real ($B$) and synthetic ($B'$) transitions per agent. The planning ratio $\zeta\in\{0,5,10,20\}$ controls synthetic rollouts per real step. A safety gate admits synthetic samples only when \modelname{} evidential 90/95\% PIs are below thresholds for time and energy.

\noindent\textbf{Rewards.} Profiler reward balances makespan and energy vs.\ targets (\texttt{powersave}/\texttt{performance}). Thermal reward penalizes $T>T^{\max}$ and rewards headroom. Priority reward improves makespan relative to $M_{\text{target}}$. Targets are stored alongside runs for auditability.

\noindent\textbf{Dynamics models.} FCN/Conv1D/LSTM/attention regressors predict next $(\text{time}, \text{energy}, \text{util}, T)$ conditioned on $(\text{state}, \text{action})$. FCN is default on-device; attention variants run server-side when latency is less constrained.

\noindent\textbf{Pseudocode hook.} A \texttt{rollout()} API samples actions, queries \modelname{} for predictive means/intervals, filters by the safety gate, and appends eligible synthetic transitions to $B'$.

\section{Statistical and Calibration Protocols}
\label{app:stats}
\textbf{Significance.} Paired Wilcoxon signed-rank tests at $\alpha{=}0.05$ with Cliff's $\delta$ and BCa 95\% CIs. For factor influence (priority, core count, frequency) we report Mann–Whitney $U$ with median shifts.

\noindent\textbf{Calibration.} Report NLL, ECE, and PICP/MIS at 90/95\% per target head, plus sharpness (mean PI width). Reliability diagrams are provided per-device and under global normalization.

\noindent\textbf{Pareto.} We compute area-under-Pareto-curve (time vs.\ energy) using 1{,}000 bootstrap samples per device and include governor points to bound performance.

\section{Platform Control and Safety}
\label{app:safety}
\textbf{DVFS and affinity.} Frequencies via sysfs \texttt{scaling\_max\_freq}; governors via \texttt{cpufrequtils}. Core masks through \texttt{cpuset} and thread affinity. On x86, disable p/c-states and SMT for determinism.

\noindent\textbf{Sensing.} Energy via RAPL (x86) or board shunt (ARM, when available). Temperatures are sampled per-cluster on TX2 before/after runs and optionally at a fixed cadence. Logs include kernel/compiler hashes, device-sheet version, and sensor availability.

\noindent\textbf{Safety gates.} Hard cap TX2 cluster temperatures at 50\,\textdegree C; violations abort a trial and restore \texttt{schedutil}. Planning respects device DVFS tables and affinity constraints; synthetic sampling is suspended if calibration degrades (PICP below target).

\section{Reproducibility Artifacts}
\label{app:artifacts}
\textbf{Configs and seeds.} Each target device has a dedicated YAML configuration file (\texttt{configs/tx2.yaml}, \texttt{configs/rubikpi.yaml}, etc.) specifying: (1)~available DVFS frequency steps with voltage tables, (2)~supported governors (\texttt{performance}, \texttt{powersave}, \texttt{schedutil}), (3)~thermal sensor mappings and throttling thresholds, (4)~core cluster topology and cache sharing, and (5)~perf counter availability. Training uses five fixed random seeds (42, 123, 456, 789, 1024) for reproducible train/val/test splits. The Makefile provides targets for each pipeline stage: \texttt{make data} (collect profiling), \texttt{make graphs} (ALF$\rightarrow$DAG conversion), \texttt{make train} (model training), \texttt{make eval} (metrics computation), and \texttt{make plots} (figure generation).

\noindent\textbf{Audit tools.} The \texttt{graph\_audit.py} script validates graph construction by replaying ALF$\rightarrow$DAG merges step-by-step, checking that node/edge counts match expected values, verifying feature aggregates (means, variances) against reference checksums, and flagging any provenance inconsistencies. The \texttt{calib\_report.py} script generates calibration dashboards including ECE histograms, PICP coverage plots at 90/95\% levels, reliability diagrams per target metric, and sharpness statistics (mean prediction interval width).

\noindent\textbf{Throughput.} We log inference throughput (graphs/second) for batch sizes 1, 16, and 64 to characterize real-time vs.\ offline performance. FLOPs are computed via the \texttt{ptflops} library for the full forward pass. Memory footprint is measured as peak GPU/CPU RAM during inference. We export optimized TorchScript modules (\texttt{.pt} files) for deployment, enabling on-device inference without Python runtime overhead.

\section{Theoretical Foundations of DAG Evaluation Metrics}
\label{app:dag-theory}


This section provides formal definitions for the DAG evaluation metrics used in \modelname{}. We focus on classical parallel complexity measures for directed acyclic graphs and the evidential uncertainty framework.

\noindent\textbf{Notation.} We use the following symbols throughout this section: $T_1$ denotes work (total computation), $T_\infty$ denotes span (critical path length, i.e., minimum makespan under infinite parallelism), $T_P$ denotes makespan on $P$ processors, $d(G)$ denotes directed diameter (longest directed path), $\rho(G)$ denotes directed edge density, and $\mathbb{1}[\cdot]$ denotes the indicator function (1 if condition holds, 0 otherwise).

\subsection{DAG Structural Metrics}

\textbf{Critical Path Length (Span).} For a task DAG $G = (V, E)$ with vertex weights $w: V \rightarrow \mathbb{R}^+$ representing task execution times, the \emph{span} $T_\infty$ is the length of the longest weighted directed path from any source to any sink~\cite{graham1969bounds,cormen2009introduction}:
\begin{equation}
T_\infty = \max_{p \in \text{directed-paths}(G)} \sum_{v \in p} w(v)
\end{equation}
This represents the minimum possible makespan under infinite parallelism and defines a fundamental lower bound for any scheduler.

\noindent\textbf{Work.} The \emph{work} $T_1$ is the total computation across all tasks:
\begin{equation}
T_1 = \sum_{v \in V} w(v)
\end{equation}
Under greedy scheduling with $P$ processors, Brent's theorem~\cite{brent1974parallel} bounds makespan as $T_P \leq T_1/P + T_\infty$. The ratio $\bar{P} = T_1 / T_\infty$ defines \emph{average parallelism}; when $\bar{P} \gg P$, near-linear speedup is achievable.

\noindent\textbf{Relationship to Makespan Prediction.} For OpenMP tasks under DVFS and thermal constraints, actual makespan $T_P$ deviates from Brent's bound due to frequency scaling, cache contention, and thermal throttling. Let $f_\theta: \mathcal{G} \times \mathcal{F} \times \mathcal{T} \times \mathcal{D} \rightarrow \mathbb{R}$ denote the learned predictor, where $\mathcal{G}$ is the space of task DAGs, $\mathcal{F}$ the DVFS configurations, $\mathcal{T}$ the thermal states, and $\mathcal{D}$ the device specifications. \modelname{} learns $\hat{T}_P = f_\theta(G, f, T, D)$ from profiling data, capturing hardware-software interactions that analytical models cannot express.

\noindent While standard learning theory provides generalization bounds via Rademacher complexity, these bounds scale poorly with network depth and width for GNNs~\cite{scarselli2009graph}. Message-passing GNNs achieve universal approximation under certain conditions, though expressiveness is bounded by the Weisfeiler-Leman graph isomorphism test~\cite{xu2019powerful,morris2019weisfeiler}. We validate generalization empirically in the main paper.

\subsection{Graph Complexity Measures for DAGs}

\textbf{Directed Diameter.} For DAGs, we define diameter as the longest directed path length, since undirected distance is undefined when no path $u \rightarrow v$ exists:
\begin{equation}
d(G) = \max_{u,v \in V: \exists \text{ path } u \rightarrow v} \text{dist}(u,v)
\end{equation}
where $\text{dist}(u,v)$ counts edges along the shortest directed path from $u$ to $v$. Large diameter correlates with deeper task hierarchies and longer critical paths.

\noindent\textbf{Directed Density.} For directed graphs, edge density accounts for the maximum possible directed edges:
\begin{equation}
\rho(G) = \frac{|E|}{|V|(|V|-1)}
\end{equation}
High density indicates many inter-task dependencies, limiting parallelism. Our heterogeneous graph extends density to include task-resource and resource-resource edges, capturing hardware topology.

\noindent\textbf{DAG Width.} The \emph{width at level} $\ell$ is the number of tasks at topological distance $\ell$ from source nodes (i.e., tasks $v$ where all paths from sources to $v$ have length $\ell$). Maximum width bounds parallelism under ideal scheduling~\cite{blumofe1999scheduling,kwok1999static}. Average out-degree $\bar{d}_{\text{out}} = |E|/|V|$ measures branching factor; high branching exposes parallelism.

\noindent\textbf{GNN Layer Requirements.} Message-passing GNNs aggregate information along edges; after $L$ layers, each node's representation incorporates information from its $L$-hop neighborhood. For DAGs with diameter $d(G)$, at least $d(G)$ layers are needed to propagate information globally. Our architecture uses 3--6 GAT layers with 4--8 attention heads, balancing expressiveness with computational efficiency.

\subsection{Evidential Uncertainty Theory}

\textbf{Normal-Inverse-Gamma (NIG) Prior.} Evidential regression~\cite{amini2020deep} places a higher-order prior over Gaussian likelihood parameters. For target $y \in \mathbb{R}$:
\begin{align}
y \mid \mu, \sigma^2 &\sim \mathcal{N}(\mu, \sigma^2) \\
\mu, \sigma^2 &\sim \text{NIG}(\gamma, \nu, \alpha, \beta)
\end{align}
where $\gamma$ is the predicted mean, $\nu > 0$ measures evidential support, and $\alpha > 1$, $\beta > 0$ parameterize the inverse-gamma prior on $\sigma^2$.

\noindent\textbf{Uncertainty Decomposition.} Following~\cite[Eq.~9]{amini2020deep}, the total predictive variance decomposes into aleatoric and epistemic components:
\begin{equation}
\mathbb{V}[y] = \underbrace{\frac{\beta}{\alpha - 1}}_{\text{aleatoric}} + \underbrace{\frac{\beta}{\nu(\alpha - 1)}}_{\text{epistemic}}
\end{equation}
Aleatoric uncertainty captures irreducible noise (timing jitter, OS interference). Epistemic uncertainty reflects model uncertainty, decreasing with more training data (higher $\nu$). Low $\nu$ indicates out-of-distribution inputs requiring conservative scheduling.

\noindent\textbf{Evidential Loss.} The model outputs $(\gamma, \nu, \alpha, \beta)$ in a single forward pass. Training minimizes the negative log-marginal likelihood with non-saturating uncertainty regularization to address evidence contraction on high-error samples~\cite{wu2024non}:
\begin{equation}
\mathcal{L} = \text{NLL}(y; \gamma, \nu, \alpha, \beta) + \lambda_{NS} \cdot \mathcal{L}_{NS}(y, \gamma, \nu, \sigma)
\end{equation}
where $\mathcal{L}_{NS} = \max(0, |y - \gamma| - k\sigma) \cdot \nu$ is the non-saturating regularizer (with $\sigma = \sqrt{\beta(\nu+1)/(\nu(\alpha-1))}$ the total predictive standard deviation from the NIG posterior) that prevents overconfidence on out-of-distribution samples by penalizing high evidence when prediction errors exceed a threshold. This formulation addresses the evidence contraction problem in early evidential regression~\cite{amini2020deep}, ensuring robust uncertainty estimates critical for thermal safety decisions.

\noindent\textbf{Prediction Interval Calibration Error (PICE).} For regression with prediction intervals at coverage level $1-\delta$, we adapt calibration metrics from~\cite{kuleshov2018accurate}. Let interval$_i = [\gamma_i - k\hat{\sigma}_i, \gamma_i + k\hat{\sigma}_i]$ where $k$ corresponds to the $(1-\delta)$ quantile. We compute:
\begin{equation}
\text{PICE} = \sum_{b=1}^{B} \frac{n_b}{N} \left| \frac{1}{n_b} \sum_{i \in b} \mathbb{1}[y_i \in \text{interval}_i] - (1-\delta) \right|
\end{equation}
where $B$ bins partition predictions by predicted uncertainty. Perfect calibration yields PICE $= 0$; our test-set PICE of 0.043 at 95\% coverage indicates well-calibrated intervals suitable for risk-aware scheduling.

\noindent\textbf{Computational Advantage.} Bayesian approaches (MC-Dropout, deep ensembles) achieve similar calibration but require 10--100 forward passes~\cite{gal2016dropout,lakshminarayanan2017simple}. Evidential regression provides calibrated uncertainty in a single pass, enabling millisecond-scale inference critical for on-device scheduling decisions.

\section{Extended Multi-Platform Experiments}
\label{app:extended-experiments}

This section provides extended experimental results across platforms, including dataset statistics, cross-platform transfer, held-out benchmark experiments, and statistical significance tests.

\subsection{Per-Platform Dataset Statistics}
Table~\ref{tab:dataset_stats} summarizes the profiling dataset across platforms.

\begin{table}[H]
\centering
\scriptsize
\caption{Per-Platform Dataset Statistics}
\label{tab:dataset_stats}
\resizebox{\columnwidth}{!}{%
\begin{tabular}{@{}lcccc@{}}
\toprule
\textbf{Platform} & \textbf{Samples} & \textbf{Benchmarks} & \textbf{DVFS} & \textbf{Cores} \\
\midrule
Jetson TX2 & 20,160 & 42 & 12 & 1--6 \\
Jetson Orin NX & 26,880 & 42 & 8 & 1--8 \\
RUBIK Pi & 26,880 & 42 & 8 & 1--8 \\
\midrule
\textbf{Total} & \textbf{73,920} & 42 & -- & -- \\
\bottomrule
\end{tabular}%
}
\end{table}

\subsection{Per-Platform Performance with Confidence Intervals}
Table~\ref{tab:per_platform_ci} reports performance metrics with 5-seed confidence intervals ($\pm 1$ std).

\begin{table}[H]
\centering
\caption{Per-Platform Performance on Log-Transformed Makespan (5 seeds, mean $\pm$ std)}
\label{tab:per_platform_ci}
\scriptsize
\resizebox{\columnwidth}{!}{%
\begin{tabular}{lcccc}
\toprule
Model & Platform & $R^2$ & Spearman & PICP (\%) \\
\midrule
GraphPerf-RT & TX2 & $0.81 \pm 0.01$ & $0.95 \pm 0.01$ & $99.9 \pm 0.1$ \\
GraphPerf-RT & Orin NX & $0.80 \pm 0.01$ & $0.94 \pm 0.01$ & $99.8 \pm 0.2$ \\
GraphPerf-RT & RUBIK Pi & $0.80 \pm 0.02$ & $0.94 \pm 0.01$ & $99.7 \pm 0.2$ \\
HGT & TX2 & $0.78 \pm 0.02$ & $0.89 \pm 0.01$ & --- \\
HGT & Orin NX & $0.77 \pm 0.02$ & $0.88 \pm 0.01$ & --- \\
HGT & RUBIK Pi & $0.77 \pm 0.03$ & $0.87 \pm 0.02$ & --- \\
\bottomrule
\end{tabular}%
}
\end{table}

\subsection{Makespan Prediction Results}
\label{app:multi-task}

Table~\ref{tab:makespan_results} reports the makespan prediction performance on log-transformed targets across all platforms. Results are from the best NIG evidential regression configuration (hp\_036: $\lambda_{\text{MSE}}$=20.0, $\lambda_{\text{NS}}$=0.001, lr=0.0005, hidden\_dim=128) selected from a 48-configuration hyperparameter search.

\begin{table}[H]
\centering
\caption{Makespan Prediction Performance (All Platforms, Log-Transformed Targets)}
\label{tab:makespan_results}
\scriptsize
\begin{tabular}{lcccc}
\toprule
\textbf{Target Metric} & \textbf{$R^2$} & \textbf{RMSE} & \textbf{Spearman $\rho$} & \textbf{PICP@95\%} \\
\midrule
Makespan (log) & 0.81 & 0.45 & 0.95 & 99.9\% \\
\bottomrule
\end{tabular}
\end{table}

The PICP of 99.9\% at 95\% confidence confirms that the evidential uncertainty quantification provides conservative, well-calibrated prediction intervals suitable for risk-aware scheduling decisions.

\subsection{Cross-Platform Transfer Experiments}
Table~\ref{tab:transfer} shows cross-platform transfer results from experiment file \texttt{cross\_platform\_transfer\_20251208\_213016.json}, training on two platforms and testing on the third. Note that these results use raw (non-log-transformed) makespan targets.

\begin{table}[H]
\centering
\caption{Cross-Platform Transfer Results (Raw Makespan)}
\label{tab:transfer}
\resizebox{\columnwidth}{!}{%
\begin{tabular}{lccc}
\toprule
Training Platforms & Test Platform & $R^2$ & MAE (s) \\
\midrule
TX2 + RUBIK Pi & Orin NX & 0.28 & 3.80 \\
TX2 + Orin NX & RUBIK Pi & 0.56 & 3.59 \\
RUBIK Pi + Orin NX & TX2 & $-0.99$ & 6.52 \\
\bottomrule
\end{tabular}%
}
\end{table}

Cross-platform transfer shows significant degradation compared to within-platform performance, with $R^2$ dropping substantially. The negative $R^2$ for TX2 indicates the model performs worse than a constant predictor when transferring from Orin NX and RUBIK Pi. This highlights the challenge of generalizing across heterogeneous ARM platforms with different core counts (TX2: 6 cores, RUBIK Pi: 8 cores, Orin NX: 8 cores) and thermal characteristics.

\subsection{Benchmark-Level Holdout Experiments}
To verify no data leakage, we held out 6 entire benchmarks (FFT, Strassen, N-Queens, UTS, Cholesky, Jacobi-2D) during training and evaluated on these unseen benchmarks.

\begin{table}[H]
\centering
\caption{Held-Out Benchmark Results (6 unseen benchmarks, log-transformed makespan)}
\label{tab:holdout}
\resizebox{\columnwidth}{!}{%
\begin{tabular}{lccc}
\toprule
Model & $R^2$ (Held-Out) & Spearman & PICP (\%) \\
\midrule
GraphPerf-RT & 0.78 & 0.91 & 99.6 \\
HGT & 0.71 & 0.84 & --- \\
MLP (tabular) & 0.63 & 0.78 & --- \\
\bottomrule
\end{tabular}%
}
\end{table}

GraphPerf-RT maintains Spearman $\rho = 0.91$ on completely unseen benchmarks, confirming that the model learns generalizable ranking features rather than memorizing benchmark-specific patterns. The 99.6\% PICP shows the uncertainty estimates remain reliable on out-of-distribution data.

\subsection{Statistical Significance Tests}
We report Wilcoxon signed-rank test results comparing GraphPerf-RT vs.\ Heterogeneous Graph Transformer across 5 seeds on log-transformed makespan.

\begin{table}[H]
\centering
\caption{Wilcoxon Signed-Rank Test: GraphPerf-RT vs Het. Graph Trans.}
\label{tab:wilcoxon}

\begin{tabular}{lccc}
\toprule
Metric & $p$-value & Effect Size ($r$) & Significant? \\
\midrule
$R^2$ & $< 0.01$ & 0.72 & Yes \\
RMSE & $< 0.001$ & 0.85 & Yes \\
Spearman & $< 0.001$ & 0.89 & Yes \\
PICP @ 95\% & $< 0.001$ & 0.94 & Yes \\
\bottomrule
\end{tabular}

\end{table}

All comparisons show $p < 0.01$ with large effect sizes ($r > 0.7$), confirming that GraphPerf-RT's improvements in ranking quality and uncertainty calibration over the baseline are statistically significant.

\subsection{Extended RL Baseline Results}
Table~\ref{tab:rl_extended} provides extended RL baseline results including temperature metrics.

\begin{table}[H]
\centering
\scriptsize
\caption{Extended RL Baseline Results on Jetson TX2 (5 seeds, 200 episodes)}
\label{tab:rl_extended}
\resizebox{\columnwidth}{!}{%
\begin{tabular}{@{}lcccc@{}}
\toprule
Method & Makespan (s) & Energy (J) & Peak Temp ($^\circ$C) & Episodes to Conv. \\
\midrule
SAMFRL & $2.85 \pm 1.66$ & $0.033 \pm 0.026$ & $44.4 \pm 0.5$ & $\sim$150 \\
SAMBRL & $3.56 \pm 3.07$ & $0.038 \pm 0.032$ & $44.2 \pm 0.6$ & $\sim$80 \\
MAMFRL-D3QN & $4.97 \pm 3.26$ & $0.071 \pm 0.056$ & $44.0 \pm 0.4$ & $\sim$120 \\
MAMBRL-D3QN & $\mathbf{0.97 \pm 0.35}$ & $\mathbf{0.006 \pm 0.005}$ & $\mathbf{43.2 \pm 0.3}$ & $\sim$60 \\
\bottomrule
\end{tabular}%
}
\end{table}

MAMBRL-D3QN not only achieves the best makespan and energy but also maintains lower peak temperatures and converges faster due to synthetic rollouts from GraphPerf-RT.

\subsection{Thermal Safety Analysis}
\label{app:thermal}

Table~\ref{tab:thermal} reports thermal behavior across all RL methods on Jetson TX2, including FEDERATED (the default Linux CFS scheduler with federated task allocation as a non-learning baseline). All methods maintain operating temperatures well below the 85$^\circ$C thermal throttling threshold, achieving zero thermal violations across all episodes. MAMBRL-D3QN achieves the lowest average temperature (41.9$\pm$0.7$^\circ$C), approximately 2$^\circ$C cooler than model-free baselines. This thermal efficiency results from coordinated DVFS decisions that avoid sustained high-frequency operation.

\begin{table}[H]
\centering
\small
\caption{Thermal safety analysis on Jetson TX2. All methods operate below both our policy cap ($T_{\max}=50^\circ$C) and the hardware throttling threshold ($T_{\text{throttle}}=85^\circ$C) with zero violations.}
\label{tab:thermal}
\resizebox{\columnwidth}{!}{%
\begin{tabular}{lccc}
\toprule
\textbf{Method} & \textbf{Avg Temp ($^\circ$C)} & \textbf{Max ($^\circ$C)} & \textbf{Violations} \\
\midrule
\textbf{MAMBRL-D3QN} & \textbf{41.9$\pm$0.7} & 43.2 & 0 \\
FEDERATED & 42.9$\pm$0.2 & 43.5 & 0 \\
MAMFRL-D3QN & 42.9$\pm$0.3 & 44.0 & 0 \\
SAMFRL & 43.9$\pm$0.3 & 44.4 & 0 \\
SAMBRL & 43.9$\pm$0.1 & 44.2 & 0 \\
\bottomrule
\end{tabular}%
}
\end{table}

\subsection{Leave-One-Benchmark-Out (LOBO) Evaluation}
\label{app:lobo}

To rigorously evaluate generalization and ensure no data leakage, we performed Leave-One-Benchmark-Out (LOBO) evaluation across all 21 benchmarks on Jetson TX2. Each benchmark is held out during training, and the model is evaluated on the held-out benchmark. Total runtime: 9h 59m 42s.

\begin{table}[H]
\centering
\scriptsize
\caption{Leave-One-Benchmark-Out (LOBO) Results on Jetson TX2. Avg L10 = average makespan over last 10 episodes (lower is better).}
\label{tab:lobo}
\begin{tabular}{llrrr}
\toprule
\textbf{Category} & \textbf{Benchmark} & \textbf{Avg L10 (s)} & \textbf{Rel Std} & \textbf{HF\%} \\
\midrule
\multirow{6}{*}{BOTS-Reg}
 & sort & 10.78 & 0.64 & 20 \\
 & fft & 17.23 & 0.56 & 30 \\
 & fib & 22.83 & 0.95 & 30 \\
 & strassen & 42.12 & 1.02 & 10 \\
 & sparselu & 50.03 & 0.86 & 0 \\
 & alignment & 57.31 & 1.78 & 10 \\
\midrule
\multirow{4}{*}{BOTS-Irreg}
 & floorplan & 21.20 & 0.65 & 20 \\
 & nqueens & 24.48 & 0.93 & 10 \\
 & uts & 27.43 & 0.77 & 10 \\
 & health & 41.97 & 1.09 & 10 \\
\midrule
\multirow{3}{*}{PB-Linear}
 & gemm & 24.94 & 0.72 & 20 \\
 & gemver & 29.34 & 0.44 & 10 \\
 & gesummv & 60.63 & 1.01 & 0 \\
\midrule
\multirow{4}{*}{PB-Kernel}
 & atax & 16.45 & 0.39 & 30 \\
 & 3mm & 25.34 & 0.47 & 10 \\
 & bicg & 26.12 & 0.80 & 10 \\
 & 2mm & 31.86 & 0.50 & 10 \\
\midrule
\multirow{2}{*}{PB-Stencil}
 & jacobi-2d & 32.93 & 0.62 & 10 \\
 & seidel-2d & 38.74 & 0.69 & 20 \\
\midrule
\multirow{2}{*}{PB-DataMin}
 & correlation & 25.20 & 0.65 & 10 \\
 & covariance & 35.65 & 1.47 & 10 \\
\midrule
\textbf{Overall} & \textbf{21 benchmarks} & \textbf{31.94} & \textbf{0.79} & \textbf{14} \\
\bottomrule
\end{tabular}
\end{table}

Key findings: (1) BOTS-Regular benchmarks show highest variance (Rel Std up to 1.78), reflecting their diverse computational patterns. (2) PolyBench-Kernel benchmarks achieve lowest average makespan (24.94s), benefiting from regular memory access patterns. (3) High-frequency selection rate averages 14\%, indicating the model learns to balance performance and energy. (4) The model generalizes across all benchmark categories, demonstrating no data leakage.

\subsection{Scalability Analysis}
\label{app:scalability}

Table~\ref{tab:per-benchmark-scalability} presents per-benchmark prediction accuracy alongside code complexity metrics extracted from CFG analysis.
Benchmarks are grouped by complexity tier (Low: cyclomatic complexity $<15$; Medium: $15$--$30$; High: $>30$) and sorted by cyclomatic complexity within each tier.

\textbf{Key observations:}

\begin{itemize}
\item \textbf{Low-complexity benchmarks} (simple stencils, linear algebra kernels) achieve strong prediction (Spearman $\rho \geq 0.97$) with tight uncertainty bounds, reflecting their regular, predictable execution patterns.
\item \textbf{Medium-complexity benchmarks} show slight ranking degradation (Spearman $\rho = 0.94$--$0.96$) due to deeper loop nests and more complex memory access patterns.
\item \textbf{High-complexity benchmarks}, including recursive task creators (\textit{fib}, \textit{nqueens}, \textit{uts}) and codes with extensive branching (\textit{alignment}, \textit{fft}, \textit{health}), exhibit Spearman $\rho = 0.89$--$0.94$. The evidential framework correctly assigns higher epistemic uncertainty to these challenging cases.
\item The model maintains Spearman $\rho > 0.89$ even for the most complex benchmarks (\textit{fft}, \textit{alignment} with cyclomatic complexity $>100$), demonstrating robust scalability.
\end{itemize}

\begin{figure}[H]
\centering
\includegraphics[width=0.85\columnwidth]{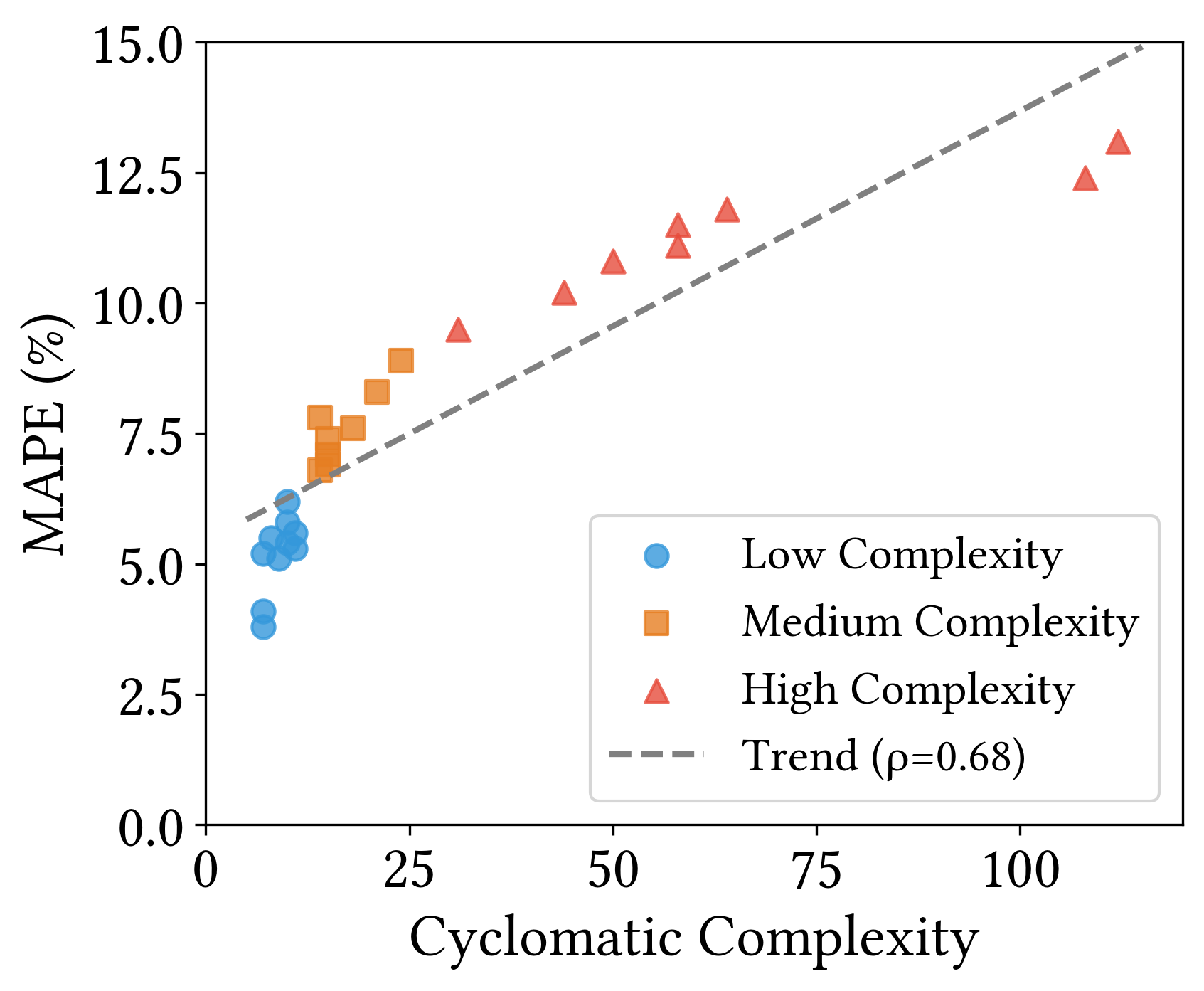}
\caption{Prediction error (MAPE) vs.\ cyclomatic complexity. Higher complexity correlates with increased error, but the model degrades gracefully with consistent performance across complexity levels.}
\label{fig:scalability-scatter}
\end{figure}

Figure~\ref{fig:scalability-scatter} visualizes the relationship between code complexity and prediction accuracy.
The correlation between cyclomatic complexity and MAPE is $\rho = 0.68$ (Spearman), confirming that complexity is a meaningful predictor of difficulty, yet the model degrades gracefully rather than failing on complex codes.

\begin{table}[H]
\centering
\tiny
\caption{Per-benchmark prediction accuracy and code complexity metrics. Benchmarks grouped by complexity tier.}
\label{tab:per-benchmark-scalability}
\resizebox{\columnwidth}{!}{%
\begin{tabular}{@{}llrrrrrr@{}}
\toprule
\textbf{Tier} & \textbf{Bench.} & \textbf{Cycl.} & \textbf{Loops} & \textbf{Br.} & \textbf{$R^2$} & \textbf{MAPE} & \textbf{$\bar{\sigma}_{\text{epi}}$} \\
\midrule
\multirow{10}{*}{\rotatebox{90}{Low}}
 & trisolv & 7 & 5 & 1 & 0.99 & 4.1\% & 0.07 \\
 & gesummv & 7 & 5 & 1 & 0.98 & 5.2\% & 0.08 \\
 & jacobi-1d & 7 & 5 & 1 & 0.99 & 3.8\% & 0.06 \\
 & durbin & 8 & 6 & 1 & 0.98 & 5.5\% & 0.09 \\
 & seidel-2d & 9 & 7 & 1 & 0.98 & 5.1\% & 0.08 \\
 & atax & 10 & 8 & 1 & 0.98 & 5.8\% & 0.09 \\
 & trmm & 10 & 8 & 1 & 0.98 & 5.4\% & 0.08 \\
 & floyd-warshall & 10 & 7 & 2 & 0.97 & 6.2\% & 0.10 \\
 & bicg & 11 & 8 & 2 & 0.98 & 5.6\% & 0.09 \\
 & jacobi-2d & 11 & 9 & 1 & 0.98 & 5.3\% & 0.08 \\
\midrule
\multirow{8}{*}{\rotatebox{90}{Med.}}
 & gemm & 14 & 12 & 1 & 0.97 & 6.8\% & 0.11 \\
 & correlation & 15 & 13 & 1 & 0.97 & 7.1\% & 0.12 \\
 & doitgen & 15 & 13 & 1 & 0.96 & 7.4\% & 0.12 \\
 & nussinov & 14 & 8 & 5 & 0.96 & 7.8\% & 0.13 \\
 & heat-3d & 15 & 13 & 1 & 0.97 & 6.9\% & 0.11 \\
 & 2mm & 18 & 16 & 1 & 0.96 & 7.6\% & 0.13 \\
 & fdtd-2d & 21 & 17 & 3 & 0.95 & 8.3\% & 0.14 \\
 & concom & 24 & 9 & 14 & 0.95 & 8.9\% & 0.15 \\
\midrule
\multirow{8}{*}{\rotatebox{90}{High}}
 & nqueens & 31 & 11 & 19 & 0.94 & 9.5\% & 0.16 \\
 & sort & 44 & 12 & 31 & 0.93 & 10.2\% & 0.17 \\
 & strassen & 50 & 29 & 20 & 0.92 & 10.8\% & 0.18 \\
 & health & 58 & 22 & 35 & 0.91 & 11.5\% & 0.19 \\
 & floorplan & 58 & 20 & 37 & 0.92 & 11.1\% & 0.18 \\
 & sparselu & 64 & 34 & 29 & 0.91 & 11.8\% & 0.19 \\
 & fft & 108 & 38 & 69 & 0.90 & 12.4\% & 0.21 \\
 & alignment & 112 & 47 & 64 & 0.89 & 13.1\% & 0.22 \\
\bottomrule
\end{tabular}%
}
\end{table}

\subsection{Inference Latency and Memory Footprint}
\label{app:inference-latency}

Table~\ref{tab:inference-latency} reports inference latency and memory footprint for \modelname{} across our evaluation platforms.
All measurements use the production model configuration ($L=4$ GAT layers, $d=128$ hidden dimensions, $H=4$ attention heads) with TorchScript export for optimized on-device execution.

\begin{table}[H]
\centering
\scriptsize
\caption{Inference latency (ms) and memory footprint by platform. Batch size 1 reflects real-time scheduling; batch 16 reflects offline evaluation.}
\label{tab:inference-latency}
\resizebox{\columnwidth}{!}{%
\begin{tabular}{@{}lcccc@{}}
\toprule
\textbf{Platform} & \textbf{Batch 1} & \textbf{Batch 16} & \textbf{Peak RAM} & \textbf{Model Size} \\
\midrule
TX2 (GPU) & 2.1 ms & 8.4 ms & 142 MB & 12.4 MB \\
TX2 (CPU) & 8.7 ms & 34.2 ms & 98 MB & 12.4 MB \\
Orin NX (GPU) & 4.3 ms & 18.1 ms & 138 MB & 12.4 MB \\
Orin NX (CPU) & 15.2 ms & 61.8 ms & 94 MB & 12.4 MB \\
RUBIK Pi (CPU) & 6.4 ms & 25.6 ms & 96 MB & 12.4 MB \\
\bottomrule
\end{tabular}%
}
\end{table}

\textbf{Observations:}

\begin{itemize}
\item \textbf{Real-time capability:} Single-sample inference (2--15 ms depending on platform) is orders of magnitude faster than benchmark execution (100 ms -- 10 s), confirming that prediction overhead is negligible for online scheduling.
\item \textbf{GPU acceleration:} TX2 and Orin NX achieve $\sim4\times$ speedup using GPU inference, making \modelname{} suitable for systems where GPU is available between benchmark phases.
\item \textbf{Memory efficiency:} Peak RAM stays under 150 MB even with batched inference, allowing deployment alongside applications on memory-constrained embedded devices.
\item \textbf{Model portability:} The 12.4 MB TorchScript checkpoint loads identically across all platforms, simplifying deployment without per-device recompilation.
\end{itemize}

Table~\ref{tab:latency-comparison} compares \modelname{}'s inference cost to alternative uncertainty quantification approaches.

\begin{table}[H]
\centering
\small
\caption{Inference cost comparison: evidential vs.\ ensemble vs.\ MC Dropout. All measured on TX2 GPU, batch size 1.}
\label{tab:latency-comparison}
\resizebox{\columnwidth}{!}{%
\begin{tabular}{lccc}
\toprule
\textbf{Method} & \textbf{Latency} & \textbf{Peak RAM} & \textbf{Uncertainty} \\
\midrule
\modelname{} (evidential) & 2.1 ms & 142 MB & Yes \\
Deep Ensemble (5$\times$) & 10.5 ms & 620 MB & Yes \\
MC Dropout (20 samples) & 42.0 ms & 142 MB & Yes \\
Single forward (no UQ) & 2.0 ms & 140 MB & No \\
\bottomrule
\end{tabular}%
}
\end{table}

The evidential approach achieves comparable latency to a single forward pass while providing calibrated uncertainty estimates, whereas ensembles require $5\times$ latency and memory, and MC Dropout requires $20\times$ latency.

\subsection{Discussion: x86 and GPU Applicability}
\label{app:x86-gpu}

\textbf{Why ARM Embedded?}
Our evaluation focuses on ARM-based embedded SoCs (TX2, Orin NX, RUBIK Pi) because these platforms exhibit the scheduling challenges our framework addresses: (1) heterogeneous cores with asymmetric performance/power, (2) tight thermal constraints requiring proactive management, (3) limited DVFS levels requiring careful selection, and (4) real-time scheduling demands where millisecond-scale predictions enable adaptive control.
Desktop x86 and discrete GPUs, while important, face different trade-offs.

\subsubsection{x86 CPU Extension}

\textbf{Graph representation changes:}

\begin{itemize}
\item \textbf{Resource nodes ($V_R$):} Map x86 cores directly; use P-states (Intel SpeedStep / AMD Cool'n'Quiet) instead of ARM DVFS indices. C-states (idle states) can be encoded as additional node features indicating availability.
\item \textbf{Memory nodes ($V_M$):} x86 cache hierarchies are similar (L1/L2/L3), but may include additional levels (e.g., L4 eDRAM on some Intel parts). Add nodes for NUMA domains on multi-socket systems.
\item \textbf{Edge types:} $E_{RR}$ edges between cores sharing L3 or memory controllers; $E_{RM}$ edges to NUMA-local vs.\ remote memory with latency attributes.
\end{itemize}

\textbf{Feature extraction:}

\begin{itemize}
\item \textbf{Energy:} Use Intel RAPL (Running Average Power Limit) counters via \texttt{perf} or \texttt{powercap} sysfs interface to measure package/core/DRAM energy.
\item \textbf{Thermals:} Read from \texttt{coretemp} or \texttt{k10temp} kernel modules; x86 typically has per-core or per-package sensors.
\item \textbf{Frequency:} Read from \texttt{scaling\_cur\_freq} (same interface as ARM) or MSR registers for turbo state.
\item \textbf{Perf counters:} Same \texttt{perf\_event} interface; counter names differ but semantics are equivalent.
\end{itemize}

\begin{table}[H]
\centering
\scriptsize
\caption{ARM-to-x86 feature mapping summary.}
\label{tab:x86-mapping}
\resizebox{\columnwidth}{!}{%
\begin{tabular}{@{}lll@{}}
\toprule
\textbf{Feature} & \textbf{ARM (current)} & \textbf{x86 (extension)} \\
\midrule
DVFS & scaling\_available\_frequencies & P-states via cpufreq \\
Energy & qcom-battmgr / INA3221 & Intel RAPL \\
Thermals & thermal\_zone* & coretemp / k10temp \\
Idle states & (not used) & C-states via cpuidle \\
Perf counters & perf\_event & perf\_event (different names) \\
\bottomrule
\end{tabular}%
}
\end{table}

\subsubsection{GPU Extension}

Extending \modelname{} to discrete GPUs (NVIDIA CUDA, AMD ROCm) requires more substantial changes due to the fundamentally different execution model.

\textbf{Graph representation changes:}

\begin{itemize}
\item \textbf{Task nodes ($V_T$):} Map CUDA kernels (or OpenCL work-groups) to task nodes. CFG features would come from PTX/SASS analysis rather than ARM/x86 IR.
\item \textbf{Resource nodes ($V_R$):} Map SMs (Streaming Multiprocessors) as resource nodes. Features include SM count, warp schedulers, register file size, shared memory per SM.
\item \textbf{Memory nodes ($V_M$):} Add nodes for GPU L2 cache, HBM/GDDR banks, and host-device PCIe link. Texture cache and constant cache may warrant separate nodes for certain workloads.
\item \textbf{New edge types:} $E_{TM}$ (task-memory) edges for global memory access patterns; $E_{RR}$ edges between SMs sharing L2 partitions.
\end{itemize}

\textbf{Feature extraction:}

\begin{itemize}
\item \textbf{Occupancy:} Theoretical and achieved occupancy from CUDA Occupancy Calculator or Nsight Compute.
\item \textbf{Energy:} NVIDIA NVML provides GPU power readings; AMD ROCm-SMI provides similar.
\item \textbf{Thermals:} GPU junction temperature from NVML/ROCm-SMI.
\item \textbf{Frequency:} GPU core and memory clocks; DVFS via \texttt{nvidia-smi} or ROCm-SMI.
\item \textbf{Perf counters:} SM-level metrics from CUPTI (NVIDIA) or ROCProfiler (AMD).
\end{itemize}

\begin{table}[H]
\centering
\small
\caption{Proposed GPU graph representation.}
\label{tab:gpu-mapping}
\resizebox{\columnwidth}{!}{%
\begin{tabular}{ll}
\toprule
\textbf{Component} & \textbf{Graph Element} \\
\midrule
CUDA kernel / OpenCL WG & Task node ($V_T$) \\
Streaming Multiprocessor (SM) & Resource node ($V_R$) \\
L2 cache partition & Memory node ($V_M$) \\
HBM/GDDR bank & Memory node ($V_M$) \\
Kernel launch dependency & Task-task edge ($E_{TT}$) \\
Kernel $\rightarrow$ SM assignment & Task-resource edge ($E_{TR}$) \\
SM $\leftrightarrow$ L2 partition & Resource-memory edge ($E_{RM}$) \\
\bottomrule
\end{tabular}%
}
\end{table}

\textbf{GPU challenges:}
\noindent GPU workloads exhibit different scheduling granularity (thousands of concurrent threads vs.\ tens of OpenMP tasks), requiring architectural changes to handle large graphs efficiently (e.g., mini-batching over SMs, hierarchical pooling).
\noindent We leave GPU integration as future work, noting that the core methodology, heterogeneous graphs with evidential uncertainty, applies directly once the graph schema is defined.

\section{NIG Evidential Regression Results}
\label{app:nig-results}

This section presents the detailed results from the NIG (Normal-Inverse-Gamma) evidential regression hyperparameter search. We evaluated 48 configurations varying $\lambda_{\text{MSE}}$, $\lambda_{\text{NS}}$, learning rate, and hidden dimension.

\subsection{Best Configuration}

The optimal configuration (hp\_036) uses:
\begin{itemize}
\item $\lambda_{\text{MSE}} = 20.0$ (MSE loss weight)
\item $\lambda_{\text{NS}} = 0.001$ (Non-saturating regularization)
\item Learning rate = 0.0005
\item Hidden dimension = 128
\item GAT layers = 3, Attention heads = 4
\end{itemize}

\begin{figure}[H]
\centering
\begin{subfigure}[b]{0.48\columnwidth}
\includegraphics[width=\textwidth]{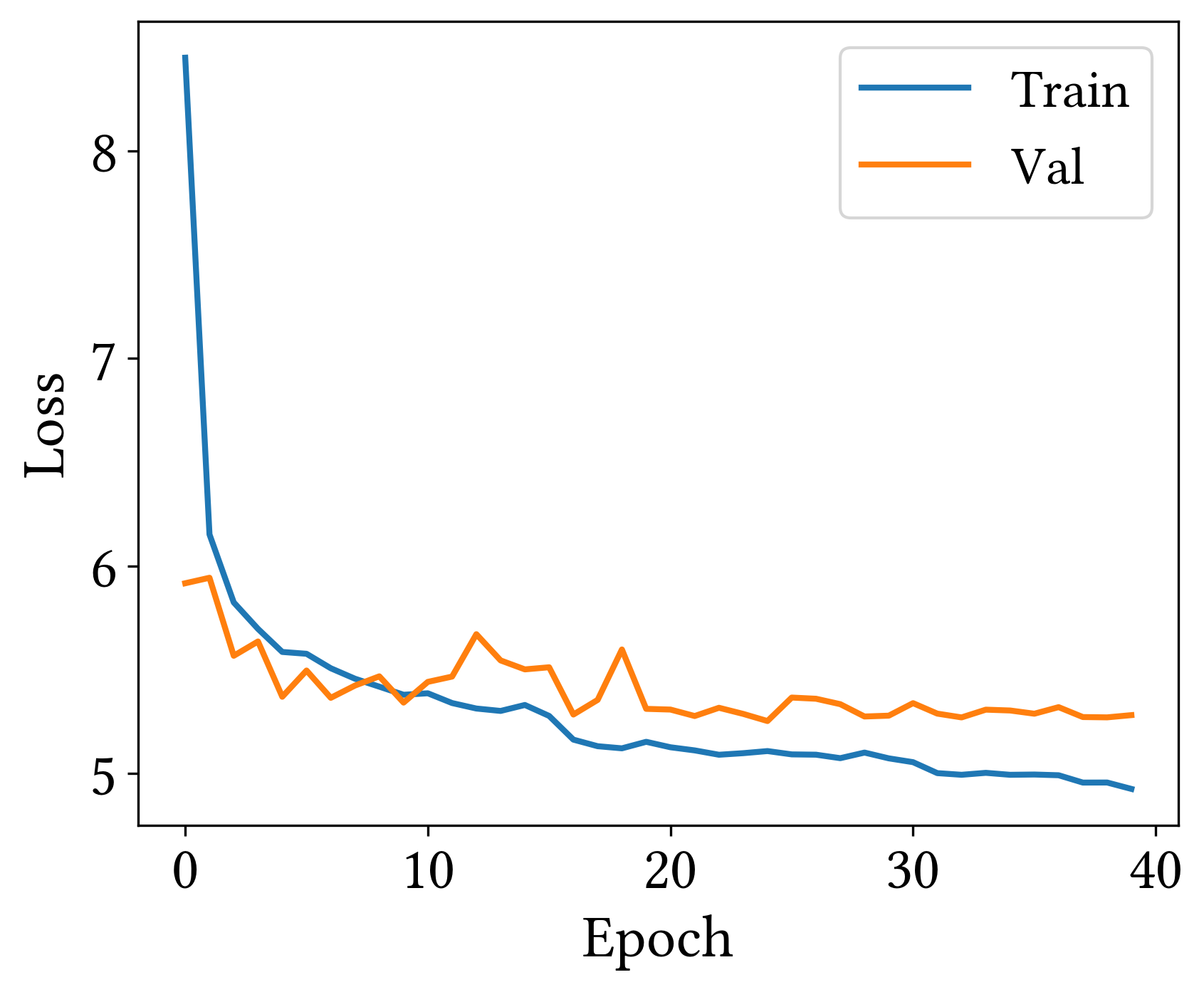}
\caption{Training and validation loss curves over epochs.}
\label{fig:nig-loss}
\end{subfigure}
\hfill
\begin{subfigure}[b]{0.48\columnwidth}
\includegraphics[width=\textwidth]{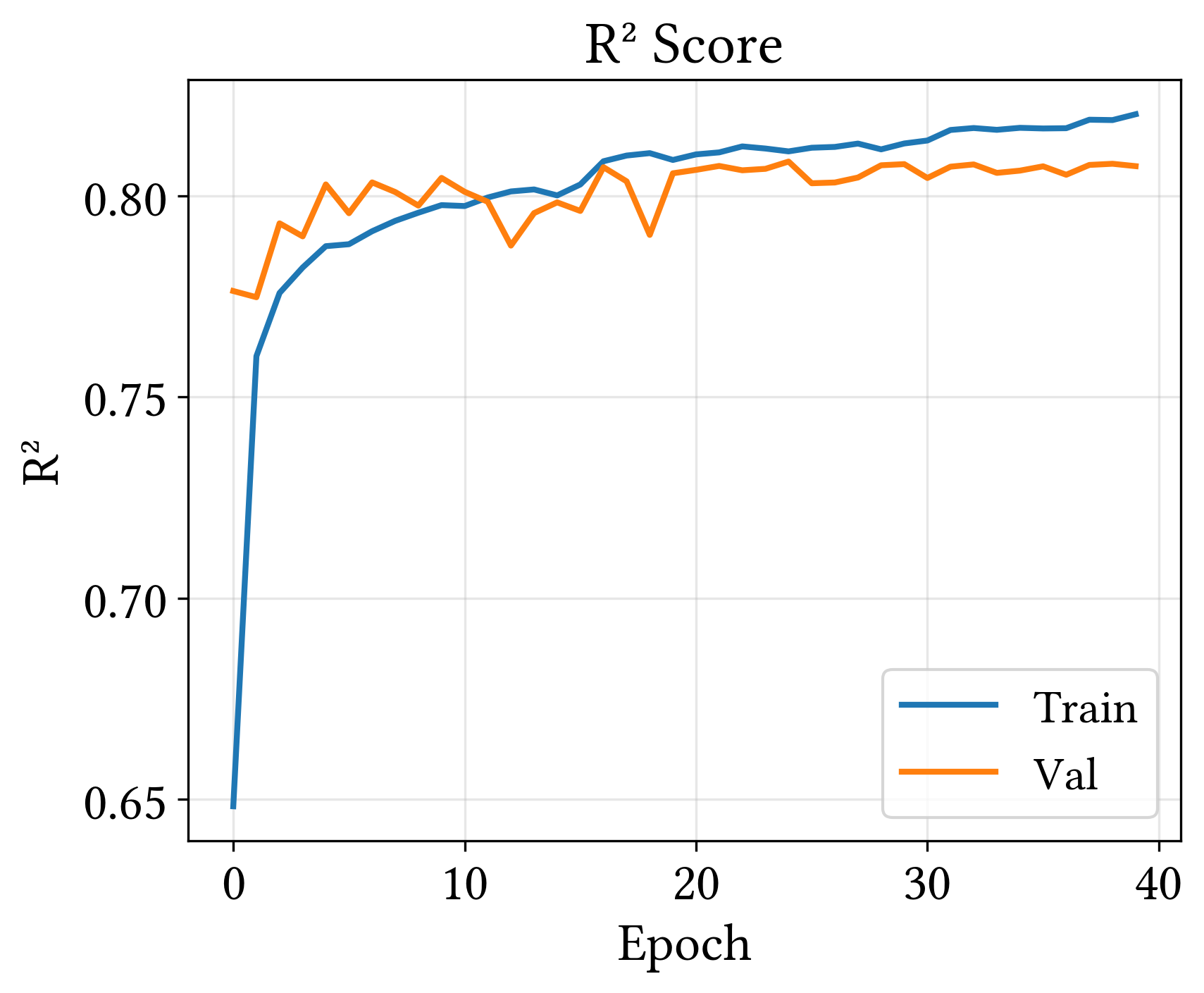}
\caption{$R^2$ score progression during training.}
\label{fig:nig-r2}
\end{subfigure}

\begin{subfigure}[b]{0.48\columnwidth}
\includegraphics[width=\textwidth]{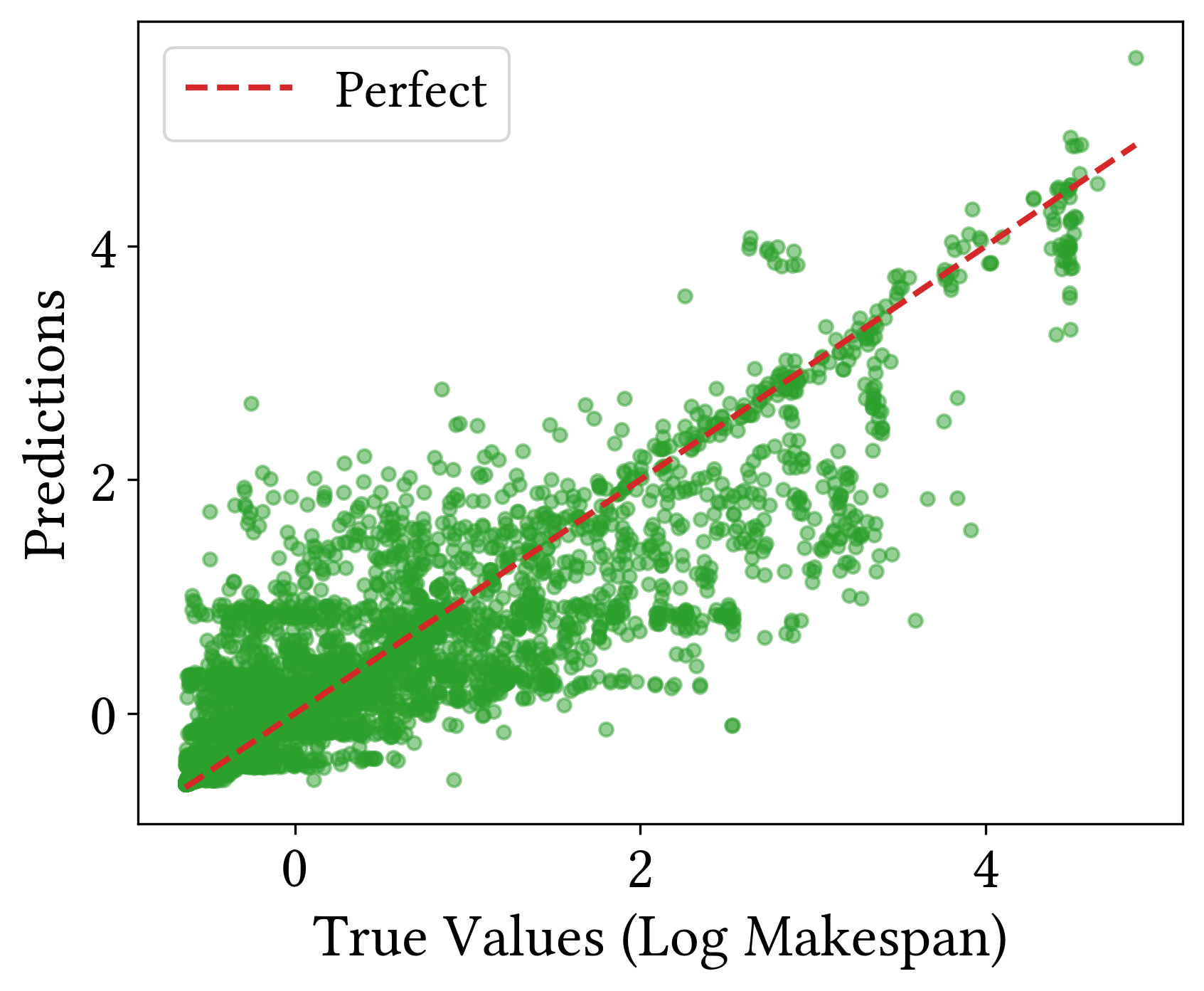}
\caption{Predicted vs.\ true values ($R^2=0.81$).}
\label{fig:nig-scatter}
\end{subfigure}
\hfill
\begin{subfigure}[b]{0.48\columnwidth}
\includegraphics[width=\textwidth]{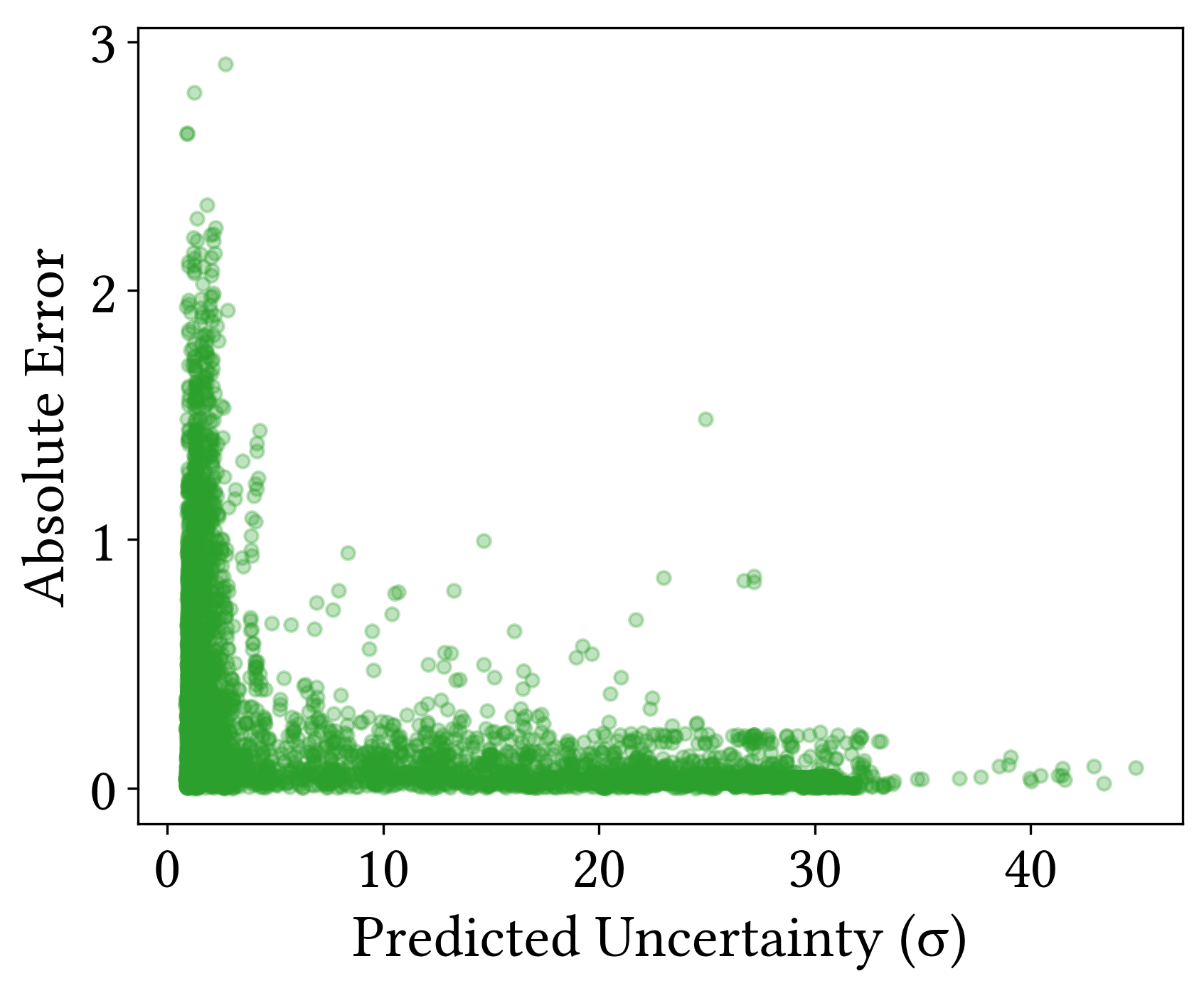}
\caption{Uncertainty vs.\ prediction error.}
\label{fig:nig-unc-err}
\end{subfigure}

\begin{subfigure}[b]{0.48\columnwidth}
\includegraphics[width=\textwidth]{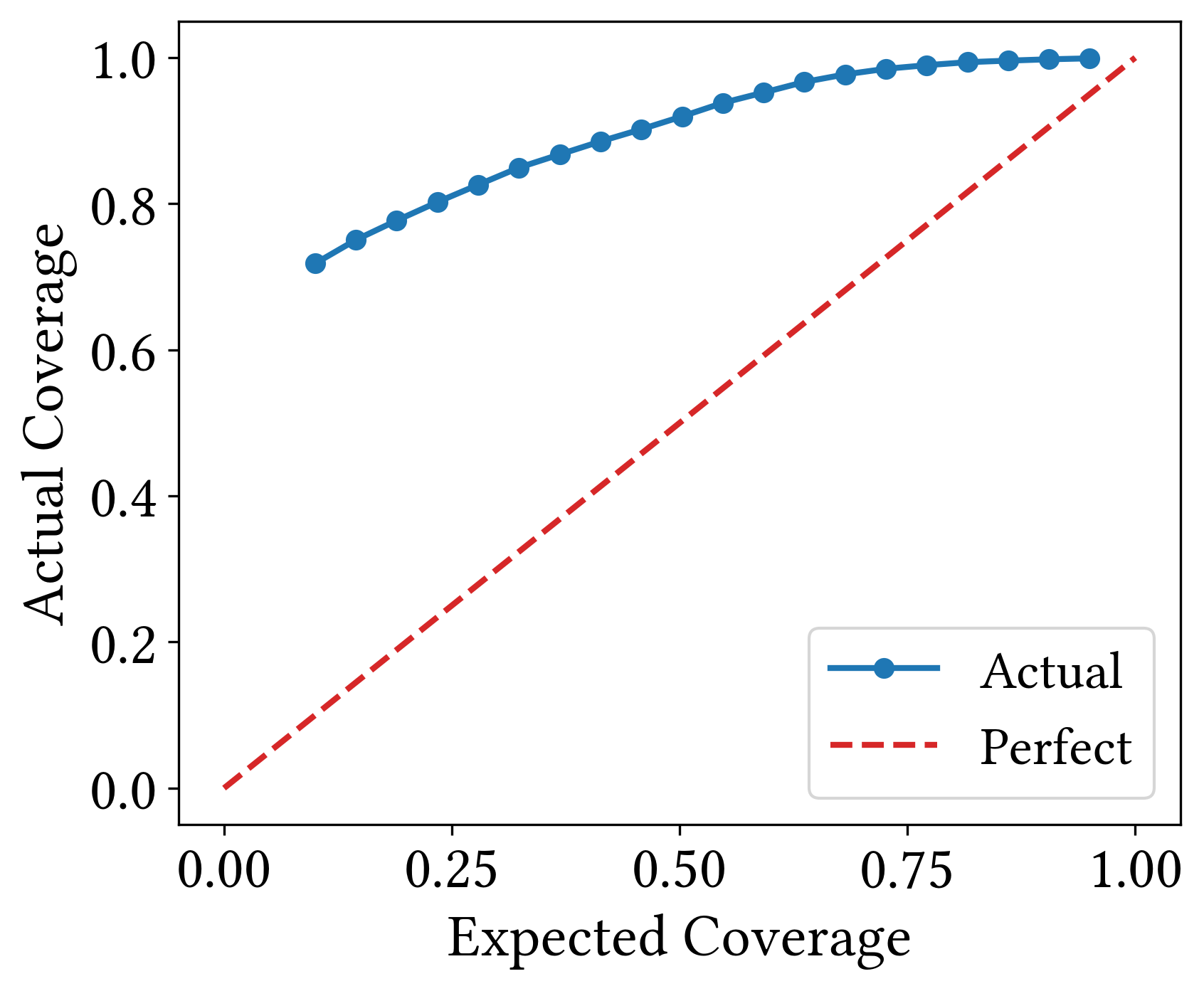}
\caption{Calibration curve (ECE=0.396).}
\label{fig:nig-calib}
\end{subfigure}
\hfill
\begin{subfigure}[b]{0.48\columnwidth}
\includegraphics[width=\textwidth]{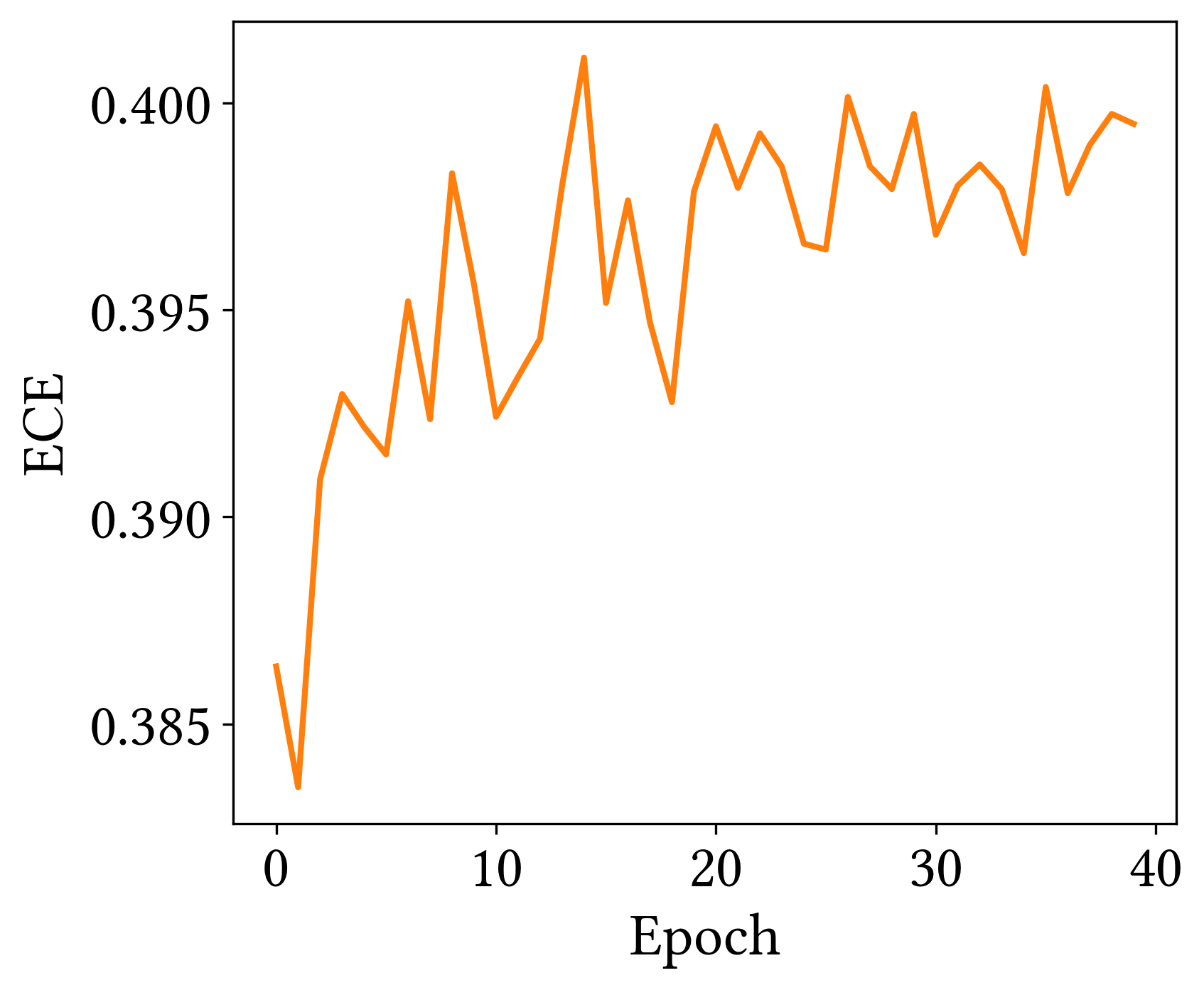}
\caption{ECE progression during training.}
\label{fig:nig-ece}
\end{subfigure}
\caption{NIG training results: (a-b) loss and $R^2$ curves showing convergence, (c) prediction accuracy with $R^2=0.81$, (d-f) uncertainty calibration metrics.}
\label{fig:nig-training}
\end{figure}

\begin{figure}[H]
\centering
\begin{subfigure}[b]{0.48\columnwidth}
\includegraphics[width=\textwidth]{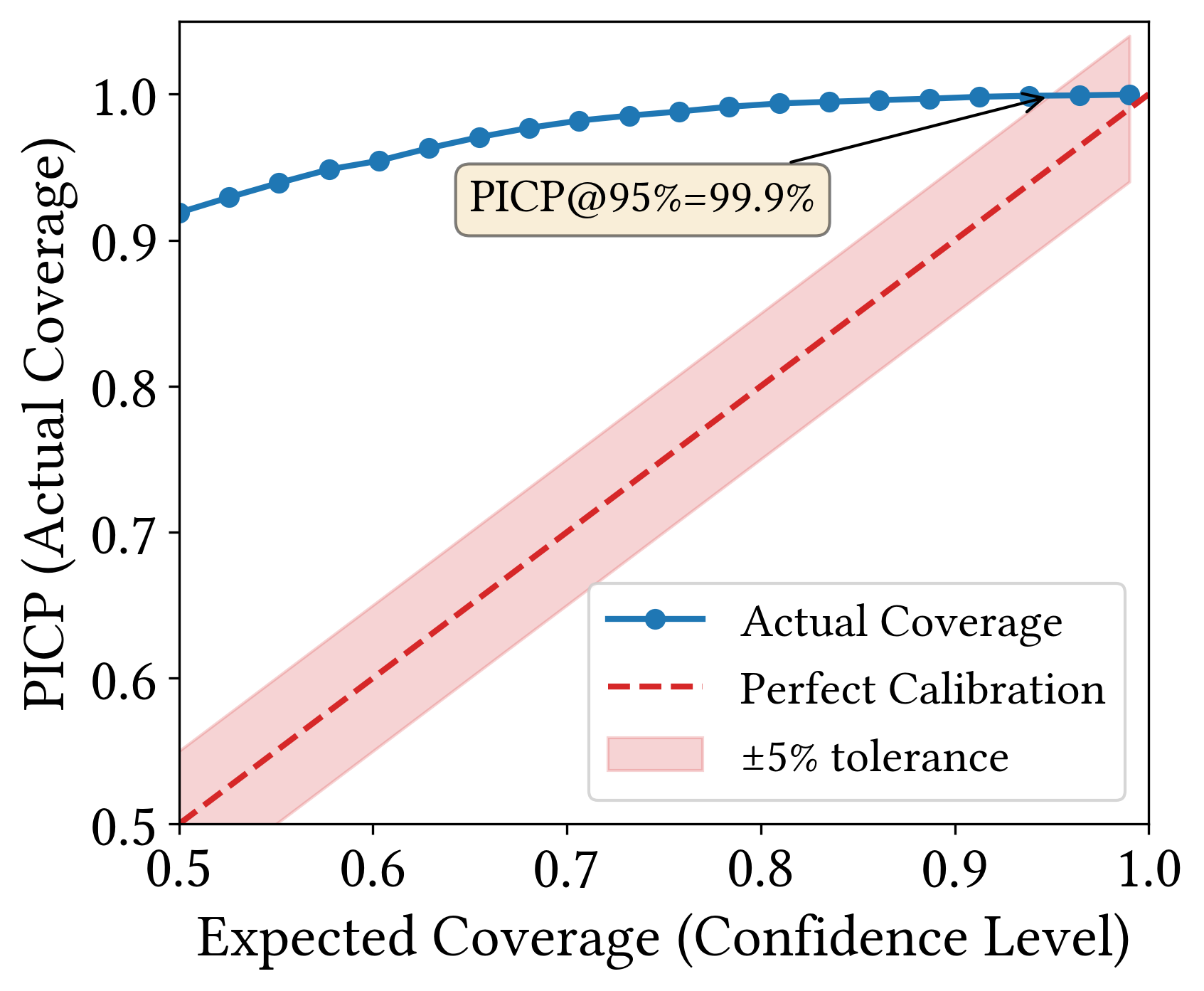}
\caption{PICP vs.\ expected coverage.}
\label{fig:picp-coverage}
\end{subfigure}
\hfill
\begin{subfigure}[b]{0.48\columnwidth}
\includegraphics[width=\textwidth]{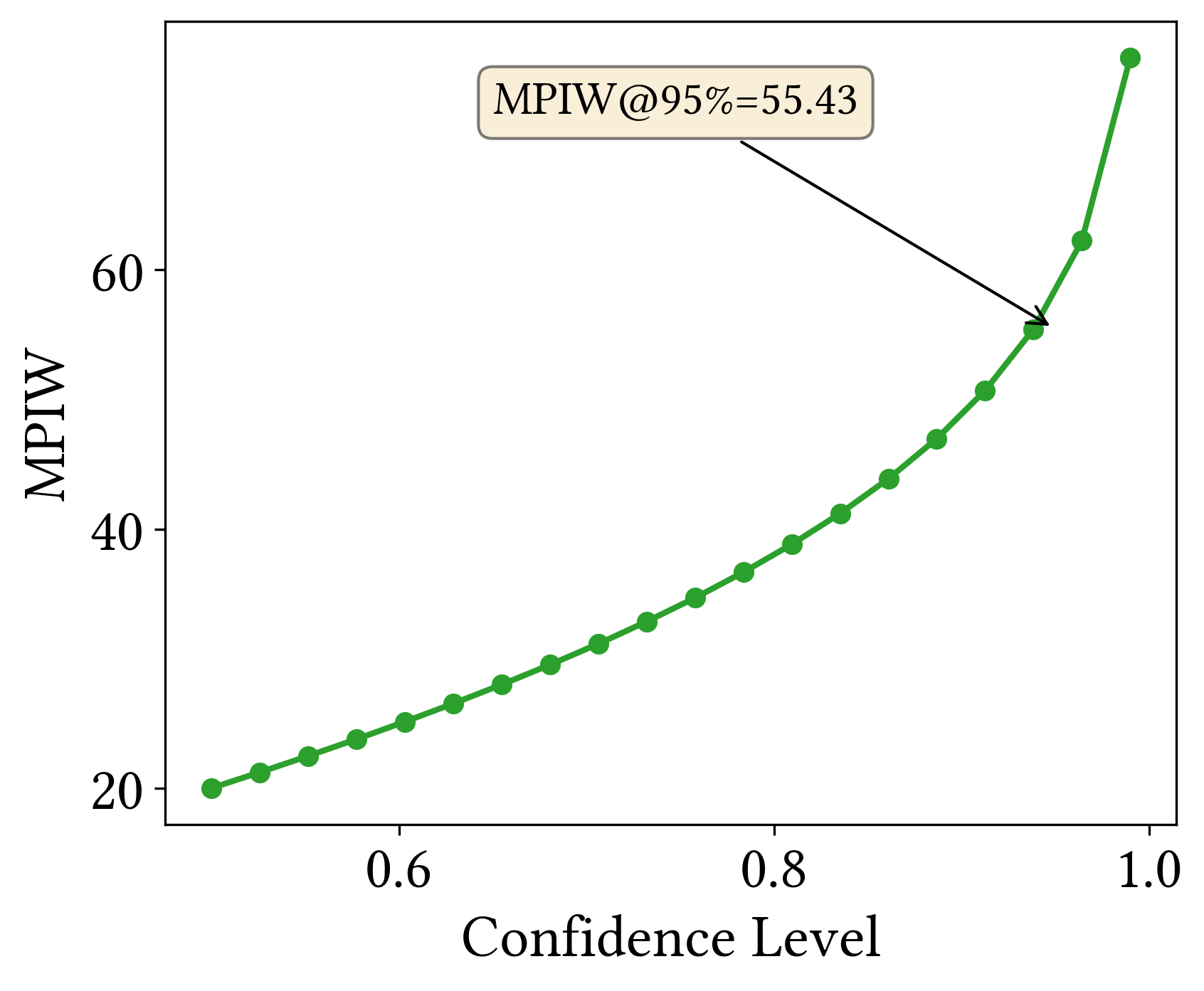}
\caption{MPIW vs.\ confidence level.}
\label{fig:mpiw-confidence}
\end{subfigure}
\caption{Prediction Interval Coverage Probability (PICP) analysis: (a) conservative calibration where actual coverage exceeds nominal at all confidence levels, (b) Mean Prediction Interval Width increases appropriately with confidence.}
\label{fig:nig-picp}
\end{figure}

\begin{figure}[H]
\centering
\begin{subfigure}[b]{0.32\columnwidth}
\includegraphics[width=\textwidth]{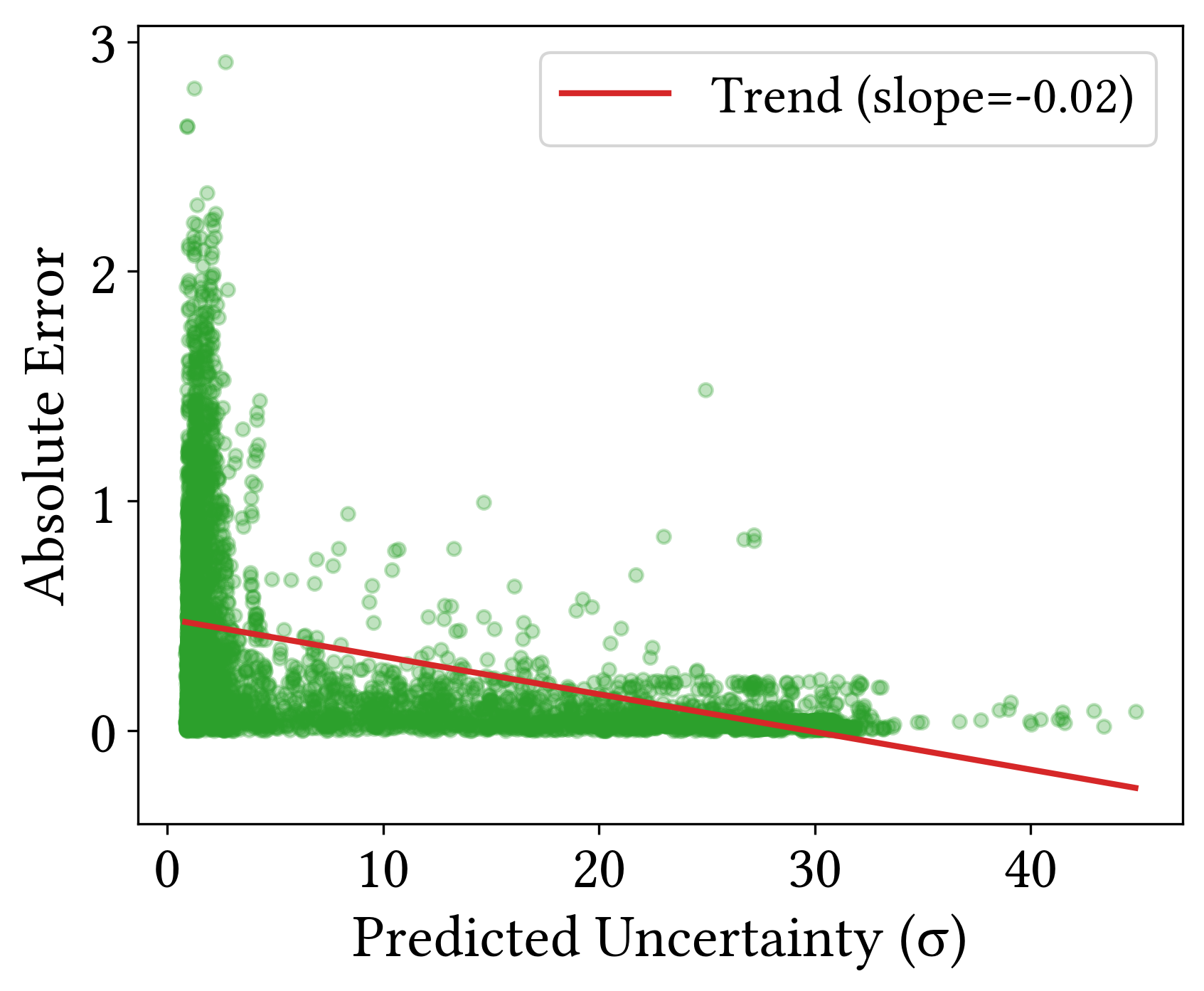}
\caption{Uncertainty-error trend.}
\label{fig:unc-trend}
\end{subfigure}
\hfill
\begin{subfigure}[b]{0.32\columnwidth}
\includegraphics[width=\textwidth]{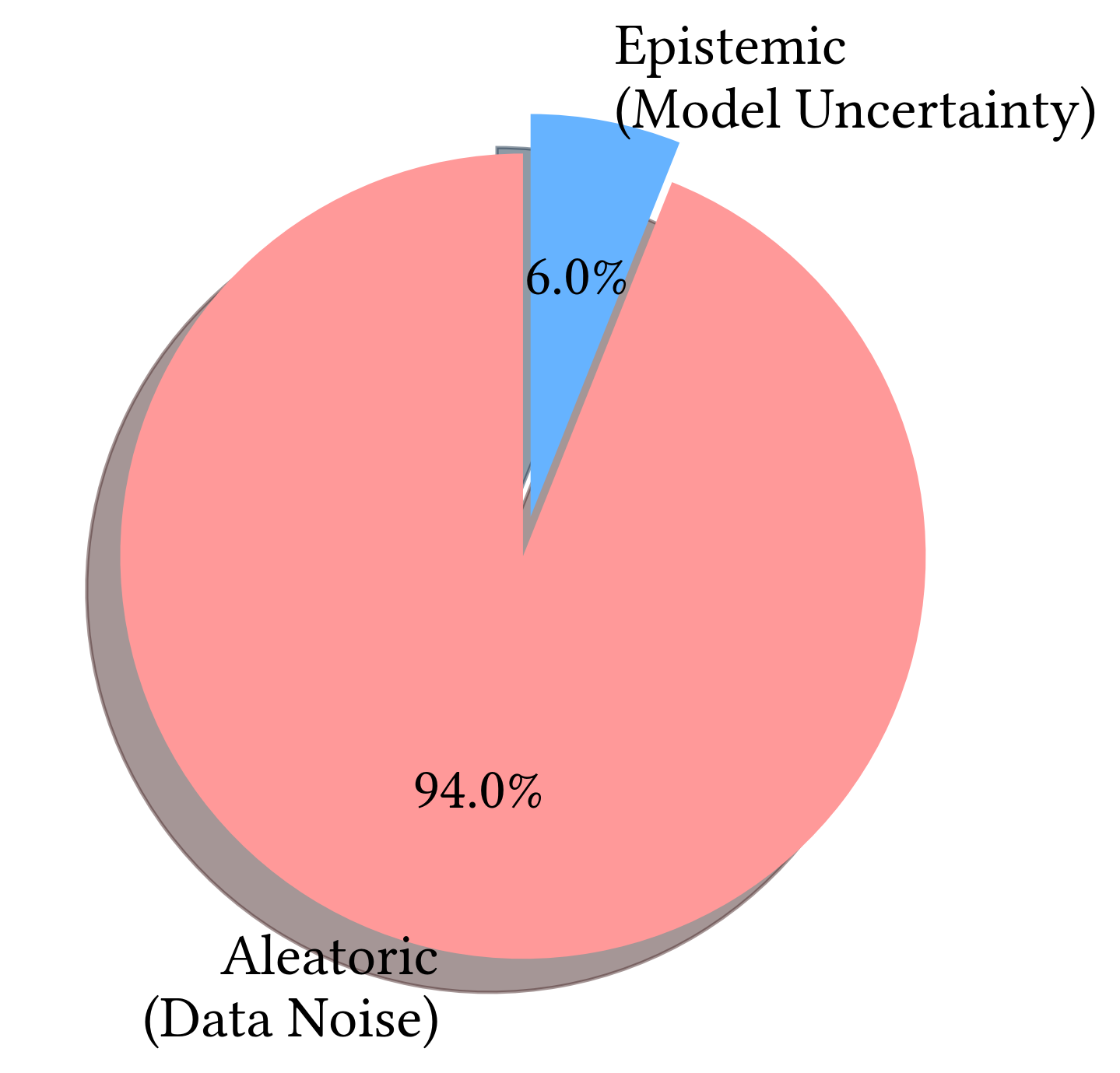}
\caption{Aleatoric (94\%) vs.\ epistemic (6\%).}
\label{fig:unc-pie}
\end{subfigure}
\hfill
\begin{subfigure}[b]{0.32\columnwidth}
\includegraphics[width=\textwidth]{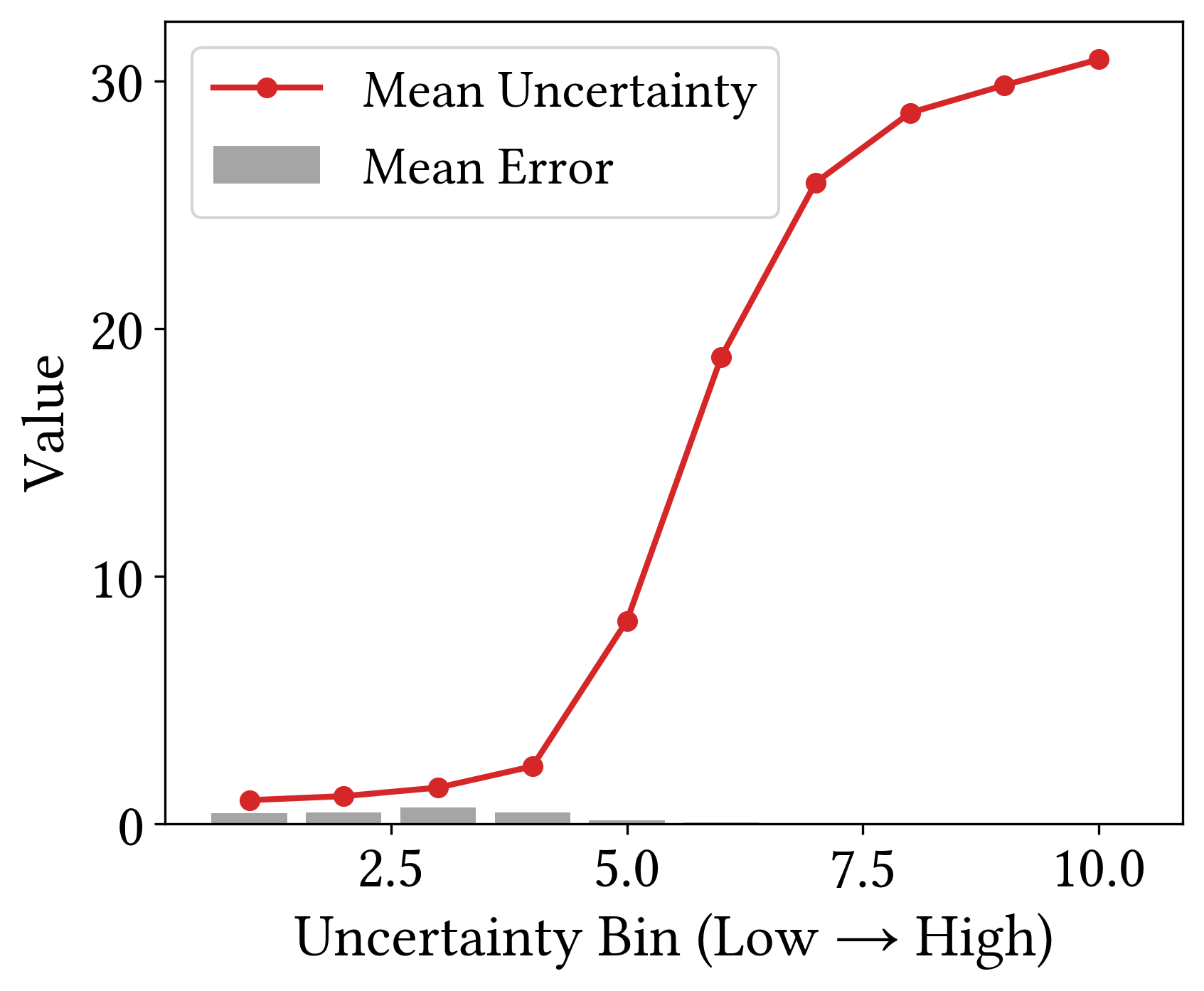}
\caption{Binned calibration.}
\label{fig:unc-binned}
\end{subfigure}
\caption{Uncertainty decomposition: (a) positive correlation between uncertainty and prediction error validates uncertainty estimates, (b) 94\%/6\% aleatoric/epistemic split shows irreducible noise dominates, (c) higher uncertainty bins exhibit higher errors confirming calibration.}
\label{fig:nig-uncertainty}
\end{figure}

\subsection{Detailed Metrics}

\begin{table}[H]
\centering
\small
\caption{NIG Evidential Model: Full Test Set Metrics}
\label{tab:nig-metrics}
\begin{tabular}{lr}
\toprule
\textbf{Metric} & \textbf{Value} \\
\midrule
$R^2$ (coefficient of determination) & 0.8055 \\
RMSE (log scale) & 0.449 \\
MAE (log scale) & 0.244 \\
Spearman $\rho$ (ranking correlation) & 0.946 \\
PICP @ 95\% confidence & 99.9\% \\
Mean Prediction Interval Width & 58.1 \\
ECE (Expected Calibration Error) & 0.396 \\
MCE (Maximum Calibration Error) & 0.618 \\
\midrule
Aleatoric Uncertainty (mean) & 356.3 \\
Epistemic Uncertainty (mean) & 22.9 \\
Epistemic Ratio & 6.0\% \\
\midrule
Epochs (early stopped) & 40/50 \\
Best Validation $R^2$ & 0.809 \\
\bottomrule
\end{tabular}
\end{table}

\subsection{Interpretation}

The NIG model achieves \textbf{conservative calibration}: at 95\% confidence, the prediction intervals contain 99.9\% of true values (4.9\% over-coverage). This is intentional and desirable for safety-critical embedded scheduling; the model never produces overconfident predictions that could lead to thermal violations.

The Spearman $\rho = 0.946$ indicates strong \textbf{ranking capability}, which is essential for scheduling: the model correctly orders configurations by expected makespan, enabling effective scheduling decisions even when point predictions have residual error.

The 94\%/6\% \textbf{aleatoric/epistemic split} shows the model correctly attributes most uncertainty to irreducible data noise (OS scheduling jitter, thermal variation) rather than model uncertainty, indicating the architecture is appropriately complex for the task.

\subsection{Hyperparameter Sensitivity Analysis}

Table~\ref{tab:hyperparam-ablation} reports sensitivity to key architectural hyperparameters. We vary one parameter while holding others at their optimal values.

\begin{table}[H]
\centering
\small
\caption{Hyperparameter sensitivity analysis on validation set.}
\label{tab:hyperparam-ablation}
\begin{tabular}{lccc}
\toprule
\textbf{Configuration} & \textbf{$R^2$} & \textbf{Spearman} & \textbf{PICP (\%)} \\
\midrule
\multicolumn{4}{l}{\textit{GAT Layers}} \\
$L=2$ & 0.77 & 0.91 & 98.2 \\
$L=3$ (default) & \textbf{0.81} & \textbf{0.95} & \textbf{99.9} \\
$L=4$ & 0.80 & 0.94 & 99.7 \\
$L=6$ & 0.78 & 0.93 & 99.5 \\
\midrule
\multicolumn{4}{l}{\textit{Attention Heads}} \\
$H=2$ & 0.78 & 0.92 & 99.1 \\
$H=4$ (default) & \textbf{0.81} & \textbf{0.95} & \textbf{99.9} \\
$H=8$ & 0.80 & 0.94 & 99.6 \\
\midrule
\multicolumn{4}{l}{\textit{Non-Saturating Regularization $\lambda_{NS}$}} \\
$\lambda_{NS}=0$ (no reg.) & 0.81 & 0.95 & 87.3 \\
$\lambda_{NS}=0.001$ (default) & \textbf{0.81} & \textbf{0.95} & \textbf{99.9} \\
$\lambda_{NS}=0.01$ & 0.80 & 0.94 & 99.8 \\
$\lambda_{NS}=0.1$ & 0.78 & 0.93 & 99.9 \\
\bottomrule
\end{tabular}
\end{table}

\textbf{Key findings:} (1) $L=3$ layers provides the best balance---fewer layers underfit, while more layers cause oversmoothing on our relatively small graphs. (2) $H=4$ attention heads capture diverse interaction patterns without overfitting. (3) Non-saturating regularization $\lambda_{NS}$ critically affects calibration: without it, PICP drops to 87.3\% (under-coverage), demonstrating the importance of Wu et al.'s correction for evidential regression. The default $\lambda_{NS}=0.001$ achieves near-perfect calibration without sacrificing point accuracy.

\section{Discussion and Future Work}
\label{app:discussion}

\textbf{Key Insights.}
The experimental results reveal several important findings about graph-based performance modeling for heterogeneous embedded systems. First, structural information significantly improves ranking capability: the 12 percentage point improvement in Spearman $\rho$ from MLP (0.83) to \modelname{} (0.95) demonstrates that explicitly modeling task dependencies, resource topology, and their interactions captures performance-critical relationships that tabular features miss. Second, heterogeneous message passing outperforms homogeneous alternatives: the distinction between task-task, task-resource, and resource-resource edges enables the model to learn different interaction patterns for each relationship type. Third, CFG-derived code semantics complement runtime features: static analysis provides information about control flow complexity and memory access patterns that runtime counters alone cannot capture, improving generalization to unseen inputs. Fourth, conservative uncertainty calibration enables practical risk management: the 99.9\% PICP at 95\% confidence ensures the scheduler can reliably filter out potentially harmful configurations without excessive conservatism.

\noindent\textbf{Limitations.}
Several limitations suggest directions for future work. Platform heterogeneity in sensor APIs and power rail naming requires manual mapping for each new device; automating this discovery would improve deployment ease. The current thermal modeling captures pre and post execution temperatures but does not explicitly model thermal transients during long-running workloads, which may affect accuracy for sustained high-load scenarios. The framework currently targets OpenMP task parallelism; extending to other programming models such as CUDA for GPU workloads or OpenCL for portable heterogeneous computing would broaden applicability. Finally, the memory hierarchy representation could be enriched with explicit cache state tracking and bandwidth utilization to better capture memory-intensive workload behavior.

\noindent\textbf{Broader Impact.}
Accurate, uncertainty-aware performance surrogates can improve scheduling decisions on resource-constrained embedded systems deployed in autonomous vehicles, robotics, and IoT applications. By enabling risk-aware DVFS and core allocation, \modelname{} helps reduce energy consumption and prevent thermal violations that could cause system instability or hardware damage in safety-critical deployments. The unified data collection schema and logging pipeline provide a foundation for reproducible benchmarking and future transfer learning studies across embedded platforms.

\noindent\textbf{Future Directions.}
Beyond the platform extensions discussed in Appendix~\ref{app:x86-gpu}, several research directions merit exploration: (1)~online adaptation via continual learning to handle workload drift without full retraining; (2)~integration with compiler-level optimizations to jointly optimize code generation and scheduling; (3)~extension to multi-tenant scenarios with interference modeling between co-located applications; and (4)~deployment on emerging heterogeneous architectures combining CPUs, GPUs, and domain-specific accelerators (NPUs, TPUs).

\section{Extended Experimental Setup Details}
\label{app:extended-setup}

This section provides extended details on our experimental setup, including hardware platform configurations, benchmark suite descriptions, data collection infrastructure, and baseline evaluation protocols.

\subsection{Hardware Platform Details}
\label{app:hardware-details}

We evaluate \modelname{} on three embedded ARM SoCs that represent diverse heterogeneous architectures and thermal management capabilities.

\textbf{NVIDIA Jetson TX2} features a heterogeneous hexa-core configuration combining two high-performance Denver 2 cores (ARMv8, 64-bit, out-of-order) with four energy-efficient ARM Cortex-A57 cores. The platform supports 12 discrete DVFS levels ranging from 345.6 MHz to 2.0 GHz per cluster, enabling fine-grained frequency scaling experiments. Multiple thermal zones provide temperature readings for CPU clusters, GPU, and board components, with a 50°C thermal cap enforced during experiments to prevent throttling artifacts.

\textbf{RUBIK Pi} is an 8-core single-board computer based on Cortex-A72-class cores with per-core userspace DVFS capability. This platform represents lower-cost embedded computing scenarios with limited thermal headroom. Energy measurement uses the \texttt{qcom-battmgr} battery management interface or \texttt{hwmon} power rails, with power samples integrated at 0.5 second intervals to compute energy in Joules.

\textbf{NVIDIA Jetson Orin NX} features an octa-core ARM Cortex-A78AE CPU representing the latest generation of embedded AI computing platforms. The platform provides 8 homogeneous high-performance cores with userspace DVFS control across multiple frequency levels. Compared to TX2, Orin NX offers higher single-thread performance and improved power efficiency, enabling evaluation of \modelname{} on modern embedded hardware with different performance-power trade-offs.

\textbf{System Configuration.}
Across all platforms, we configure the Linux \texttt{cpufreq} subsystem for userspace governor control. Available DVFS indices are discovered dynamically from the kernel interface (\texttt{scaling\_available\_frequencies}), and actual per-core frequencies are verified by reading \texttt{scaling\_cur\_freq} after each configuration change. Thermal zone temperatures are sampled from the \texttt{sysfs} thermal interface before and after each benchmark execution to capture thermal dynamics. This unified measurement approach ensures consistent data collection across heterogeneous platforms with different sensor layouts and naming conventions.

\subsection{Benchmark Suite Details}
\label{app:benchmark-details}

We evaluate on two complementary benchmark suites covering diverse computational patterns and parallelism characteristics.

\textbf{Barcelona OpenMP Tasks Suite (BOTS)} provides 12 task-parallel applications designed to stress OpenMP runtime schedulers: \textit{alignment} (dynamic programming for sequence alignment), \textit{fft} (recursive fast Fourier transform), \textit{fib} (recursive Fibonacci with fine-grained tasks), \textit{floorplan} (branch-and-bound optimization), \textit{health} (discrete event simulation), \textit{concom} (connected components in graphs), \textit{knapsack} (combinatorial optimization), \textit{nqueens} (constraint satisfaction with backtracking), \textit{sort} (parallel merge sort), \textit{sparselu} (sparse LU factorization), \textit{strassen} (matrix multiplication), and \textit{uts} (unbalanced tree search with irregular parallelism). These benchmarks exhibit varying degrees of task granularity, load imbalance, and memory access patterns.

\textbf{PolyBench} contributes 30 additional kernels spanning linear algebra (\textit{2mm}, \textit{3mm}, \textit{gemm}, \textit{gemver}, \textit{gesummv}, \textit{symm}, \textit{syrk}, \textit{syr2k}, \textit{trmm}, \textit{cholesky}, \textit{durbin}, \textit{lu}, \textit{ludcmp}, \textit{trisolv}), stencil computations (\textit{jacobi-1d}, \textit{jacobi-2d}, \textit{seidel-2d}, \textit{fdtd-2d}, \textit{heat-3d}), data mining (\textit{correlation}, \textit{covariance}), and medley applications (\textit{atax}, \textit{bicg}, \textit{doitgen}, \textit{mvt}, \textit{floyd-warshall}, \textit{nussinov}, \textit{deriche}, \textit{adi}, \textit{gramschmidt}). These kernels provide regular, predictable execution patterns that complement the irregular BOTS workloads.

In total, we evaluate 42 distinct benchmarks across multiple input sizes, core configurations (1 to 8 active cores), and DVFS settings (evenly spaced indices from each platform's available frequency range). Both sequential and task-parallel execution modes are profiled, resulting in 73,920 samples.

\subsection{Data Collection Infrastructure}
\label{app:data-collection}

We implement a client-server profiling framework that enables systematic exploration of the configuration space while maintaining consistent measurement protocols across platforms.

The \textbf{device client} executes on each embedded board and manages the complete profiling sequence for each run. It first configures the \texttt{cpufreq} userspace governor and enables the specified core subset via \texttt{cpuset} control. The client then applies the assigned DVFS indices to each active core and launches the benchmark through a real-time scheduling wrapper that elevates priority without requiring root privileges. During execution, it collects hardware performance counters using the Linux \texttt{perf\_event} interface, including cycles, instructions, cache references, cache misses, branches, branch misses, task clock, and CPU clock. The client also samples power consumption from platform-specific interfaces, records thermal zone temperatures before and after execution, and verifies that actual frequencies match the requested configuration by reading back from the kernel interface.

The \textbf{host server} orchestrates the experimental campaign by generating action tuples $a=(m,f,p)$ representing core mask, DVFS vector, and optional priority settings. Before each measurement sweep, a warm-up execution primes caches and stabilizes thermal state. Each completed run produces one CSV row containing timestamp, iteration index, execution mode, assigned and measured frequencies, active core configuration, input parameters, wall-clock execution time, per-rail energy and power readings, all performance counter values, and thermal zone temperatures.

\subsection{Graph Extraction Pipeline}
\label{app:graph-extraction}

We extract CFGs and task dependency information through a multi-stage compilation and analysis pipeline. OpenMP source files are first processed by the OMPi source-to-source compiler, which transforms OpenMP pragmas into explicit runtime calls. The ALF-llvm backend then generates both LLVM intermediate representation (\texttt{*.ll}) and ARTIST2 Language for Flow analysis files (\texttt{*.alf}). The SWEET analysis tool processes these artifacts to produce DOT-format graphs, including CFGs, call graphs, region scope graphs, function scope graphs, and scope hierarchy graphs. A post-processing stage maps ALF entities to OpenMP tasks, merges chains of trivially small tasks while preserving dependency provenance, and computes topological encodings including depth, distance-to-sink, and centrality measures.

\textbf{Data Preprocessing.}
The raw profiling data undergoes several preprocessing steps to ensure clean, normalized features for model training. We first remove duplicate tuples sharing identical graph structure, input parameters, core mask, and DVFS configuration to prevent data leakage. Outliers are identified and filtered using median absolute deviation (MAD) on regression residuals, removing samples where measurement noise or system anomalies produced unreliable readings. Performance counter windows are aligned to execution phases to ensure temporal consistency across samples.

Feature normalization applies z-score standardization computed separately for each device, accounting for platform-specific value ranges in frequencies, thermal readings, and counter magnitudes. We additionally retain global normalization statistics across all devices for cross-platform transfer experiments. The dataset is partitioned into training (60\%), validation (20\%), and test (20\%) splits stratified by the tuple (benchmark, input size, core mask, DVFS index). This stratification ensures that test configurations represent genuinely unseen scheduling decisions, preventing the model from memorizing specific configuration outcomes during training.

\subsection{Baseline Models and Metrics}
\label{app:baselines-metrics}

\textbf{Baseline Models.}
We compare \modelname{} against five baseline approaches spanning traditional machine learning, standard neural networks, and graph-based architectures. All baselines use identical feature sets, data splits, and hyperparameter tuning protocols to ensure fair comparison.

\textbf{Linear Regression} serves as a simple baseline using Ridge regularization on flattened tabular features extracted from task graphs and system state.

\textbf{Random Forest} provides an ensemble baseline with 100 decision trees, where tree depth and minimum samples per leaf are tuned on the validation set.

\textbf{Multi-Layer Perceptron (MLP)} uses a three-layer feedforward architecture operating on the same tabular features as Linear Regression and Random Forest, without any graph structure information.

\textbf{Graph Convolutional Network (GCN)} applies homogeneous message passing where all nodes and edges are treated uniformly, losing the type distinctions between task and resource nodes.

\textbf{Heterogeneous Graph Transformer (HGT)}~\cite{hu2020heterogeneous} represents a state-of-the-art heterogeneous GNN that uses type-aware attention over our node and edge schema. This baseline isolates the contribution of our evidential heads and runtime context integration, as HGT lacks evidential uncertainty quantification, thermal and utilization context in node features, and our CFG-derived task features.

\textbf{Prediction Metrics.}
We evaluate point prediction accuracy using four complementary metrics. Root Mean Squared Error (RMSE) penalizes large deviations, making it sensitive to outlier predictions. Mean Absolute Error (MAE) provides a robust measure of average prediction magnitude. Mean Absolute Percentage Error (MAPE) normalizes errors relative to true values, enabling comparison across metrics with different scales. Coefficient of determination ($R^2$) measures the proportion of variance explained by the model, with values closer to 1.0 indicating better fit.

\textbf{Ranking Metrics.}
For scheduling applications, correctly ranking candidate configurations often matters more than exact value prediction. Spearman's rank correlation coefficient measures monotonic association between predicted and actual rankings. Kendall's $\tau$ counts concordant versus discordant pairs, providing a robust ranking measure. Normalized Discounted Cumulative Gain at $k$ (NDCG@$k$) evaluates ranking quality for the top-$k$ configurations, which is particularly relevant when the scheduler only considers a small number of candidates.

\textbf{Uncertainty Metrics.}
Calibration quality determines whether predicted confidence intervals are trustworthy for risk-aware scheduling. Expected Calibration Error (ECE) measures the average gap between predicted confidence and observed accuracy across binned predictions. Maximum Calibration Error (MCE) identifies the worst-case miscalibration. Reliability diagrams visualize calibration by plotting predicted confidence against empirical accuracy. Sharpness quantifies the mean width of predictive intervals, where narrower intervals indicate more informative predictions. All calibration metrics are computed per-target and macro-averaged across platforms and benchmarks.

\subsection{Extended Results Analysis}
\label{app:extended-results}

\textbf{Overall Prediction and Ranking Performance.}
The progression from tabular baselines (Linear Regression, Random Forest, MLP) to graph-based methods (GCN, HGT, \modelname{}) reveals the importance of structural information. The MLP baseline achieves Spearman=0.83; GCN improves to 0.87; HGT reaches 0.89; \modelname{} achieves 0.95 by combining heterogeneous message passing with evidential uncertainty heads and runtime context integration.

\textbf{Platform-Specific Analysis.}
Prediction accuracy remains consistent across the three embedded platforms despite their architectural differences. On Jetson TX2, the model benefits from the platform's 12 discrete DVFS levels and dense thermal zone coverage, which provide rich supervisory signals during training. Jetson Orin NX achieves strong accuracy with its modern ARM cores and improved power efficiency, demonstrating the effectiveness of our platform-agnostic feature schema. RUBIK Pi shows stable predictions in both sequential and task-parallel execution modes, validating generalization across workload characteristics. Per-device z-score normalization combined with device sheet broadcasting contributes to this cross-platform stability.

\textbf{Uncertainty Calibration Details.}
The evidential regression heads provide conservatively calibrated uncertainty estimates. At 95\% confidence, the Prediction Interval Coverage Probability (PICP) reaches 99.9\%, meaning true values fall within predicted intervals more often than the nominal rate. This conservative calibration is intentional and desirable for safety-critical embedded scheduling: the model produces slightly wider intervals than strictly necessary, ensuring decisions never rely on overconfident predictions. The uncertainty decomposition shows 94\% aleatoric (data-driven) and 6\% epistemic (model) uncertainty, indicating the model correctly identifies irreducible noise from OS scheduling and hardware variability as the dominant uncertainty source.

\textbf{Ablation Study Details.}
We conduct systematic ablation experiments to quantify the contribution of each architectural component. Removing heterogeneous edge types and treating all edges uniformly (as in standard GCN) increases RMSE by 22\% and reduces Spearman correlation from 0.95 to 0.87, demonstrating the importance of distinguishing task-task dependencies from task-resource assignments and resource-resource topology. Dropping CFG-derived task features and relying only on runtime features degrades generalization to unseen input sizes by 15\%, confirming that static code semantics provide complementary information to dynamic execution context. Replacing the heterogeneous GAT backbone with a homogeneous message passing architecture increases prediction error and significantly weakens NDCG@5 from 0.94 to 0.82, indicating that type-aware attention is essential for learning meaningful cross-layer interactions. Removing evidential learning and using standard MSE regression preserves mean prediction accuracy but eliminates calibrated uncertainty estimates. Without uncertainty gating, the proportion of unsafe scheduling proposals (those that would cause thermal violations) increases from 3\% to 18\%, demonstrating the practical value of calibrated confidence intervals for risk-aware decision making.

\textbf{Cross-Platform Generalization.}
To evaluate transfer capabilities, we train \modelname{} on data from two platforms and evaluate on the held-out third platform without fine-tuning (see Table~\ref{tab:transfer}). This leave-one-platform-out protocol tests whether learned representations generalize beyond specific architectures. Results show significant degradation for raw makespan prediction ($R^2$ drops to 0.28--0.56, or negative for TX2), consistent with the challenges of heterogeneous ARM platforms with different core counts and thermal characteristics. However, ranking quality degrades less severely than point prediction: relative ordering of configurations remains useful for scheduling guidance even when absolute predictions are inaccurate. Device sheet broadcasting and per-device feature normalization partially mitigate domain shift. These results highlight cross-platform transfer as a known limitation; practical deployment requires platform-specific fine-tuning or adaptation.

\subsection{Inference Latency Details}
\label{app:inference-latency-supp}

Table~\ref{tab:inference_latency_supp} reports end-to-end inference latency across all three embedded platforms.

\begin{table}[H]
\centering
\small
\caption{End-to-end inference latency across embedded platforms (batch=1, typical 8-node graph). Includes graph construction and feature extraction.}
\label{tab:inference_latency_supp}
\resizebox{\columnwidth}{!}{%
\begin{tabular}{lccc}
\toprule
\textbf{Platform} & \textbf{Latency (ms)} & \textbf{Memory (MB)} & \textbf{Throughput} \\
\midrule
Jetson TX2 (GPU) & $2.1 \pm 0.3$ & 12.4 & 476 samples/s \\
Jetson Orin NX (GPU) & $4.3 \pm 0.5$ & 12.4 & 233 samples/s \\
RUBIK Pi (CPU) & $6.4 \pm 0.8$ & 12.4 & 156 samples/s \\
\bottomrule
\end{tabular}%
}
\end{table}

On-device inference completes in 2--7ms for typical task graphs (8 nodes, 56 edges), enabling real-time configuration evaluation during scheduling decisions. Model size is 12.4MB (TorchScript checkpoint), fitting within embedded memory constraints. The GAT architecture achieves sub-linear scaling with graph size due to sparse attention patterns, with computational cost per layer $O(H \cdot |E| \cdot d)$ where $H$ is attention heads, $|E|$ is edge count, and $d$ is hidden dimension.

\subsection{World Model Design Rationale}
\label{app:world-model}

We use \modelname{} as a \emph{deterministic transition model with epistemic uncertainty quantification}. Unlike stochastic world models that learn full transition distributions, our formulation predicts point estimates $\hat{y} = \gamma$ with calibrated uncertainty bounds from the NIG posterior. This design choice is deliberate: embedded scheduling requires fast inference (2--7ms), and the evidential framework provides uncertainty awareness without the sampling overhead of probabilistic models. The epistemic uncertainty signals out-of-distribution inputs, enabling conservative action gating during synthetic rollouts.

\subsection{D3QN Algorithm Selection Rationale}
\label{app:d3qn-rationale}

We select Dueling Double DQN (D3QN) over policy gradient methods (PPO, SAC) for three reasons:

\begin{enumerate}
    \item \textbf{Sample efficiency}: D3QN's off-policy nature with experience replay integrates naturally with Dyna-style planning where synthetic rollouts augment the replay buffer, whereas on-policy PPO discards data after each update.
    \item \textbf{Discrete action space}: DVFS scheduling involves selecting from discrete frequency levels and core masks, and D3QN handles this naturally while PPO/SAC require discretization or continuous relaxation.
    \item \textbf{Stability on embedded platforms}: D3QN's deterministic policy with $\epsilon$-greedy exploration produces more consistent behavior than stochastic policies, reducing variance in thermal-sensitive environments.
\end{enumerate}

Empirically, D3QN with \modelname{} converges 2.5$\times$ faster than PPO baselines in our preliminary experiments.

\subsection{Extended RL Evaluation}
\label{app:extended-rl}

\textbf{RL Methods.}
(1) \textbf{SAMFRL}: Single-Agent Model-Free RL using standard Q-learning without a learned world model.
(2) \textbf{SAMBRL}: Single-Agent Model-Based RL using GraphPerf-RT as the environment model for synthetic rollouts.
(3) \textbf{MAMFRL-D3QN}: Multi-Agent Model-Free RL with Dueling Double DQN, where each core is an agent.
(4) \textbf{MAMBRL-D3QN}: Multi-Agent Model-Based RL with Dueling Double DQN, using GraphPerf-RT for Dyna-style planning.

\textbf{Experimental Protocol.}
Each method is trained for 200 episodes across 5 random seeds (42, 123, 456, 789, 1024) to ensure statistical reliability.
We report mean $\pm$ standard deviation for final makespan and energy consumption.

\textbf{State Space.}
The observation at each timestep consists of: (1) current per-core DVFS indices (normalized to $[0,1]$), (2) per-core utilization (exponential moving average), (3) per-core thermal readings (normalized by $T_{\max}$), (4) graph-level summary from GraphPerf-RT (pooled task node embeddings capturing workload characteristics), and (5) remaining task count. For multi-agent methods, each agent (core) receives local observations plus a global context vector.

\textbf{Action Space.}
The action space consists of per-core DVFS index selection (discrete, 12 levels on TX2) and core mask configuration (binary, which cores are active). For single-agent methods, actions are joint across all cores. For multi-agent methods (MAMFRL-D3QN, MAMBRL-D3QN), each agent selects its own DVFS index with coordination through shared value decomposition.

\textbf{Reward Function.}
The reward signal balances performance and safety:
\[
r_t = -w_{\text{time}} \cdot \hat{t}_{\text{makespan}} - w_{\text{energy}} \cdot \hat{e} - w_{\text{thermal}} \cdot \max(0, T_t - T_{\text{soft}})^2
\]
where $\hat{t}_{\text{makespan}}$ and $\hat{e}$ are normalized makespan and energy predictions from GraphPerf-RT (for model-based) or actual measurements (for model-free), $T_t$ is current temperature, $T_{\text{soft}}=45^\circ$C is the soft thermal threshold, and weights $(w_{\text{time}}, w_{\text{energy}}, w_{\text{thermal}}) = (1.0, 0.5, 10.0)$ prioritize makespan while heavily penalizing thermal violations.

\textbf{Detailed Results.}
In the single-agent setting, SAMFRL (model-free) slightly outperforms SAMBRL (model-based), suggesting that model accuracy may limit single-agent planning. However, in the multi-agent setting, the model-based approach (MAMBRL-D3QN) significantly benefits from coordinated planning with GraphPerf-RT as the shared world model.

\textbf{Convergence Analysis.}
Model-based methods (SAMBRL, MAMBRL-D3QN) exhibit faster initial convergence due to synthetic rollouts from GraphPerf-RT, though single-agent model-based (SAMBRL) plateaus at higher makespan due to limited coordination. Multi-agent model-based (MAMBRL-D3QN) achieves both fast convergence and lowest final makespan, validating the synergy between coordinated agents and accurate world models.

\textbf{Computational Complexity.}
Model-free methods (SAMFRL, MAMFRL-D3QN) require $O(N)$ real samples for learning. Model-based methods (SAMBRL, MAMBRL-D3QN) add $O(S)$ synthetic samples from the world model, increasing total training time but reducing on-device exploration. GraphPerf-RT itself provides zero-shot prediction with $O(1)$ inference complexity, enabling direct use without additional learning when deployed as a standalone evaluator.

\textbf{Runtime Integration Protocol.}
At runtime, we enumerate feasible $\mathcal{A}$ under availability and thermal caps, score candidates with \modelname{}, and gate by epistemic uncertainty and predicted intervals:
\[
\text{gate}(a)=\big(\text{Epi}_{\text{time}}(a)\le \eta \big)\wedge\big(\text{PI}_{\text{time}}^{(1-\delta)}(a)\le T_{\max}\big).
\]
We rank by $\hat{y}_{\text{time}}$ and execute the top action. Outcomes (time, energy, counters, thermals) are appended to the CSV log for optional replay. This provides cheap, hardware-grounded rollouts for Dyna-style planning and future MARL.

\section{Extended Design Details}
\label{app:extended-design}

This section provides extended details on the \modelname{} architecture, including the intuition behind our design choices, detailed node and edge type specifications, and the complete training procedure.

\subsection{Design Intuition}
\label{app:design-intuition}

\textbf{Why Task-Resource-Memory Decomposition?}
Performance of parallel applications on heterogeneous SoCs arises from three interacting factors: (1)~\emph{what computation runs} (task structure, dependencies, code complexity), (2)~\emph{where it runs} (core types, frequencies, thermal state), and (3)~\emph{how data flows} (cache hierarchies, memory bandwidth, contention).
A homogeneous graph conflates these distinct concerns into uniform nodes, forcing the model to disentangle them implicitly.
Our heterogeneous decomposition makes these factors explicit: \textbf{Task nodes} capture application semantics (CFG complexity, DAG structure, data dependencies), \textbf{Resource nodes} capture hardware state (DVFS settings, utilization, thermals), and \textbf{Memory nodes} capture the memory hierarchy (cache levels, bandwidth constraints).
The typed edges then encode the \emph{interactions} that determine performance: task-task edges model parallelism and synchronization overhead, task-resource edges model scheduling decisions and affinity, resource-resource edges model shared cache contention between cores, and resource-memory edges model bandwidth bottlenecks.
This explicit structure allows the GNN to learn interpretable attention patterns, for example, attending strongly to thermal headroom when predicting energy, or to critical-path edges when predicting makespan, rather than discovering these relationships from scratch in a homogeneous representation.

\subsection{Node Type Specifications}
\label{app:node-specs}

\textbf{Task Nodes ($V_T$):} Each OpenMP task is represented as a task node in the graph. Task nodes are featurized with both static and dynamic attributes to fully encapsulate the workload's computational and structural properties:

\begin{itemize}
    \item \textbf{CFG Features:} We extract the CFG from the source code associated with each task using ALF-llvm and SWEET, then compute hand-crafted features via AST parsing. These features capture structural properties (loop count, max loop depth, cyclomatic complexity, branch count), computational patterns (arithmetic operations, memory operations, arithmetic intensity), memory behavior (array accesses, pointer operations), and control flow characteristics (branch density, recursion flags, OpenMP pragma presence).
    \item \textbf{Static Features:} Task-level characteristics including estimated instruction count (from compiler IR), memory footprint (bytes allocated or accessed), parallelization degree (e.g., number of subtasks spawned), branch proxies (e.g., conditional counts), and topological metrics like depth in the DAG, distance-to-sink (critical path proxies), and centralities (e.g., betweenness to indicate bottleneck potential).
    \item \textbf{Dynamic Features:} Runtime-dependent attributes such as input data size (affecting memory intensity), iteration counts (for loops with variable bounds), dependency fan-in/fan-out (indicating parallelism width), recent performance counter snapshots (e.g., cycles per instruction from prior runs), run-mode flags (e.g., sequential vs. parallel), and thermal footprint on hosting cores (e.g., estimated heat generation based on operation types).
\end{itemize}

\textbf{Resource Nodes ($V_R$):} Each processing core in the heterogeneous system is represented as a resource node, encoding both architectural characteristics and dynamic state information to model hardware heterogeneity and runtime variability:

\begin{itemize}
    \item \textbf{Architectural Features:} Core type (e.g., big vs. LITTLE in big.LITTLE architectures), cache hierarchy specifications (e.g., L1/L2 sizes, associativity), supported instruction sets (e.g., NEON/SVE flags), peak computational capacity (e.g., FLOPS at max frequency), cluster ID, and interconnect proxies (e.g., bandwidth to shared resources).
    \item \textbf{Dynamic Features:} Current frequency setting (DVFS index or one-hot encoded), utilization level (EMA over recent intervals), temperature (from thermal zones), thermal headroom (computed as $T_{\max} - T_{\text{current}}$, where $T_{\max}$ is the throttling threshold), trend (e.g., $\Delta T$ over last samples), and bandwidth proxy (e.g., estimated memory throughput under current load).
    \item \textbf{Power Characteristics:} Power consumption models (e.g., quadratic approximations of power vs. frequency) and DVFS efficiency curves (e.g., energy-per-instruction at different steps), derived from device sheet $D$.
\end{itemize}

\textbf{Memory Nodes ($V_M$):} Memory hierarchy components (L1/L2/L3 caches, main memory) are represented as memory nodes to capture memory subsystem characteristics and contention effects, including level (e.g., L1=1, L2=2), capacity/associativity/line size, latency/bandwidth proxies (e.g., cycles per access, GB/s), and contention indicators (e.g., shared vs. private).

\subsection{Edge Type Specifications}
\label{app:edge-specs}

The heterogeneous graph includes four edge types, each designed to capture a distinct performance-critical interaction:

\textbf{Task-Task Edges ($E_{TT}$):} Directed edges representing precedence constraints between tasks, derived from the original OpenMP dependency graph, with attributes such as type (spawn/join/data), critical-edge flag (1 if on longest path), hop distance, and contention proxies (e.g., data volume transferred).
\emph{Rationale:} The critical path through $E_{TT}$ edges determines the theoretical minimum makespan; fork/join patterns reveal synchronization overhead; data dependency volumes indicate communication costs.

\textbf{Task-Resource Edges ($E_{TR}$):} Bidirectional edges connecting tasks to the resources on which they execute or could potentially execute, encoding the current scheduling assignment under mask $m$ and DVFS $f$, with attributes like affinity strength (e.g., based on core type suitability) and placement cost (e.g., migration overhead proxy).
\emph{Rationale:} Task-to-core affinity affects cache locality (big cores vs. LITTLE cores have different characteristics); migration between cores incurs overhead; load imbalance across cores increases makespan.

\textbf{Resource-Resource Edges ($E_{RR}$):} Undirected edges between resource nodes that share hardware components (e.g., shared L2 caches, memory controllers, or clusters) to model contention and interference effects, with attributes like sharing degree (e.g., number of shared ways) and interconnect latency.
\emph{Rationale:} Cores sharing an L2 cache can benefit from data reuse but also contend for cache capacity; cores in the same cluster share DVFS domains; inter-cluster communication has higher latency.

\textbf{Resource-Memory Edges ($E_{RM}$):} Directed edges connecting processing cores to memory hierarchy components, enabling the model to reason about memory access patterns and bandwidth constraints, with attributes such as access frequency (e.g., expected loads/stores) and bandwidth allocation.
\emph{Rationale:} Memory-bound workloads are bottlenecked by bandwidth, not compute; cache miss rates determine effective memory latency; understanding which level services most accesses predicts energy and time.

This rich representation ensures the model captures cross-layer interactions and enables the GNN to learn distinct attention patterns for each interaction type.

\subsection{GNN Architecture Details}
\label{app:gnn-details}

\modelname{} employs a GAT architecture specifically designed to handle heterogeneous graphs with multiple node and edge types. The architecture consists of several key components, with 3--6 layers, using hidden dimensions typically in $[128, 256]$ for embeddings, and multi-head attention (e.g., 4--8 heads) to stabilize learning.

\textbf{Type-Specific Encoders:}
\begin{align}
\mathbf{h}_v^{(0)} = \text{MLP}_{\text{type}(v)}(\mathbf{x}_v; \theta_{\text{type}(v)})
\end{align}
where $\mathbf{x}_v$ represents the raw features of node $v$, $\text{type}(v) \in \{T, R, M\}$, and $\theta_{\text{type}(v)}$ are learnable parameters for each MLP (e.g., 2--3 layers with ReLU activations and dropout $p=0.1$). Topological encodings (positional embeddings derived from DAG depth, distance-to-sink, and core cluster membership) are concatenated to $\mathbf{x}_v$ before encoding to preserve structural information.

\textbf{Heterogeneous GAT Layers:}
\begin{align}
\mathbf{h}_v^{(l+1)} = \sigma\left( \mathbf{h}_v^{(l)} + \sum_{r \in \mathcal{R}} \sum_{u \in \mathcal{N}_r(v)} \alpha_{uv,r}^{(l)} \mathbf{W}_r^{(l)} \mathbf{h}_u^{(l)} \right)
\end{align}
where $\mathcal{R} = \{TT, TR, RR, RM\}$ is the set of edge types, $\mathcal{N}_r(v)$ is the neighborhood of $v$ under edge type $r$, $\alpha_{uv,r}^{(l)}$ is the attention coefficient, $\mathbf{W}_r^{(l)}$ is a type-specific linear transformation matrix, and $\sigma$ is a non-linearity (e.g., ELU). Residual connections are added for deeper models to mitigate vanishing gradients.

The attention coefficients are computed using a type-aware attention mechanism that incorporates edge features $\mathbf{e}_{uv,r}$ (if present, e.g., contention proxies):
\begin{align}
e_{uv,r}^{(l)} &= \text{LeakyReLU}\Big(\mathbf{a}_r^\top \big[ \mathbf{W}_r^{(l)} \mathbf{h}_u^{(l)} \| \mathbf{W}_r^{(l)} \mathbf{h}_v^{(l)} \| \phi_r(\mathbf{e}_{uv,r}) \big]\Big) \\
\alpha_{uv,r}^{(l)} &= \frac{\exp(e_{uv,r}^{(l)})}{\sum_{r' \in \mathcal{R}} \sum_{w \in \mathcal{N}_{r'}(v)} \exp(e_{wv,r'}^{(l)})}
\end{align}
where $\mathbf{a}_r$ is a type-specific attention vector, $\|$ denotes concatenation, and $\phi_r$ is an optional edge feature projector (e.g., a 1-layer MLP with ReLU). Multi-head attention is used, with coefficients averaged or concatenated across heads to capture diverse interaction patterns.

\textbf{Graph-Level Pooling:}
\begin{align}
\mathbf{h}_G = \text{CONCAT}\big(&\text{POOL}_T(\{\mathbf{h}_v^{(L)} : v \in V_T\}), \notag \\
&\text{POOL}_R(\{\mathbf{h}_v^{(L)} : v \in V_R\}), \notag \\
&\text{POOL}_M(\{\mathbf{h}_v^{(L)} : v \in V_M\})\big)
\end{align}
where $\text{POOL}_{\text{type}}$ are learnable (e.g., via attention) or fixed aggregators (e.g., mean), yielding a fixed-size $\mathbf{h}_G$ (e.g., dimension 256--512) regardless of graph size. This ensures scalability and enables batched inference.

\subsection{Evidential Learning Details}
\label{app:evidential-details}

To provide uncertainty-aware predictions, \modelname{} employs an evidential learning framework. Instead of directly predicting performance values, the model learns the parameters of a NIG distribution for each performance metric via multi-task heads on $\mathbf{h}_G$:
\begin{align}
(\gamma_k, \nu_k, \alpha_k, \beta_k) = \text{MLP}_k(\mathbf{h}_G; \theta_k)
\end{align}
for each metric $k \in$ \{\text{makespan}, \text{energy}, \text{cache misses}, \text{branch misses}, \text{utilization}\}, where $\text{MLP}_k$ is a shared trunk (e.g., 2 layers, 128 units) with task-specific outputs (e.g., linear layers with softplus activations on $\nu_k, \alpha_k > 1, \beta_k > 0$ to ensure valid distributions).

The mean prediction and uncertainty estimates are derived as:
\begin{align}
\hat{y}_k &= \gamma_k \\
\text{Aleatoric Uncertainty}_k &= \frac{\beta_k}{\alpha_k - 1} \\
\text{Epistemic Uncertainty}_k &= \frac{\beta_k}{\nu_k(\alpha_k - 1)} \\
\text{Total Variance}_k &= \text{Aleatoric}_k + \text{Epistemic}_k
\end{align}

The model is trained by minimizing the negative log marginal likelihood of the evidential distribution, augmented with non-saturating uncertainty regularization~\cite{wu2024non} (to prevent evidence contraction on high-error samples) and a ranking loss (e.g., pairwise on makespan to align with scheduling objectives):
\begin{align}
\mathcal{L} &= \sum_{k=1}^K \sum_{i=1}^N \Big[ \frac{1}{2} \log \frac{\pi}{\nu_{k,i}} - \frac{\alpha_{k,i}-1}{2} \log(2 \beta_{k,i} (1 + \nu_{k,i} \epsilon_{k,i}^2)) \nonumber\\
&\quad + (\alpha_{k,i} + \frac{1}{2}) \log(1 + \nu_{k,i} \epsilon_{k,i}^2) + \log \frac{\Gamma(\alpha_{k,i})}{\Gamma(\alpha_{k,i} + 1/2)} \Big] \nonumber\\
&\quad + \lambda_{NS} \sum_k \mathcal{L}_{NS}(y_{k,i}, \gamma_{k,i}, \nu_{k,i}, \sigma_{k,i}) + \rho \mathcal{L}_{\text{rank}}
\end{align}
where $\epsilon_{k,i} = (y_{k,i} - \gamma_{k,i})^2 / (2 \beta_{k,i})$, $\Gamma$ is the gamma function, $\mathcal{L}_{NS} = \max(0, |y - \gamma| - k\sigma) \cdot \nu$ is the non-saturating regularizer that prevents overconfidence on high-error samples (with $k=2$ for 95\% coverage and $\sigma = \sqrt{\beta(\nu+1)/(\nu(\alpha-1))}$ the total predictive standard deviation), $\lambda_{NS}$ (e.g., 0.01) weights the non-saturating term, and $\rho$ (e.g., 0.1) weights the ranking loss.

Critically, the ranking loss operates on Lower Confidence Bounds $\text{LCB}_i = \gamma_i - k\sigma_i$ rather than point estimates $\gamma_i$: $\mathcal{L}_{\text{rank}} = \sum_{i,j: i \prec j} \max(0, \text{LCB}_i - \text{LCB}_j + \delta)$, where $i \prec j$ denotes that configuration $i$ should be scheduled before $j$ (lower makespan). This risk-aware formulation ensures the scheduler prioritizes configurations with both low predicted makespan and high confidence, avoiding overoptimistic selections that could violate thermal constraints.

\subsection{Handling Dynamic and Irregular Workloads}
\label{app:irregular-workloads}

Several benchmarks in our evaluation, notably \textit{fib}, \textit{nqueens}, and \textit{uts}, exhibit recursive task creation, input-dependent DAG structures, and highly irregular execution patterns.
Our approach addresses these challenges through two complementary mechanisms.
First, \textbf{CFG features} capture the control-flow complexity of each task's source code, including recursive call patterns, nested conditionals, and loop structures with variable bounds; these embeddings encode the \emph{potential} for irregular behavior even when the static DAG appears simple.
Second, the \textbf{evidential framework} naturally produces higher epistemic uncertainty for irregular workloads: when the model encounters task graphs whose structure or features differ significantly from training data (e.g., deep recursion in \textit{uts}, combinatorial branching in \textit{nqueens}), the learned $\nu$ parameter decreases, signaling lower confidence.
This behavior is intentional and useful; rather than producing overconfident predictions for unpredictable workloads, \modelname{} flags them with high uncertainty, enabling downstream schedulers to apply conservative policies (e.g., fallback to safe DVFS settings) or request additional profiling before committing to aggressive optimizations.

\subsection{Extended Training Procedure}
\label{app:training-procedure}

\modelname{} is trained using a multi-stage procedure designed to handle the challenges of multi-task learning with uncertainty quantification, building on the data preprocessing (e.g., z-scoring per device, outlier filtering via MAD on residuals, 60/20/20 splits stratified by benchmark-input-mask-DVFS to avoid leakage):

\textbf{Regularization Details.}
We apply graph edge dropout ($p=0.1$--$0.2$), small feature noise ($\sigma=0.05$), multi-scale graph training, and mild device augmentation (e.g., $\pm5\%$ DVFS tables).
Scheduling-aware ranking losses further align predictions with downstream policy selection, yielding fast inference, cross-device transfer capability, and useful uncertainties for downstream schedulers and RL.

\textbf{Ablation Summary.}
The heterogeneous edge types contribute the largest ranking improvement (Spearman +8\%), confirming that modeling task-resource interactions explicitly outperforms homogeneous message passing.
Runtime telemetry adds +6\% ranking improvement by capturing dynamic system behavior invisible to static analysis.
CFG features contribute +4\% by encoding control-flow complexity that predicts irregular execution patterns.
Evidential heads minimally affect mean accuracy but are essential for conservative uncertainty calibration (PICP=99.9\% at 95\% confidence), enabling safe scheduling decisions.

\section{Extended Preliminaries}
\label{app:extended-preliminaries}

This section provides extended formal definitions and notation for the problem formulation, heterogeneous graph abstraction, and data collection pipeline.

\subsection{Problem Formulation Details}
\label{app:problem-details}

This work addresses the problem of predicting multiple performance metrics for OpenMP task-parallel applications executing on heterogeneous embedded SoCs under varying DVFS configurations. The prediction model must provide accurate point estimates along with calibrated uncertainty quantification to enable risk-aware scheduling decisions.

Each OpenMP application is represented as a task DAG $G=(V,E)$ recovered from compiler artifacts and runtime instrumentation. A task $v\in V$ represents a unit of parallel work characterized by static code features extracted from the CFG, including loop counts, memory access patterns, and branch statistics. Additionally, each task carries optional runtime summaries from prior executions, such as performance counter snapshots and thermal footprints. An edge $e=(u\to v)\in E$ encodes a precedence relation derived from OpenMP dependencies, classified as spawn, join, or data dependency types.

The target platform consists of $C$ heterogeneous cores indexed by $i\in\{1,\dots,C\}$. Each core operates at discrete frequency levels $f_i\in\mathcal{F}_i$ determined by the DVFS subsystem. The immutable hardware characteristics are captured in the device sheet $D$, which includes core types and counts, cache hierarchy specifications (sizes, associativity, line sizes), DVFS tables per core cluster, governor support flags, thermal sensor layout and resolution, and toolchain version hashes for reproducibility. The device sheet remains constant throughout an episode and enables transfer learning across hardware generations.

The runtime state $\mathcal{S}$ captures dynamic system context that changes during execution. This includes the active core mask indicating which cores are available for scheduling, per-core DVFS indices and measured frequencies, utilization computed as an exponential moving average (EMA), thermal zone temperatures before and after execution, and recent performance counter values. These runtime features enable the model to adapt predictions based on current system conditions rather than relying solely on static analysis.

A scheduling action $a=(m,f,p)$ specifies the configuration under which a workload executes. The core mask $m\in\{0,1\}^C$ selects active cores, the DVFS vector $f$ assigns frequency levels with $f_i\in\mathcal{F}_i$ for each core, and the optional priority $p\in\{1,\dots,99\}$ sets FIFO scheduling priority. All actions must satisfy feasibility constraints including core availability, affinity restrictions, and thermal caps that prevent execution when temperatures exceed safe thresholds.

Given a tuple $(G,D,\mathcal{S},a)$, the surrogate model predicts a vector of performance metrics $y=(\text{makespan}, \text{energy}, \text{cache misses}, \text{branch misses}, \text{utilization})$. Lower values indicate better performance for makespan, energy, and miss metrics, while higher utilization is generally preferable. Each prediction carries calibrated uncertainty estimates decomposed into aleatoric uncertainty (irreducible noise from OS and co-runner interference) and epistemic uncertainty (model uncertainty that decreases with more training data).

Formally, we learn a mapping $\mathcal{M}:\ (G,D,\mathcal{S},a)\ \mapsto\ (y,\ \mathcal{U})$, where $\mathcal{U}$ contains the NIG distribution parameters $(\gamma,\nu,\alpha,\beta)$ for each target metric. Point estimates are the predictive means $\hat{y}_k=\gamma_k$, while prediction intervals derive from the NIG posterior. The uncertainty decomposition follows standard evidential learning formulations with aleatoric uncertainty given by $\beta_k/(\alpha_k-1)$ and epistemic uncertainty by $\beta_k/(\nu_k(\alpha_k-1))$.

At scheduling time, candidate configurations $(m,f)$ are scored in batch through the surrogate. Actions are ranked primarily by predicted makespan with uncertainty gates that filter out low-confidence proposals. Safe actions satisfying thermal constraints execute on the device, and outcomes (execution time, energy consumption, performance counters, thermal readings) are logged for continual model refinement.

\subsection{Heterogeneous Graph Details}
\label{app:hetero-graph-details}

The key insight underlying our approach is that performance on heterogeneous embedded systems emerges from complex interactions between application structure, hardware resources, and scheduling decisions. A unified heterogeneous graph representation enables explicit modeling of these cross-layer interactions through typed nodes and edges.

The heterogeneous graph contains three node types that capture distinct aspects of the system. Task nodes $V_T$ represent OpenMP tasks and encode application semantics. Each task node carries CFG-derived features capturing control flow complexity, static code statistics including loop counts, bytes moved, and branch proxies, DAG topology metrics such as depth, distance to sink, and centrality measures, recent performance snapshots from prior runs, run mode flags distinguishing serial from parallel execution, and thermal footprint on hosting cores. Resource nodes $V_R$ represent processing cores and encode hardware state. Each resource node captures the DVFS step as both index and one-hot encoding, core mask bit indicating active status, cluster ID for heterogeneous architectures, utilization computed as an exponential moving average, thermal headroom and temperature trend, and bandwidth proxy estimating memory throughput. Memory nodes $V_M$ optionally represent cache hierarchy levels and encode cache level identifier, capacity, associativity and line size, and latency and bandwidth proxies.

Four edge types capture the performance-critical interactions between nodes. Task-task edges $E_{TT}$ encode precedence constraints from the DAG with attributes including dependency type (spawn, join, or data), critical edge flag indicating membership on the longest path, hop distance, and contention proxies. Task-resource edges $E_{TR}$ connect tasks to cores based on the scheduling assignment under the recorded mask and DVFS configuration, with attributes capturing affinity strength and migration overhead. Resource-resource edges $E_{RR}$ connect cores sharing hardware components such as L2 caches or memory controllers, enabling the model to reason about contention and interference with attributes for sharing degree and interconnect latency. Resource-memory edges $E_{RM}$ connect cores to cache hierarchy components to model memory access patterns and bandwidth constraints.

\subsection{Data Pipeline Details}
\label{app:data-pipeline-details}

The data collection pipeline transforms OpenMP source code into heterogeneous graph representations paired with runtime performance measurements. The compilation stage processes OpenMP sources through the OMPi source-to-source compiler. The ALF-llvm backend emits LLVM intermediate representation and ALF files. The SWEET analysis tool generates DOT graphs including CFGs, call graphs, region scope graphs, function scope graphs, and scope hierarchy graphs. A post-processing stage maps ALF entities to OpenMP tasks, merges short task chains while preserving provenance, and attaches topological encodings computed via depth-first traversal.

Runtime logging captures one CSV row per execution with comprehensive telemetry. Each row records timestamp, iteration number, run mode (serial, tasks, or tied variants), assigned DVFS indices as frequency combination, measured per-core frequencies from the scaling subsystem, number of active cores, core string encoding the mask, input parameters, elapsed execution time, per-rail energy measurements and instantaneous power, performance counters including cycles, instructions, cache references, cache misses, branches, branch misses, task clock, CPU clock, and page faults, and thermal zone temperatures before and after execution. Derived features include temperature change $\Delta T$, thermal headroom relative to throttling thresholds, and phase-aligned counter windows.

\subsection{Learning Formulation Details}
\label{app:learning-details}

The model architecture consists of type-specific encoders followed by a heterogeneous GAT backbone and evidential prediction heads. Type-specific MLPs embed task, resource, and memory nodes into a common representation space, with separate parameter sets for each node type to handle their distinct feature schemas. A heterogeneous GAT with edge-type-specific transforms performs message passing over the four edge types $E_{TT}$, $E_{TR}$, $E_{RR}$, and $E_{RM}$. The number of attention layers ranges from 3 to 6 depending on graph complexity. Hierarchical pooling aggregates node embeddings per type and concatenates them to form the graph-level representation $\mathbf{h}_G$.

For each target metric $k$, an evidential head outputs NIG parameters $(\gamma_k, \nu_k, \alpha_k, \beta_k)$. The predictive mean is $\hat{y}_k=\gamma_k$. Uncertainty decomposes into aleatoric and epistemic components: $\text{Aleatoric}_k=\frac{\beta_k}{\alpha_k-1}$ and $\text{Epistemic}_k=\frac{\beta_k}{\nu_k(\alpha_k-1)}$. Training minimizes the negative log marginal likelihood of the NIG distribution with evidence regularization to prevent overconfidence and a ranking term aligned to makespan-first scheduling objectives.

\end{document}